\documentclass[sigconf]{acmart}
\usepackage{cite}
\usepackage{algorithmic}
\usepackage{graphicx}
\usepackage{textcomp}
\usepackage{xcolor}

\usepackage{microtype}

\usepackage{subfigure}
\usepackage{booktabs} 
\usepackage{multicol}
\usepackage{multirow}
\usepackage{makecell}
\usepackage{caption}
\usepackage{enumitem}
\newtheorem{definition}{Definition}
\usepackage{wrapfig}
\usepackage{paralist}

\DeclareMathOperator*{\argmax}{argmax} 

\usepackage{microtype}

\AtBeginDocument{%
  \providecommand\BibTeX{{%
    \normalfont B\kern-0.5em{\scshape i\kern-0.25em b}\kern-0.8em\TeX}}}

\setcopyright{none}

\begin{document}

\title{Time Series Forecasting with Hypernetworks Generating Parameters in Advance}


\author{Jaehoon Lee$^1$, Chan Kim$^1$, Gyumin Lee$^1$, Haksoo Lim$^1$, Jeongwhan Choi$^1$, Kookjin Lee$^2$, Dongeun Lee$^3$, Sanghyun Hong$^4$, Noseong Park$^1$}
\authornote{1: Yonsei University ,2: Arizona State University, 3: Texas A\&M University-Commerce, 4: Oregon State University}
\email{{jaehoonlee, rlacks10, minzzang68, limhaksoo96, jeongwhan.choi, noseong}@yonsei.ac.kr}
\email{kjlee8344@gmail.com, dongeun.lee@tamuc.edu, sanghyun.hong@oregonstate.edu}











\begin{abstract}
Forecasting future outcomes from recent time series data is not easy, especially when the future data are different from the past (i.e. time series are under temporal drifts). Existing approaches show limited performances under data drifts, and we identify the main reason:
It takes time for a model to collect sufficient training data and \emph{adjust} its parameters for complicated temporal patterns whenever the underlying dynamics change.
To address this issue, we study a new approach; 
instead of adjusting model parameters (by continuously re-training a model on new data), we build a \emph{hypernetwork} that generates other target models' parameters expected to perform well on the future data. Therefore, we can adjust the model parameters beforehand (if the hypernetwork is correct).
We conduct extensive experiments with 6 target models, 6 baselines, and 4 datasets, and show that our \textit{HyperGPA} outperforms other baselines.

%
%

\end{abstract}

\keywords{Time-series forecasting, Hypernetwork, Temporal drifts}

\maketitle

\section{INTRODUCTION}
\label{sec:intro}


Time series forecasting is one of the most fundamental problems in deep learning, ranging from classical climate modeling~\citep{debrouwer2019gruodebayes,REN2021100178} and stock price forecasting~\citep{7046047,VIJH2020599} to recent pandemic forecasting~\citep{wu2020deep,pmlr-v144-wang21a}. Since these tasks are important in many real-world applications, diverse methods have been proposed, such as recurrent neural networks (RNNs)~\citep{lstm,cho2014learning}, neural ordinary differential equations (NODEs)~\citep{chen2019neural}, neural controlled differential equations (NCDEs)~\citep{kidger2020neural}, and so on. Owing to these novel models, the forecasting accuracy has been significantly enhanced over the past several years.

\begin{figure}[!ht]
    \centering
    \includegraphics[width=\columnwidth]{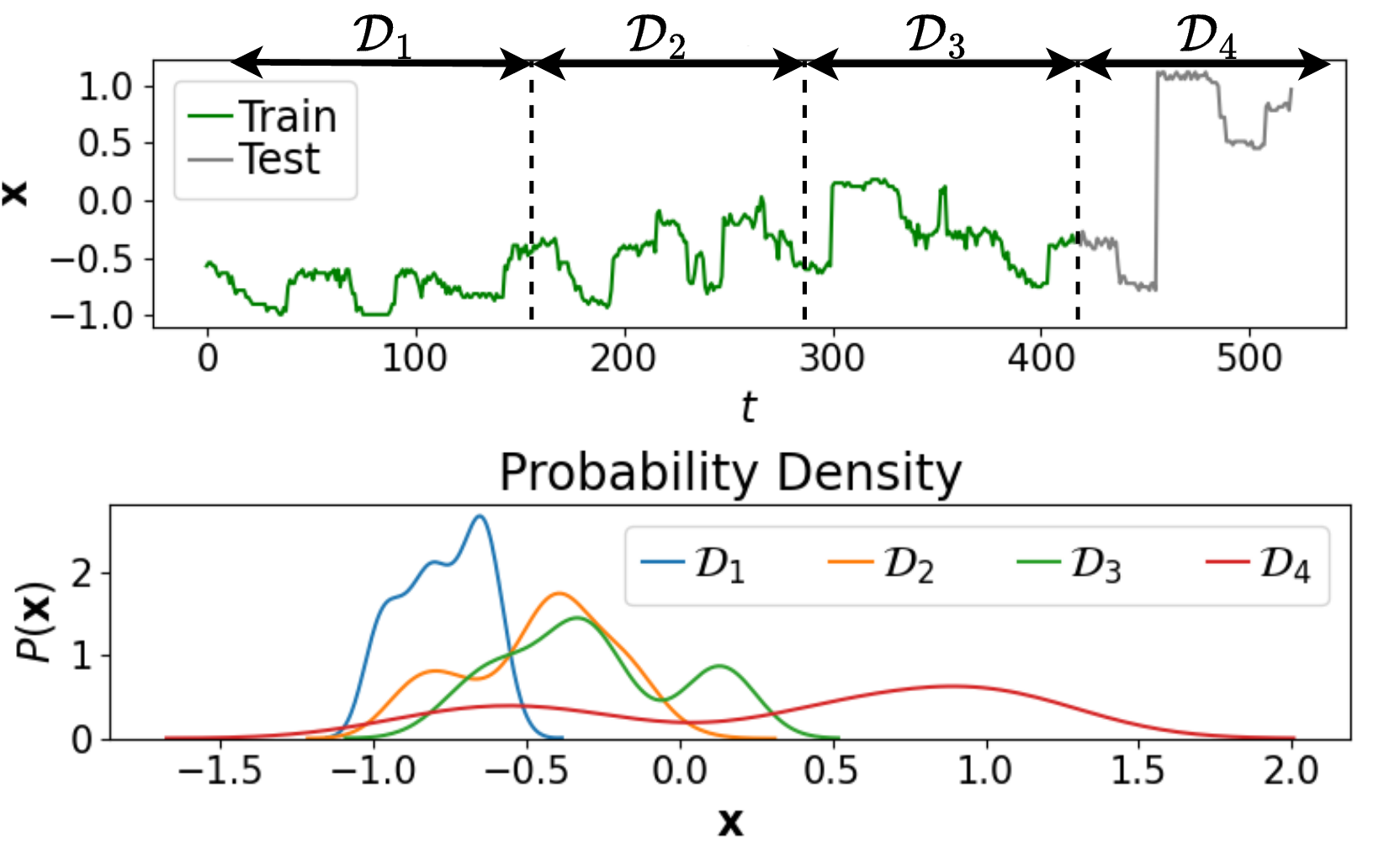}
    \caption{Time series data under temporal drifts. $\mathcal{D}_4$ is an unseen period to be forecast. Each period $\mathcal{D}_j$ shows a different probability density.}
    \label{fig:archia}
\end{figure}

%
However, this forecasting is challenging, especially when \emph{temporal drifts}, i.e.,  a data distribution changes over time by an underlying latent dynamics, exist in time series data (cf. Fig.~\ref{fig:archia})~\citep{10.5555/3042817.3043028,Oh2019RelaxedPS,DBLP:journals/corr/abs-2111-03422,10.1007/978-3-319-11662-4_19,NIPS2015_41f1f191}. 
For example, the number of COVID-19 patients fluctuates severely over time, and a dynamics behind the daily patient numbers is governed by complicated factors. 

\begin{figure}[!ht]
    \centering
    \includegraphics[width=\columnwidth]{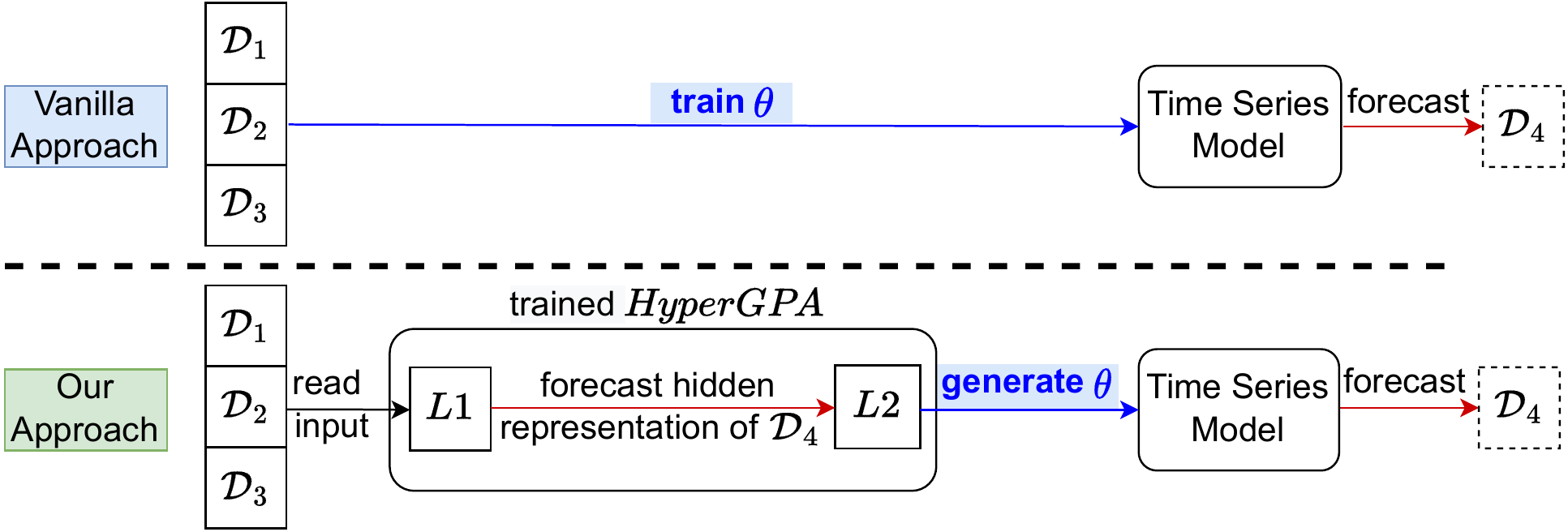}
    \caption{In a vanilla setting, a time series model is trained with all historical periods. On the contrary, in \textit{HyperGPA}, the first layer ($L1$) understands the recent periods ($\mathcal{D}_1,\mathcal{D}_2,\mathcal{D}_3$) and 
    forecasts a hidden representation for the next period ($\mathcal{D}_4$) with the discovered underlying dynamics.
    Based on the forecast hidden representation, the second layer ($L2$) generates the parameters of a time series model which would work well for $\mathcal{D}_4$.}
    \label{fig:archib}
\end{figure}

To defend against the latent dynamics, we propose a new approach that enables highly accurate time series forecasting under temporal drifts.
Specifically, we present a \underline{\textbf{Hyper}}network \underline{\textbf{G}}enerating \underline{\textbf{P}}arameters in \underline{\textbf{A}}dvance (\textbf{HyperGPA}). 
In usual settings, to forecast future data, one can train the parameters of a target time series model with data from all historical periods. On the other hand, \textit{HyperGPA} is based on hypernetworks which are neural networks generating the parameters of other neural networks (called \emph{target models})~\citep{ha2016hypernetworks}. Our hypernetwork has two parts as depicted in Fig.~\ref{fig:archib}: one ($L1$) captures the hidden dynamics given recent time series data, and the other ($L2$) generates (i.e., forecasts)  a set of parameters for a target time series forecasting model.  To our knowledge, we are the first proposing a hypernetwork-based approach for addressing temporal drift problems.


In \textit{HyperGPA}, $L1$ is responsible for discovering a hidden underlying dynamics and forecasting a future period's characteristic from data in recent periods. 
We use NCDEs for $L1$ since they are a continuous analogue to RNNs and are known to be superior to RNNs in processing a complicated dynamics~\citep{exit,gtgan}. 
To reflect real-world scenarios,
we assume $M$ loosely coupled time series data. For instance, the United States flu dataset~\citep{CDC} contains weekly reports for each state, which consists of 51 loosely coupled time series, i.e., one time series for each state. Because the underlying dynamics of the correlated multiple time series are also correlated, considering them at the same time can make the task easier by increasing the number of training samples.
To process multiple time series simultaneously, we integrate recent graph convolutional networks (GCNs) and NCDEs into a single framework --- we learn the latent graph of those $M$ loosely coupled time series from data and apply the GCN technology~\citep{bai2020adaptive}. Therefore, the first part of our hypernetwork can be understood as a \emph{shared multi-task} layer in that it processes multiple time series simultaneously.

From the hidden representation of the future period forecast by $L1$, $L2$ generates the future parameters of a target time series model. The parameters are generated based on GCNs since target neural network models are typically represented by computation graphs. 
However, if we generate parameters for each time series and for each period independently, the target models equipped with the generated parameters are vulnerable to an overfitting problem.
This is because we assume $M$ loosely coupled time series and different periods of a time series are also correlated to each other to some extent, although they may have different distributions. To solve this problem, we propose an attention-based generation method~\citep{attention2}, where we first generate several candidate sets of parameters and then combine them via the attention coefficients to obtain the final parameters.
Our proposed attention-based method has the following advantages: i) the number of parameters of \textit{HyperGPA} itself is not large; ii) \textit{HyperGPA} can relieve an overfitting problem of target models;  iii) \textit{HyperGPA} can be successfully applied to almost all existing popular time series forecasting models, ranging from RNNs and NODEs to NCDEs, which is a significant enhancement in comparison with other hypernetwork-based approaches which assume a specific target model type~\citep{ha2016hypernetworks,zhang2019anodev2}.
To summarize, we make the following contributions:

\begin{compactenum}
    \item \textbf{Novel Approach to Temporal Drifts:} We propose a hypernetwork, called \emph{HyperGPA}, that can forecast the future parameters of target time series models from previous periods. Aided by the NCDE technology,
    we show that our model is able to capture the pattern of temporal drift dynamics.
    \item \textbf{Novel Approach to Hypernetwork:} \emph{HyperGPA} generates parameters for each period via our proposed attention-based method, alleviating overfitting and resulting in a small model size. Also, it can be applied to almost all existing popular time series forecasting models, ranging from LSTMs/GRUs to NODEs/NCDEs.
    \item \textbf{Effectiveness:} \emph{HyperGPA} improves target models' errors up to 64.1\% with their small parameter sizes, compared to six baselines which include not only existing approaches addressing the temporal drift problem, but also hypernetwork-based approaches.
    \item \textbf{Reproducibility:} Our code is available in the supplementary material and we refer the reader to Appendix~\ref{appendix:reproducibility} for the details of reproducing our experimental results.
\end{compactenum}

%
\section{RELATED WORK \& MOTIVATION}
\label{subsec:related-work}

\subsection{Hypernetwork}\label{sec:hypernetwork}

In general, neural networks generating the parameters of other neural networks are called \emph{hypernetworks}. We call a neural network whose parameters are generated by a hypernetwork as a \emph{target model}. Hypernetworks can be used for many purposes.  
Interpreted as implicit distributions, hypernetworks give help for uncertainty estimations, which are new methods for more flexible variational approximation learning~\citep{pawlowski2018implicit,Henning2018ApproximatingTP,bayeshyper}. \citep{vonoswald2020continual} used hypernetworks for continual learning. Hypernetworks were also used for neural architecture search~\citep{nas}, and for personalized federated learning which aims to train models for multiple clients with respective data distributions~\citep{pmlr-v139-shamsian21a}.


Futhermore, hypernetworks helped target models to have high representation learning capabilities, by allowing the parameters of target models to be changed~\citep{decodehyper,3dhyper,kddhyperst}.  In particular, time series models were considered as applications in \textit{HyperRNN}~\citep{ha2016hypernetworks} and \textit{ANODEV2}~\citep{zhang2019anodev2}. One of the limitations of RNNs is that their parameters remain the same for every time step and every input sequence. \textit{HyperRNN} was presented to overcome this limitation, by generating their parameters for each input sample and time. \textit{HyperLSTM} and \textit{HyperGRU}, a hypernetwork for LSTMs and GRUs, are defined in~\citep{ha2016hypernetworks} and Appendix~\ref{appendix:hypergru}, respectively.

\textit{ANODEV2} was devised for improving the performance of NODEs. \textit{ANODEV2} makes NODEs more flexible, evolving both a hidden path and a parameter path. As such, the parameters of NODEs evolve over time, governed by another neural network with a set of learnable parameters. 


\subsection{Temporal Drifts in Time Series}\label{sec:tempdrift}

When source and target distributions are different, domain adaptation~\citep{NIPS2006_b1b0432c,da1,da2} or domain generalization~\citep{8578664,dg1,dg2} can be a solution. 
However, these methods cannot be used for time series data under temporal drifts, because the distribution (domain) of this data continuously changes.
Therefore, different strategies are needed to handle temporal drifts.

Adaptive RNN (\textit{AdaRNN}) is a variant of RNNs for addressing temporal drifts~\citep{adarnn}. With the principle of maximum entropy, it splits a time series into different periods, each of which is expected to have a unique distribution. To find commonalities in the different periods, \textit{AdaRNN} tries to reduce discrepancies in a hidden space between the different periods. 
The reversible instance normalization (\textit{RevIN}) is a state-of-the-art method to defend against the distribution drifts of time series and is an adaptation layer that can be used for any time series model~\citep{anonymous2022reversible}. Before a time series model takes an input, \textit{RevIN} normalizes the input across the temporal dimension. In addition, \textit{RevIN} de-normalizes the output returned by the time series model. By the normalization, changing statistical properties which cause drift problems can be relieved. 

Meanwhile, \citep{icdmdrift} proposed a general framework that considers not only temporal drifts but also evolving features where the feature space of a time series dynamically changes. The framework addresses the temporal drifts and evolving features with dynamically changing micro-clusters.
However, it was devised mainly for clustering or classification, whereas our object is future forecasting. As such, we do not consider this method in our experiment.

\begin{table}
\centering
\scriptsize
\setlength{\tabcolsep}{0.6pt}
\caption{Comparison of approaches tackling temporal drifts}\label{tbl:diff}
\begin{tabular}{|c|cc|}
\hline
Method & How to address temporal drifts & \# of time series models \\ \specialrule{1pt}{1pt}{1pt} 

\textit{AdaRNN} & Focuses on commonalities across all periods & 1 for all periods \\
\textit{RevIN} & Removes varying properties from input & 1 for all periods \\
\hline
\textit{HyperGPA} & Discovers a drift dynamics \& generates params. & 1 for each period \\
\hline
\end{tabular}
\end{table}
\textit{AdaRNN} and \textit{RevIN} solve a temporal drift problem to some extent. However, their common limitation is that they learn a general time series model applicable to all periods, each of which may have a unique distribution. As a result, the general model may not be optimal for each period because of the distribution difference. 
%
%
By using hypernetworks, we can overcome this problem, generating a specific time series model for each period.
However, as experimental results in Sec.~\ref{sec:experiments} show, existing approaches (\textit{HyperRNN} and \textit{ANODEV2}) cannot solve a temporal drift problem effectively. 
In addition, they cannot utilize two pieces of information: the correlated underlying hidden dynamics of multiple loosely coupled time series and the computation graph of a target model.
In Sec.~\ref{sec:abl}, we observe that these two kinds of information are all essential for accurate forecasting. To address these issues, we propose \textit{HyperGPA}, which can discover underlying hidden dynamics behind inter-correlated multiple time series. It generates parameters with the computation graph of target models depending on the characteristics of each period forecast by the discovered underlying hidden dynamics. Table~\ref{tbl:diff} compares the characteristics of each method dealing with temporal drifts.


\section{HYPERGPA}
\label{sec:hypergpa}

\subsection{Definitions \& Notations}
\label{sec:probdef}

First, the temporal drift in time series is defined as follows:
\begin{definition}

(Temporal Drifts)
Let a time series be spilt into $N$ disjoint periods (e.g., years or months) as in Fig.~\ref{fig:archia}, i.e., $\{\mathcal{D}_{j}\}_{j=1}^N$.
 There are temporal drifts in the time series, when there is at least one pair of periods that have different distributions \textit{i.e.}, 
 $\exists (j,k): \{ \mathbf{x} \in \mathbb{R}^{\dim(\mathbf{x})} | P_{\mathcal{D}_{k}}(\mathbf{x}) \neq P_{\mathcal{D}_{j}}(\mathbf{x})\} \neq \varnothing.$
 
 
\end{definition}\label{problem1} 

\begin{figure}[!ht]
    \centering
    \subfigure[An example of multiple time series data  $\mathcal{D}_{i,j}$ (the U.S. flu dataset) where each year constitutes a period]{\includegraphics[width=\columnwidth]{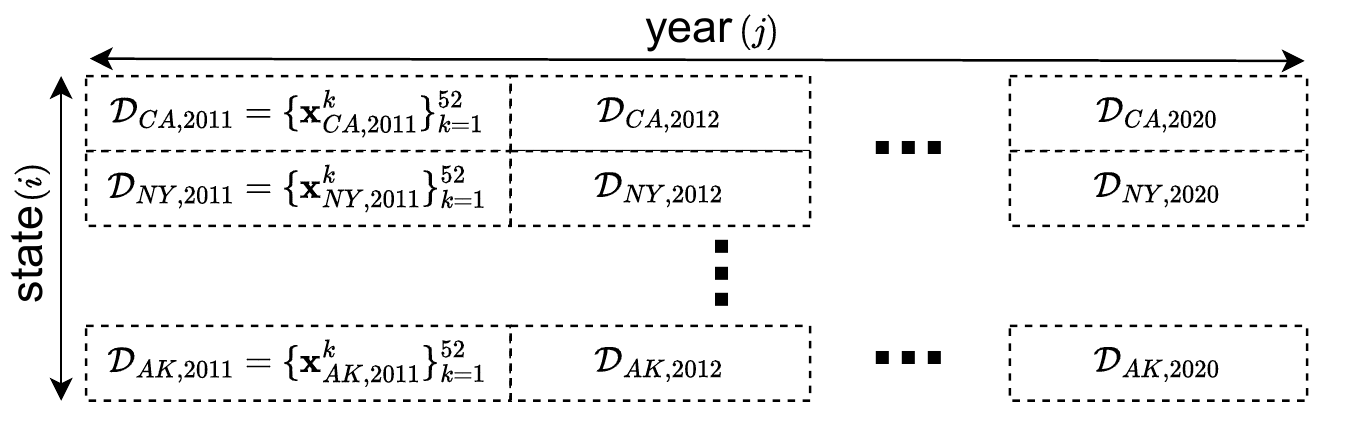}}
    \subfigure[ Overall workflow of \textit{HyperGPA}
     ]{\includegraphics[width=\columnwidth]{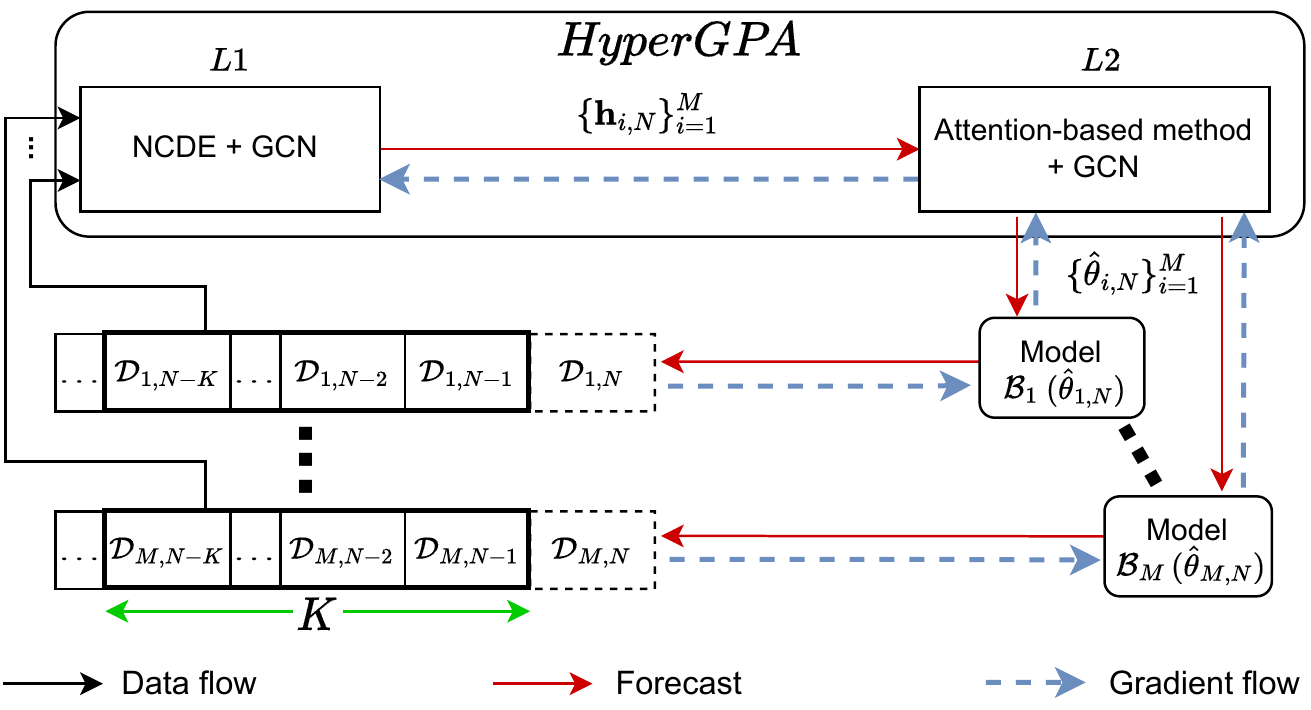}}
    \subfigure[Our training method]{\includegraphics[width=\columnwidth]{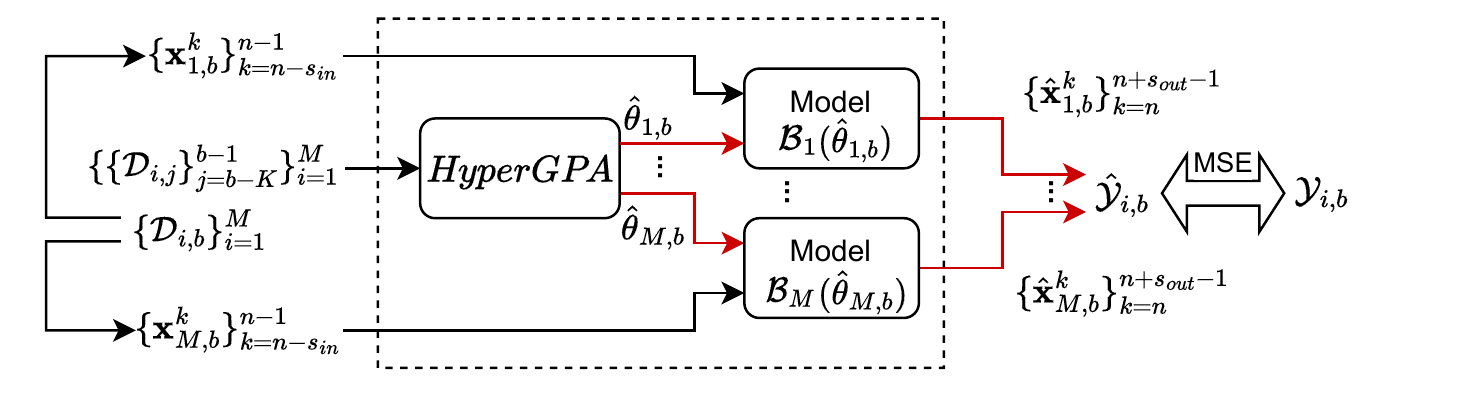}}
    \caption{(a) An illustration of multiple time series data. (b) \textit{HyperGPA} reads $K$ recent periods to generate target models' parameters for the next period. (c) \textit{HyperGPA} is trained by the MSE loss of target models in a seq-to-seq forecasting task.}
    \label{fig:archi3}
\end{figure}

As mentioned in Sec.~\ref{sec:intro}, we assume $M$ correlated time series. Given the $M$ time series, we define our time series and period concepts as follows (cf. Fig.~\ref{fig:archi3} (a)):
\begin{definition}\label{def:window}
(Time Series and Period)
Let $\mathcal{D}_{i,j} = \{\mathbf{x}_{i,j}^k\}_{k=1}^{|\mathcal{D}_{i,j}|}$ be the $i$-th time series at the $j$-th (disjoint) temporal period, where $\mathbf{x}_{i,j}^k \in \mathbb{R}^{\dim(\mathbf{x})}$ denotes the $k$-th observation. For simplicity but without loss of generality, we assume the observations of time series to be sampled regularly.
\end{definition} 

We use $M$ for the number of time series, i.e., $i=1,\ldots,M$, and $N$ for the number of periods, i.e., $j=1,\ldots,N$. Since we have $M$ time series, there are $M$ target models.
Then, for $M$ time series with $N$ periods 
$\{\{\mathcal{D}_{i,j}\}_{j=1}^{N}\}_{i=1}^{M}$, our problem definition is defined as follows:
\begin{definition}
(Forecasting under Temporal Drifts) Let the first $N-1$ periods $\{\{\mathcal{D}_{i,j}\}_{j=1}^{N-1}\}_{i=1}^{M}$ be train data, and the last $\{\mathcal{D}_{i,N}\}_{i=1}^{M}$ be test data. Provided that the time series are under temporal drifts, our goal is to train a hypernetwork to i) understand coupled underlying hidden dynamics causing the drifts, ii) predict the characteristics of $\{\mathcal{D}_{i,N}\}_{i=1}^{M}$ (as a form of latent vectors $\{\mathbf{h}_{i,N}\}_{i=1}^{M}$), and iii) generate the parameters of $M$ target models, each of which works well for $\mathcal{D}_{i,N}$.
\end{definition}\label{problem2}


 We represent the target model of the $i$-th time series as $\mathcal{B}_i$ whose parameters are generated by \textit{HyperGPA}. $\mathcal{B}_i$ can be either an RNN, NODE, or NCDE-based time series forecasting model. $\hat{\boldsymbol{\theta}}_{i,j}$ denotes the parameters of $\mathcal{B}_i$ which are generated by \textit{HyperGPA} and expected to work well in the $j$-th period.
 $K$ is an input period size, which is the number of previous periods for \textit{HyperGPA} to read (cf. Fig.~\ref{fig:archi3} (b)).
 
 The task of $\mathcal{B}_i$ is to forecast the next $s_{out}$ observations given past $s_{in}$ observations, i.e., seq-to-seq forecasting with an input size $s_{in}$ and an output size $s_{out}$. We represent the adjacency matrix of the computation graph of $\mathcal{B}_i$ as $\mathbf{A}$ --- we assume that all $\mathcal{B}_i$ have the same structure but with different parameters, and we simply use $\mathbf{A}$ without a subscript. And $v_l$ is the $l$-th vertex of $\mathbf{A}$, which denotes the $l$-th parameter of $\mathcal{B}_i$ (e.g., a weight or a bias). We also assume there are $L$ nodes in $\mathbf{A}$, i.e., $l=1,\ldots,L$.



\subsection{Overall Workflow}

We sketch our proposed method and introduce its step-by-step procedures as follows (cf. Fig.~\ref{fig:archi3} (b)):
\begin{compactenum}
    \item Let $\mathcal{D}_{input} = \{\{\mathcal{D}_{i,j}\}_{j=N-K}^{N-1}\}_{i=1}^M$ be the collected time series data for $K$ recent periods. 
    The task of $\mathcal{B}_i$ is seq-to-seq forecasting in $\mathcal{D}_{i,N}$.
    In this situation, our hypernetwork forecasts the model parameters of $\{\mathcal{B}_i\}_{i=1}^M$, each of which is expected to work well for $\mathcal{D}_{i,N}$. One can consider that our task is a \emph{multi-task learning problem} since there are $M$ similar tasks, i.e., forecasting the parameters for $M$ target models.
    
    \item In short, \textit{HyperGPA} can be expressed as $\{\hat{\boldsymbol{\theta}}_{i,N}\}_{i=1}^{M}$    
    $=L2(L1(\mathcal{D}_{input}), \mathbf{A})$, where $L1$ is a shared multi-task layer, and $L2$ is a parameter generating layer.

    \begin{compactenum}
        \item $\{\mathbf{h}_{i,N}\}_{i=1}^M = L1(\mathcal{D}_{input})$. The shared multi-task layer reads the previous periods $\mathcal{D}_{input}$ and forecasts $\{\mathbf{h}_{i,N}\}_{i=1}^M$, the hidden representation of the next period $\{\mathcal{D}_{i,N}\}_{i=1}^M$. At this step, we utilize the \emph{adaptive graph convolutional} (AGC) technique to discover \emph{latent} relationships among those $M$ different correlated time series~\citep{bai2020adaptive}. Therefore, our method can be used even when the explicit relationships among $M$ time series are unknown, which significantly improves the applicability of our method. In order to simultaneously process $M$ time series data, we integrate AGC and NCDEs into a single framework.
        
        \item  $\{\hat{\boldsymbol{\theta}}_{i,N}\}_{i=1}^{M} = L2(\{\mathbf{h}_{i,N}\}_{i=1}^M,\mathbf{A})$. The parameter generating layer forecasts the parameters of each target model $\hat{\boldsymbol{\theta}}_{i,N}$ from the forecast characteristics of the future period $\mathbf{h}_{i,N}$. We use an attention-based parameter generation method. Also, GCNs are utilized to consider connections between different nodes (e.g., weight, bias) in the adjacency matrix of a computation graph $\mathbf{A}$.

    \end{compactenum}    
    
    \item With the $i$-th target model $\mathcal{B}_i$ equipped with the forecast parameters $\hat{\boldsymbol{\theta}}_{i,N}$, we perform the seq-to-seq forecasting task in $\mathcal{D}_{i,N}$. 
\end{compactenum}

\subsection{$\boldsymbol{L1}$: Shared Multi-Task Layer}\label{sub:propose32}

We first design a shared multi-task layer to discover underlying dynamics and forecast hidden representations $\{\mathbf{h}_{i,N}\}_{i=1}^M$ from past periods. The representation $\mathbf{h}_{i,N}$ contains the expected future period characteristics of $\mathcal{D}_{i,N}$, from which the 
next
parameter generating layer forecasts the future model parameters $\hat{\boldsymbol{\theta}}_{i,N}$. If the forecast hidden representation is correct, i.e., if it represents the future period characteristics well, $\hat{\boldsymbol{\theta}}_{i,N}$ will also work well for the future period $\mathcal{D}_{i,N}$.

We use the following method to combine NCDEs and AGC into a single framework:
\begin{align}
    \mathbf{h'}_{i}(1) &= \Gamma_{i}(\mathbf{x}_{i}(1); \boldsymbol{\theta}_{\Gamma_{i}}),\label{eq:gamma}\\
    \mathbf{h'}_{i}(T) &= \mathbf{h'}_{i}(1) + \int_1^T G(\{\mathbf{h'}_{i}(t)\}_{i=1}^M;\boldsymbol{\theta}_G)\mathrm d\mathbf{X}_i(t) ,\label{eq:agc} \\
    \mathbf{h}_{i,N} &= \mathbf{h'}_{i}(T),\label{eq:omega}
\end{align}where $\mathbf{h'}_i(t)$ is an NCDE hidden vector of the $i$-th time series at integral time $t$; and the final integral time $T$ is the total time length of input for recent periods, $\sum_{j=N-K}^{N-1}|\mathcal{D}_{i,j}|$--- note that $T$ is the total number of observations because we assume regularly sampled time series (in general cases, we can set $T$ as the difference between the physical time of the first observation and that of the last observation in $\mathcal{D}_{input}$).
$T$ should be the same for all $i$-th time series because they should evolve together with a graph function in NCDEs. In addition, $\{\mathbf{x}_{i}(1)\}_{i=1}^M$ are feature vectors at initial time in $\mathcal{D}_{input}$, and $\{\mathbf{X}_i(t)\}_{i=1}^M$ are interpolated lines of $\mathcal{D}_{input}$. Each $\Gamma_i$ 
maps
the initial feature vector $\mathbf{x}_{i}(1)$ into the initial NCDE hidden vector $\mathbf{h'}_{i}(1)$. Along the interpolated path, NCDEs generate the final NCDE hidden vector, $\mathbf{h'}_{i}(T)$, which is $\mathbf{h}_{i,N}$. The important point to note is that $G$, which is shared by all $i$-th time series, includes an AGC function. Thus, all $i$-th time series are processed simultaneously through NCDEs and help each other to forecast the next hidden representation. 
We train the NCDE function with the adjoint method, so the space complexity for forward evaluations in NCDEs is constant.
\subsection{$\boldsymbol{L2}$: Parameter Generating Layer}\label{sub:propose33}
Given $\mathbf{h}_{i,N}$, we use an attention-based parameter generation method and GCNs to generate $\hat{\boldsymbol{\theta}}_{i,N}$:
\begin{align}
    \{\mathbf{z}_{i,N}^l\}_{l=1}^L &= \Phi(\mathbf{h}_{i,N};\boldsymbol{\theta}_{\Phi}),\label{eq:initquery}\\
    \{\mathbf{q}_{i,N}^l\}_{l=1}^L &= \text{GCN}(\{\mathbf{z}_{i,N}\}_{l=1}^L;\mathbf{A}, \boldsymbol{\theta}_{\text{GCN}}),\label{eq:gat}\\
    \mathbf{a}_{c,i,N}^l &= Softmax_c (\mathbf{q}_{i,N}^l \boldsymbol{\theta}_F^l), \label{eq:qk1}\\
    \mathbf{\hat{\boldsymbol{\theta}}}_{i,N} &= \{ \sum_{c=1}^{C} \mathbf{a}_{c,i,N}^l\Tilde{\boldsymbol{\theta}}_{c}^{l}\}_{l=1}^L.\label{eq:qk2}
\end{align} 
For parameter generation, we use an attention-based method. In this method, there are query, keys, and values, like~\citep{attention2}. Values are candidate parameters and each candidate has a key. $\mathbf{h}_{i,N}$ is mapped into query vectors. Because $\mathbf{h}_{i,N}$ includes the information of $\mathcal{D}_{i,N}$, query vectors should take the information of the period into account. Then, a generated target model $\mathbf{\hat{\boldsymbol{\theta}}}_{i,N}$ is the weighted sum of candidate parameters with attention coefficients from the key-query evaluation. (cf. Eq.~\ref{eq:qk1} and~\ref{eq:qk2})





$\mathbf{z}_{i,N}^l$, an initial query vector for the $l$-th parameter vertex $v_l$, is generated from $\mathbf{h}_{i,N}$ by a mapping
function $\Phi$. Then, $\{\mathbf{z}_{i,N}^l\}_{l=1}^L$ are transformed into final query vectors $\{\mathbf{q}_{i,N}^l\}_{l=1}^L$, considering the connection of each node with the adjacency matrix of a computation graph $\mathbf{A}$ via GCNs. $Softmax_c$ denotes the $c$-th elements after the softmax operation. $\Tilde{\boldsymbol{\theta}}_{c}^{l}$ is the $c$-th learnable candidate parameter for $v_l$, $C$ is the number of candidate parameters, and $\boldsymbol{\theta}_F^l \in \mathbb{R}^{dim(\mathbf{q}) \times C}$ is a learnable key matrix. According to query-key evaluations, the attention coefficients of candidate parameters is determined. This approach has several advantages over a simple mapping method where $\mathbf{h}_{i,N}$ are directly converted into $\mathbf{\hat{\boldsymbol{\theta}}}_{i,N}$ through several linear layers. We will show the improvement in the attention-based method in Sec.~\ref{sec:abl}.



One can use various graph networks, e.g., AGC, GAT~\citep{velickovic2018gatgraph}, and GCN~\citep{kipf2017semisupervised}. 
In all experiments, we select GAT as the graph convolutional network of $L2$. This is because the experiments in Appendix~\ref{appendix:adddesign} show that GAT is an reasonable choice although others sometimes produce better results than GAT.

\subsection{Training Algorithm}\label{sub:propose34}
Let $\{\{\mathcal{D}_{i,j}\}_{j=1}^{N-1}\}_{i=1}^M$ be the entire training time series data.  We use the following procedures to train our hypernetwork (cf. Fig.~\ref{fig:archi3} (c)):
\begin{compactenum}
    \item We create several mini-batches $(\{\{\mathcal{D}_{i,j}\}_{j=b-K}^{b-1}\}_{i=1}^M,$ $\{\mathcal{D}_{i,b}\}_{i=1}^M)$, where $K+1 \leq b \leq N-1$. Our hypernetwork reads $\{\{\mathcal{D}_{i,j}\}_{j=b-K}^{b-1}\}_{i=1}^M$ to generate the target model's parameters and the loss is computed with the target model's forecasting accuracy on $\{\mathcal{D}_{i,b}\}_{i=1}^M$. Given the range $[1, N-1]$, we can create multiple such mini-batches for training.
    
    \item We feed $\{\{\mathcal{D}_{i,j}\}_{j=b-K}^{b-1}\}_{i=1}^M$ into our hypernetwork, which produces $\{\hat{\boldsymbol{\theta}}_{i,b}\}_{i=1}^M$. Each target model $\mathcal{B}_i$ equipped with $\hat{\boldsymbol{\theta}}_{i,b}$ performs a seq-to-seq forecasting task in $\mathcal{D}_{i,b} = \{\mathbf{x}_{i,b}^k\}_{k=1}^{|\mathcal{D}_{i,b}|}$. $\mathcal{B}_i$ reads $\{\mathbf{x}_{i,b}^k\}_{k=n-s_{in}}^{n-1}$ to forecast $\{\mathbf{x}_{i,b}^k\}_{k=n}^{n+s_{out}-1}$. 
    
    \item Let $\mathcal{Y}_{i,b}$ be the ground-truth for $\mathcal{B}_i$ and $\hat{\mathcal{Y}}_{i,b}$ be the forecasting results by $\mathcal{B}_i$ configured with $\hat{\boldsymbol{\theta}}_{i,b}$. Our training method
    lets
    the hypernetwork produce the best parameters, which leads to the smallest possible mean squared error (MSE), $MSE_1$, between the ground-truth $\{\mathcal{Y}_{i,b}\}_{i=1}^M$ and the inference $\{\hat{\mathcal{Y}}_{i,b}\}_{i=1}^M$. 
    
    \item Also, we train \textit{HyperGPA} with another MSE, $MSE_2$, between $\{\mathcal{Y}_{i,b}\}_{i=1}^M$ and inference by a target model configured with $\{\Tilde{\boldsymbol{\theta}}_{\hat{c}_{i,b}^{l}}^{l}\}_{l=1}^L$, where $\hat{c}_{i,b}^{l}=\argmax\limits_c \mathbf{a}_{c,i,b}^l$. With $MSE_2$, each candidate becomes meaningful (i.e. forecasts well) in our task. The final loss is $MSE_1 + \lambda MSE_2$ where $\lambda$ is the regularization coefficient of $MSE_2$.

    
    
\end{compactenum}

\section{EXPERIMENTS}
\label{sec:experiments}
We now evaluate \textit{HyperGPA} on time series forecasting benchmarks. Our experiments are primarily designed to show how accurate our method is compared to other baselines. Specifically, we aim to answer the following research questions: 
\textbf{RQ.1}) How accurately does a target model generated by \textit{HyperGPA} forecast future outcomes, compared to other baselines? \textbf{RQ.2}) How does the size of a target model affect the performance under two possible approaches: \textit{HyperGPA} and a vanilla approach? \textbf{RQ.3}) How do various design choices of \textit{HyperGPA}, \textit{e.g.}, the type of graph functions, influence the ability to forecast future parameters?


\begin{table*}[t]
\scriptsize
\setlength{\tabcolsep}{13pt}
\centering
\caption{MSE improvements w.r.t. \textit{Vanilla}. Target models are in the \texttt{typewriter} font and baselines are in \textit{italics}.}\label{tbl:summary}
\begin{tabular}{|c|c|cc|cc|c|}

\hline
\multirow{2}{*}{Target Model} & \multirow{2}{*}{\textit{Vanilla Training}} & \multicolumn{2}{c|}{Hypernetwork} & \multicolumn{2}{c|}{Countermeasure for Temporal Drifts} & \multirow{2}{*}{\textit{HyperGPA}} \\
 &  & \makecell{\textit{HyperLSTM} (GRU)} & \textit{ANODEV2} & \textit{RevIN} & \makecell{\textit{AdaLSTM (GRU)}} &  \\ 
 \specialrule{1pt}{1pt}{1pt} 

\texttt{LSTM} & 0\%  & -105.3\% & N/A & -28.0\% & -217.6\% & \textbf{64.1\%} \\ 
\texttt{GRU} & 0\% & -137.5\% & N/A & -26.3\% & -178.8\% & 45.5\% \\ 
\makecell{\texttt{SeqToSeq(LSTM)}} & 0\% & -115.5\% & N/A & -20.0\% & N/A & 59.1\% \\ 
\makecell{\texttt{SeqToSeq(GRU)}} & 0\% & -131.3\% & N/A & -26.3\% & N/A & 44.3\% \\ 
\texttt{ODERNN} & 0\% & N/A & 23.1\% & -58.9\% & N/A & 63.8\% \\ 
\texttt{NCDE} & 0\% & N/A & -139.7\% & -41.1\% & N/A & 50.6\% \\ 
\hline

\end{tabular}
\end{table*}

\begin{table*}[t]
\scriptsize
\setlength{\tabcolsep}{3.5pt}
\centering
\caption{Experimental results. The best result for each target model is in \textbf{boldface} and for all target models with asterisk$^\ast$.}\label{tbl:all}
\begin{tabular}{|cc|cc|cc|cc|cc|}

\hline
\multirow{2}{*}{\makecell{Target \\ Model}} & \multirow{2}{*}{\makecell{Generating/Training \\Method}} & \multicolumn{2}{c|}{\texttt{Flu}} & \multicolumn{2}{c|}{\texttt{Stock-US}} & \multicolumn{2}{c|}{\texttt{Stock-China}} & \multicolumn{2}{c|}{\texttt{USHCN}} \\ 
& & MSE & PCC & MSE & PCC & MSE & PCC & MSE & PCC \\ \specialrule{1pt}{1pt}{1pt} 

\multirow{5}{*}{\texttt{LSTM}} & \textit{Vanilla} & 0.367±0.016 & 0.910±0.003 & 0.213±0.010 & 0.902±0.005 & 0.050±0.001 & 0.975±0.000 & 0.239±0.003 & 0.841±0.001\\
& \textit{HyperLSTM} & 0.582±0.019 & 0.852±0.005 & 0.751±0.103 & 0.605±0.071 & 0.103±0.007 & 0.949±0.003 & 0.249±0.004 & 0.824±0.003\\
& \textit{RevIN} & 0.506±0.097 & 0.917±0.010 & 0.063±0.001 & 0.966±0.000 & 0.049±0.002 & 0.977±0.001 & 0.589±0.015 & 0.693±0.009\\
& \textit{AdaLSTM} & 0.740±0.075 & 0.814±0.019 & 0.379±0.102 & 0.787±0.068 & 0.321±0.064 & 0.834±0.037 & 0.595±0.026 & 0.594±0.015\\
& \textit{HyperGPA} & \textbf{0.118±0.004} & \textbf{0.971±0.001}$^\ast$ & \textbf{0.050±0.002} & \textbf{0.973±0.000}$^\ast$ & \textbf{0.026±0.001} & \textbf{0.987±0.000} & \textbf{0.221±0.003} & \textbf{0.844±0.002}$^\ast$\\ \hline
\multirow{5}{*}{\texttt{GRU}} & \textit{Vanilla} & 0.275±0.006 & 0.933±0.002 & 0.102±0.003 & 0.952±0.002 & 0.037±0.002 & 0.982±0.001 & 0.232±0.001 & 0.840±0.001\\
& \textit{HyperGRU} & 0.520±0.033 & 0.863±0.010 & 0.462±0.046 & 0.772±0.030 & 0.075±0.002 & 0.963±0.002 & 0.244±0.004 & 0.832±0.002\\
& \textit{RevIN} & 0.379±0.039 & 0.938±0.004 & 0.060±0.001 & 0.967±0.000 & 0.040±0.000 & 0.981±0.000 & 0.465±0.007 & 0.699±0.008\\
& \textit{AdaGRU} & 0.616±0.035 & 0.849±0.009 & 0.233±0.043 & 0.875±0.028 & 0.170±0.017 & 0.913±0.009 & 0.472±0.026 & 0.676±0.018\\
& \textit{HyperGPA} & \textbf{0.116±0.004}$^\ast$ & \textbf{0.971±0.001}$^\ast$ & \textbf{0.052±0.002} & \textbf{0.972±0.001} & \textbf{0.026±0.001} & \textbf{0.987±0.000} & \textbf{0.229±0.004} & \textbf{0.841±0.003}\\ \hline
\multirow{4}{*}{\makecell{\texttt{SeqToSeq} \\ \texttt{(LSTM)}}} & \textit{Vanilla} & 0.353±0.005 & 0.914±0.002 & 0.167±0.005 & 0.926±0.003 & 0.045±0.001 & 0.978±0.001 & 0.236±0.003 & 0.835±0.002\\
& \textit{HyperLSTM} & 0.559±0.021 & 0.855±0.007 & 0.643±0.084 & 0.674±0.052 & 0.097±0.045 & 0.952±0.023 & 0.243±0.004 & 0.826±0.003\\
& \textit{RevIN} & 0.345±0.021 & 0.944±0.004 & 0.061±0.001 & 0.967±0.000 & 0.044±0.002 & 0.979±0.001 & 0.585±0.014 & 0.682±0.016\\
& \textit{HyperGPA} & \textbf{0.128±0.006} & \textbf{0.969±0.001} & \textbf{0.048±0.001}$^\ast$ & \textbf{0.973±0.001}$^\ast$ & \textbf{0.026±0.001} & \textbf{0.987±0.000} & \textbf{0.220±0.007}$^\ast$ & \textbf{0.843±0.002}\\ \hline
\multirow{4}{*}{\makecell{\texttt{SeqToSeq} \\ \texttt{(GRU)}}} & \textit{Vanilla} & 0.250±0.005 & 0.939±0.001 & 0.112±0.006 & 0.948±0.003 & 0.035±0.001 & 0.983±0.000 & 0.232±0.001 & 0.839±0.001\\
& \textit{HyperGRU} & 0.502±0.023 & 0.872±0.007 & 0.464±0.092 & 0.769±0.057 & 0.073±0.008 & 0.964±0.004 & 0.236±0.003 & 0.831±0.001\\
& \textit{RevIN} & 0.291±0.032 & 0.954±0.003 & 0.060±0.002 & 0.967±0.001 & 0.039±0.003 & 0.981±0.001 & 0.519±0.006 & 0.678±0.003\\
& \textit{HyperGPA} & \textbf{0.130±0.014} & \textbf{0.968±0.003} & \textbf{0.049±0.003} & \textbf{0.973±0.001}$^\ast$ & \textbf{0.025±0.000}$^\ast$ & \textbf{0.988±0.000}$^\ast$ & \textbf{0.222±0.004} & \textbf{0.844±0.005}$^\ast$\\ \hline
\multirow{4}{*}{\texttt{ODERNN}} & \textit{Vanilla} & 0.361±0.039 & 0.907±0.010 & 0.200±0.015 & 0.894±0.014 & 0.056±0.007 & 0.972±0.004 & 0.235±0.004 & 0.840±0.001\\
& \textit{ANODEV2} & 0.298±0.014 & 0.926±0.004 & 0.120±0.013 & 0.940±0.009 & 0.037±0.001 & 0.982±0.001 & 0.233±0.004 & \textbf{0.842±0.001}\\
& \textit{RevIN} & 0.549±0.226 & 0.905±0.008 & 0.068±0.001 & 0.964±0.000 & 0.048±0.002 & 0.978±0.001 & 0.855±0.046 & 0.529±0.021\\
& \textit{HyperGPA} & \textbf{0.134±0.016} & \textbf{0.967±0.004} & \textbf{0.050±0.001} & \textbf{0.972±0.000} & \textbf{0.026±0.001} & \textbf{0.987±0.000} & \textbf{0.226±0.007} & 0.840±0.004\\ \hline
\multirow{4}{*}{\texttt{NCDE}} & \textit{Vanilla} & 0.387±0.042 & 0.900±0.011 & 0.130±0.015 & 0.929±0.007 & 0.040±0.001 & 0.980±0.001 & 0.234±0.004 & 0.829±0.002\\
& \textit{ANODEV2} & 0.821±0.011 & 0.778±0.004 & 0.244±0.002 & 0.850±0.001 & 0.141±0.001 & 0.929±0.001 & 0.483±0.002 & 0.616±0.001\\
& \textit{RevIN} & 0.439±0.022 & 0.919±0.003 & 0.060±0.001 & 0.967±0.000 & 0.040±0.001 & 0.981±0.000 & 0.713±0.008 & 0.467±0.015\\
& \textit{HyperGPA} & \textbf{0.167±0.020} & \textbf{0.959±0.004} & \textbf{0.049±0.002} & \textbf{0.973±0.001}$^\ast$ & \textbf{0.027±0.001} & \textbf{0.987±0.001} & \textbf{0.227±0.004} & \textbf{0.837±0.003}\\ \hline

\end{tabular}
\end{table*}

\subsection{Experimental Setup}
\label{subsec:setup}

\noindent \textbf{Datasets.}
%
We evaluate our approach on four popular time series forecasting benchmarks ranging from pandemic and stock to climate forecasting datasets: \texttt{Flu}, \texttt{Stock-US}, \texttt{Stock-China}, and \texttt{USHCN}. These are all popular benchmark datasets. Refer to Appendix~\ref{appendix:data} for their detailed descriptions, including the configurations of the temporal window sizes, $s_{in}$ and $s_{out}$ in each dataset. In our evaluations for all these datasets, we use the last and the second to the last window as a test and a validation set, respectively.




\noindent \textbf{Target Models \& Baselines.}
We first explain the difference between a target model and a baseline. The target model is a time series model which performs a seq-to-seq forecasting task. On the other hand, the baseline is a way to generate or train the target model's parameters as done in \textit{HyperGPA}.

For target models, we use six time series models, ranging from RNN to NODE and NCDE-based models. For RNNs, we use \texttt{LSTM} and \texttt{GRU}~\citep{lstm,cho2014learning}. Because the task of a target model is to read a sequence and forecast its next sequence, we also use a seq-to-seq learning model (\texttt{SeqToSeq}) as a target model~\citep{sutskever2014sequence} --- note that this model can be based on either \texttt{LSTM} or \texttt{GRU}. In addition, we use a latent ODE model, \texttt{ODERNN}~\citep{rubanova2019latent}, and an NCDE-based model, \texttt{NCDE}~\citep{kidger2020neural}.

For baselines, there are 3 categories: i) In a vanilla method, one can directly train the parameters of a target model with recent data. There are neither hypernetworks nor approaches addressing temporal drifts. Target models trained this way is denoted by \textit{Vanilla} (cf. Fig.~\ref{fig:archib}). ii) There are two hypernetwork-based methods. \textit{HyperLSTM} or \textit{HyperGRU}~\citep{ha2016hypernetworks}, depending on the underlying cell types, can be applied to RNN-based models only, i.e., \texttt{LSTM}, \texttt{GRU}, \texttt{SeqToSeq(LSTM)}, and \texttt{SeqToSeq(GRU)}. Also, \textit{ANODEV2}~\citep{zhang2019anodev2} is a hypernetwork for NODEs that can be applied to NODE variants only (i.e., \texttt{ODERNN} and \texttt{NCDE}).
iii) \textit{RevIN}~\citep{anonymous2022reversible}, \textit{AdaLSTM}, and \textit{AdaGRU}~\citep{adarnn} are designed to address the temporal drifts. \textit{RevIN} can be used for various time series forecasting models, whereas \textit{AdaLSTM} and \textit{AdaGRU} are for \texttt{LSTM} and \texttt{GRU} only, respectively. 
Additionally, there is a traditional statistical model, \textit{ARIMA}~\citep{arima}. However, we include its results only in an appendix, because \textit{ARIMA} has the worst score overall.
All baselines are trained separately for each target model, whereas our method, which internally has a shared multi-task layer, is trained collectively and generates multiple target models' parameters simultaneously.

 

\noindent \textbf{Hyperparameters.}
We have two key hyperparameters for the target models, a hidden size and the number of layers.
The hidden size is in $\{16,32,64\}$ and the number of layers is in $\{1,2,3\}$. We use the same settings for the baselines. However, for \textit{HyperGPA}, we fix their hidden size and the number of layers to $16$ and $1$ respectively, which leads to the smallest target model size --- surprisingly, the smallest setting can also produce the best forecasting outcomes in many cases of our experiments, which shows the efficacy of our design. This is related to another advantage of \textit{HyperGPA} which will be described in Sec.~\ref{section:sens} (\textbf{RQ.2}). Refer to Appendix~\ref{appendix:hyperparam} for detailed hyperparameter configurations.

\noindent \textbf{Metrics.} We train \textit{HyperGPA} and the baselines with MSE --- the validation metric is also MSE. For testing, we use the pearson correlation coefficient (PCC) and MSE. We run 5 times for each evaluation case and report their mean and standard deviation (std. dev.) values.

\subsection{RQ.1: Time Series Forecasting Accuracy}
\label{section:exp}
\begin{figure}
    \centering
    \subfigure[Data: \texttt{Flu}, Target Model: \texttt{SeqToSeq (LSTM)} ]{\includegraphics[width=0.49\columnwidth]{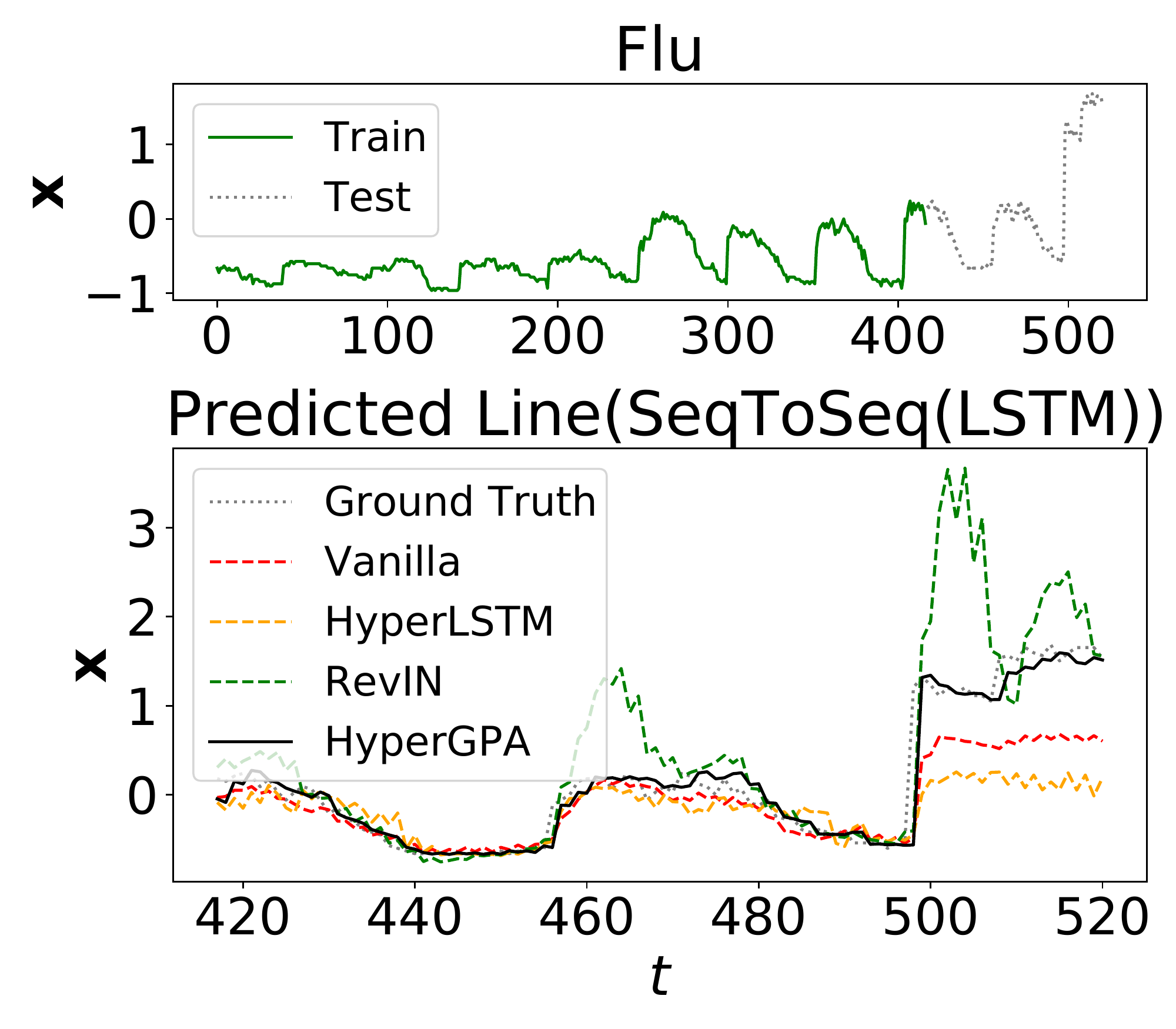}}
    \subfigure[\texttt{Stock-US}, \texttt{GRU}]{\includegraphics[width=0.49\columnwidth]{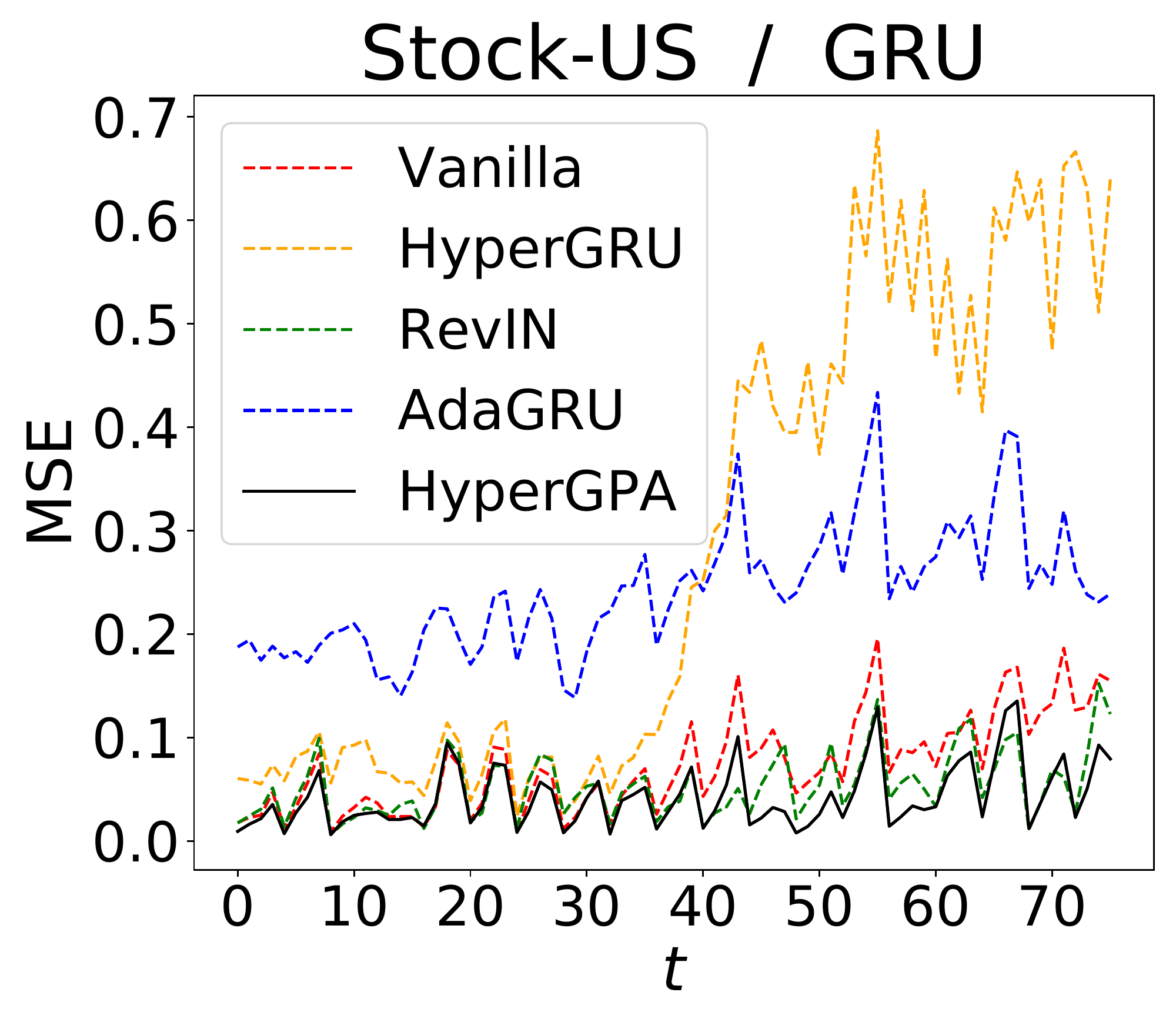}}
    \caption{(a) Forecast lines of various methods. The upper figure shows a train and test line. A validation line is included in the test line. The lower figure shows the predicted line of baselines and \textit{HyperGPA}. (b) MSE over time in various methods.}
    \label{fig:visual}
\end{figure}

Table~\ref{tbl:summary} shows the summary evaluation results which contain the improvement ratio of the baselines and HyperGPA over \textit{Vanilla} in MSE, i.e., $\frac{\textit{Vanilla's}\ \text{MSE} - \textit{HyperGPA(or baseline)'s}\ \text{MSE}}{\textit{Vanilla's}\ \text{MSE}}$. We average the ratio values across all datasets. 
Table~\ref{tbl:all} shows detailed experimental results. Also, there are visualizations. Fig.~\ref{fig:visual} (a) shows forecast lines, and Fig.~\ref{fig:visual} (b) shows MSE values over time. 
We refer the reader to Appendix~\ref{appendix:detailedresult} for full experimental results with all evaluation metrics, and Appendix~\ref{appendix:visual} for full visualizations.

\noindent \textbf{Summary.}
\textit{HyperLSTM/GRU} shows worse scores than those by \textit{Vanilla}. In all cases, they fail to increase the model accuracy. 
For \texttt{ODERNN}, \textit{ANODEV2} achieves better scores than those of \textit{Vanilla} but not for \texttt{NCDE}. One possible reason is that \textit{ANODEV2} was originally designed for NODEs only.

In the baseline approaches addressing temporal drifts, \textit{AdaLSTM/GRU} fails to address temporal drifts. We think that this is because the temporal drifts of the experimental datasets are too large for this approach to overcome, as in Fig.~\ref{fig:visual} (a). Also, \textit{RevIN} has poor performance inferior to \textit{Vanilla}. However, in spite of the poor summary scores, \textit{RevIN} is still an effective method for temporal drifts. We show relevant results in the following detailed result paragraphs.


Our model, \textit{HyperGPA}, consistently outperforms all baselines for all target models. Especially for \texttt{LSTM} and \texttt{ODERNN}, there are about 60\% improvements, proving the efficacy of the proposed method.


\noindent \textbf{Detail (\textbf{Flu}).}
In Table~\ref{tbl:all}, among the baselines, except for \textit{ANODEV2} in \texttt{ODERNN}, all baselines are worse than \textit{Vanilla} in terms of MSE. However, when checking PCC and the visualization in Fig.~\ref{fig:visual} (a), \textit{RevIN} alleviates the temporal drift problem to some extent. \textit{RevIN} catches a changed dynamics more accurately in Fig.~\ref{fig:visual} (a), and achieves a higher PCC than \textit{Vanilla} in Table~\ref{tbl:all}.

Although \textit{RevIN} alleviates temporal drifts, it has worse MSE scores. We think that in \textit{RevIN}, target models do not consider the data magnitude because of the normalization, and when the data magnitude is drastically changing, the target models become unstable. The unstable forecasting visualization can be checked in Fig.~\ref{fig:visual} (a) around $t\approx460$ or $500$.



Meanwhile, our \textit{HyperGPA} outperforms others in almost all cases, and achieves the lowest MSE. In Fig.~\ref{fig:visual} (a), \textit{HyperGPA} forecasts the changed dynamics almost exactly. 
Also, \textit{HyperGPA} performs better in all types of the target model,
which shows the adaptiveness of our method.


\noindent \textbf{Detail (\textbf{Stock}).}
Among baselines, in the two stock datasets, \textit{RevIN} mostly shows the lowest MSE scores in Fig.~\ref{fig:visual} (b) over time, which leads to improvement over \textit{Vanilla} in Table~\ref{tbl:all}. We observe that \textit{RevIN} is helpful for temporal drifts in the two datasets. However, \textit{HyperGPA} attains more enhanced performance in all aspects, showing better evaluation scores in Table.~\ref{tbl:all} and Fig.~\ref{fig:visual} (b).

\noindent \textbf{Detail (\textbf{USHCN}).}
\textit{HyperGPA} achieves the best scores in almost all cases, whereas
there is no improvement in other baselines, compared to \textit{Vanilla}, except for \textit{ANODEV2} in \texttt{ODERNN}. Even \textit{RevIN} shows poor performance. These results in all datasets show that \textit{HyperGPA} can successfully generate the parameters of all target models which work well in the future.

\subsection{RQ.2: Accuracy by Varying Target Model Size}
\label{section:sens}

\begin{figure}[t]
    \centering
    \includegraphics[width=\columnwidth]{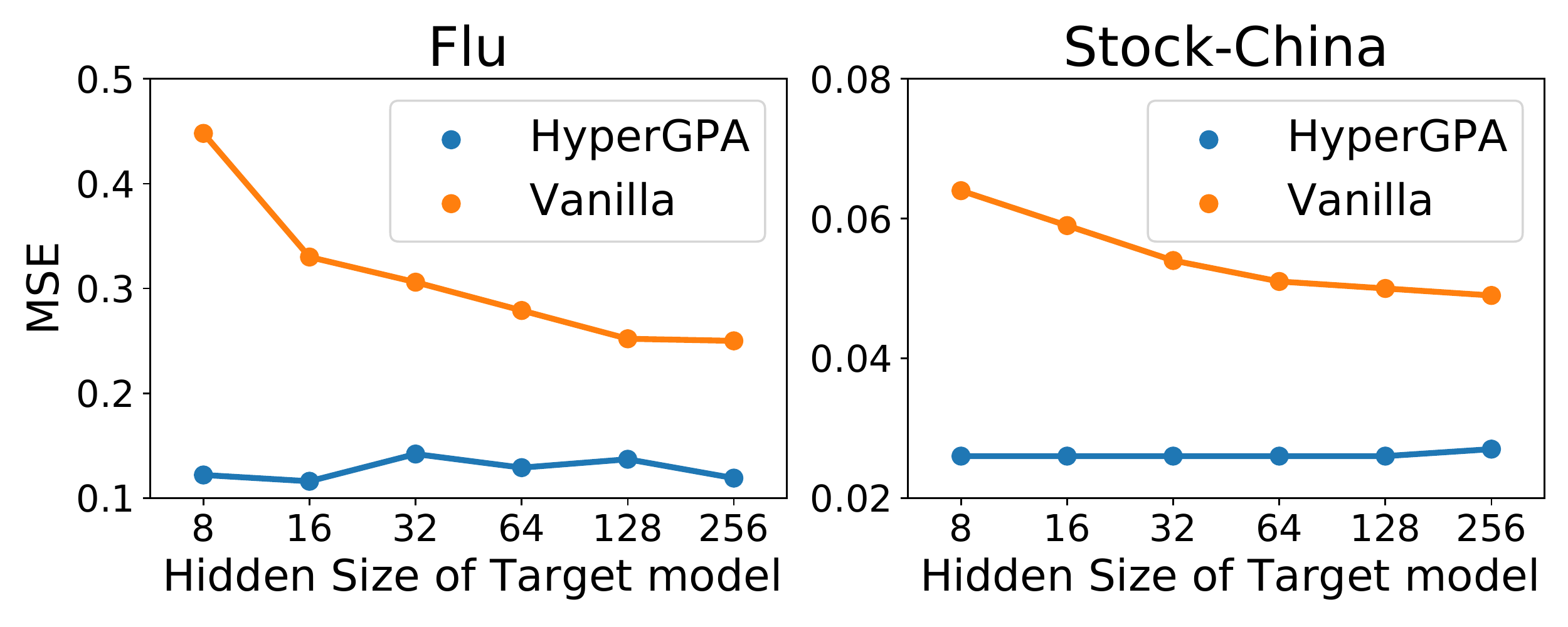}
    \caption{Sensitivity to the hidden size of the two target models, \texttt{GRU}(\texttt{Flu}), and \texttt{LSTM}(\texttt{Stock-China})}
    \label{fig:sens}
\end{figure}

We compare \textit{Vanilla} and \textit{HyperGPA}, varying the hidden size of target models from 8 to 256 --- the larger the hidden size, the larger the target model size. As in Fig.~\ref{fig:sens}, \textit{HyperGPA} shows stable errors regardless of the hidden size of a target model whereas \textit{Vanilla} does not show reliable forecasting accuracy when the target model size is small. This result shows that we can use small target models with \textit{HyperGPA}, which drastically reduces the overheads for maintaining target models in practice.

\begin{figure}
    \centering
    \subfigure[Ablation w.r.t. using GCNs or not]{\includegraphics[width=0.49\columnwidth]{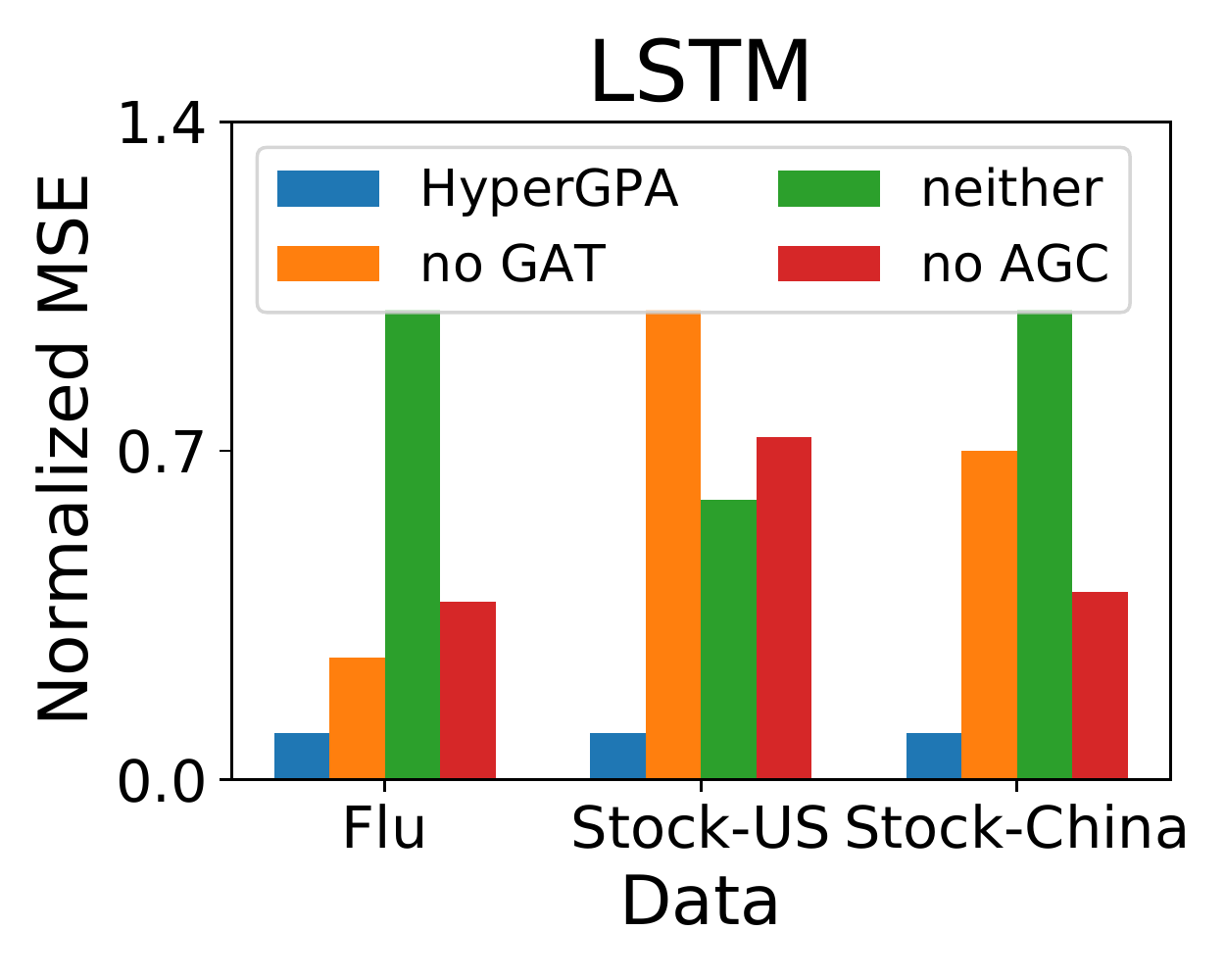}}
    \subfigure[The required model size of \textit{HyperGPA} vs. the hidden size of target models]{\includegraphics[width=0.49\columnwidth]{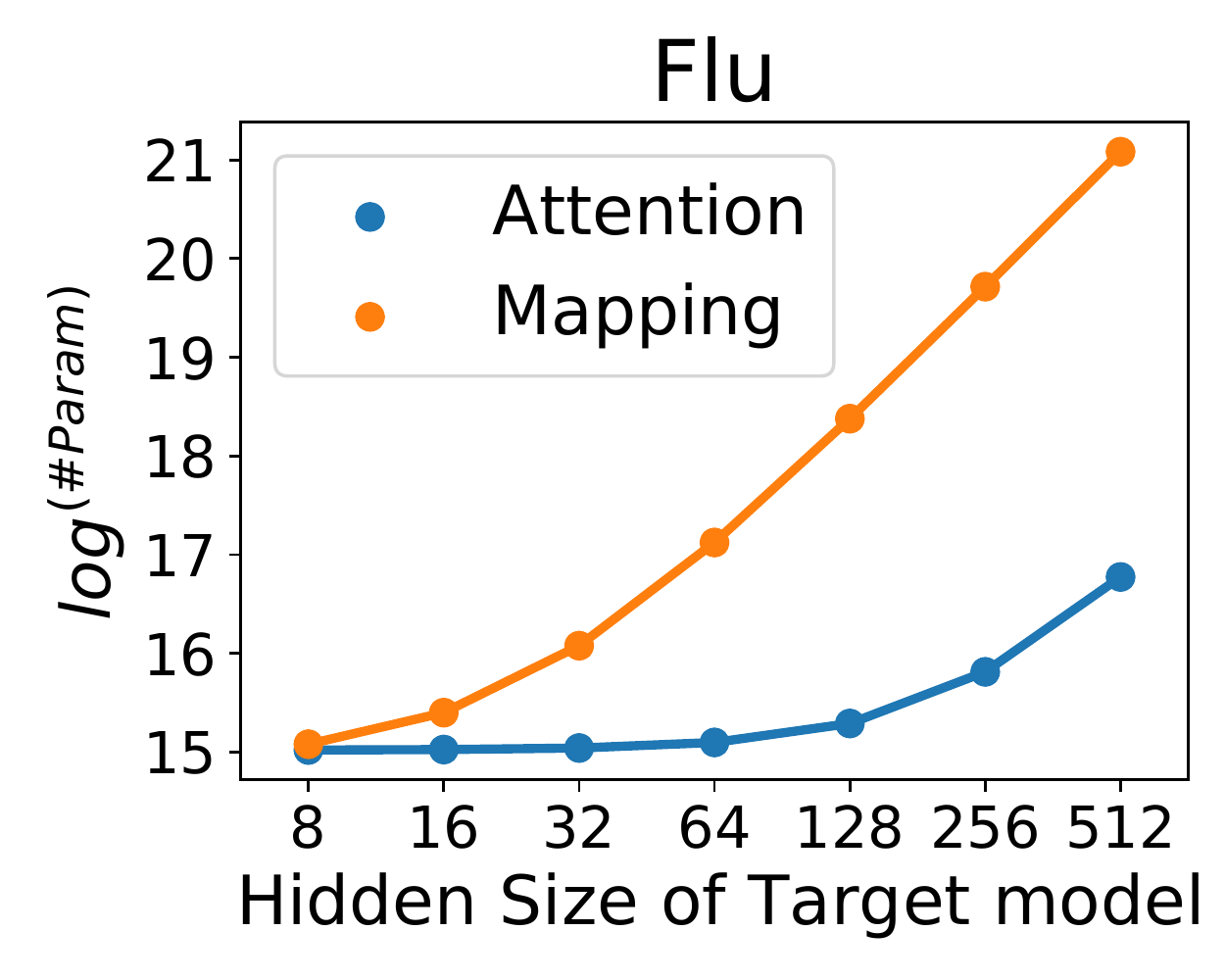}}
    \subfigure[Train and test MSE of the two possible parameter generation methods in \texttt{Flu}. `Attention' means our full model.]{\includegraphics[width=\columnwidth]{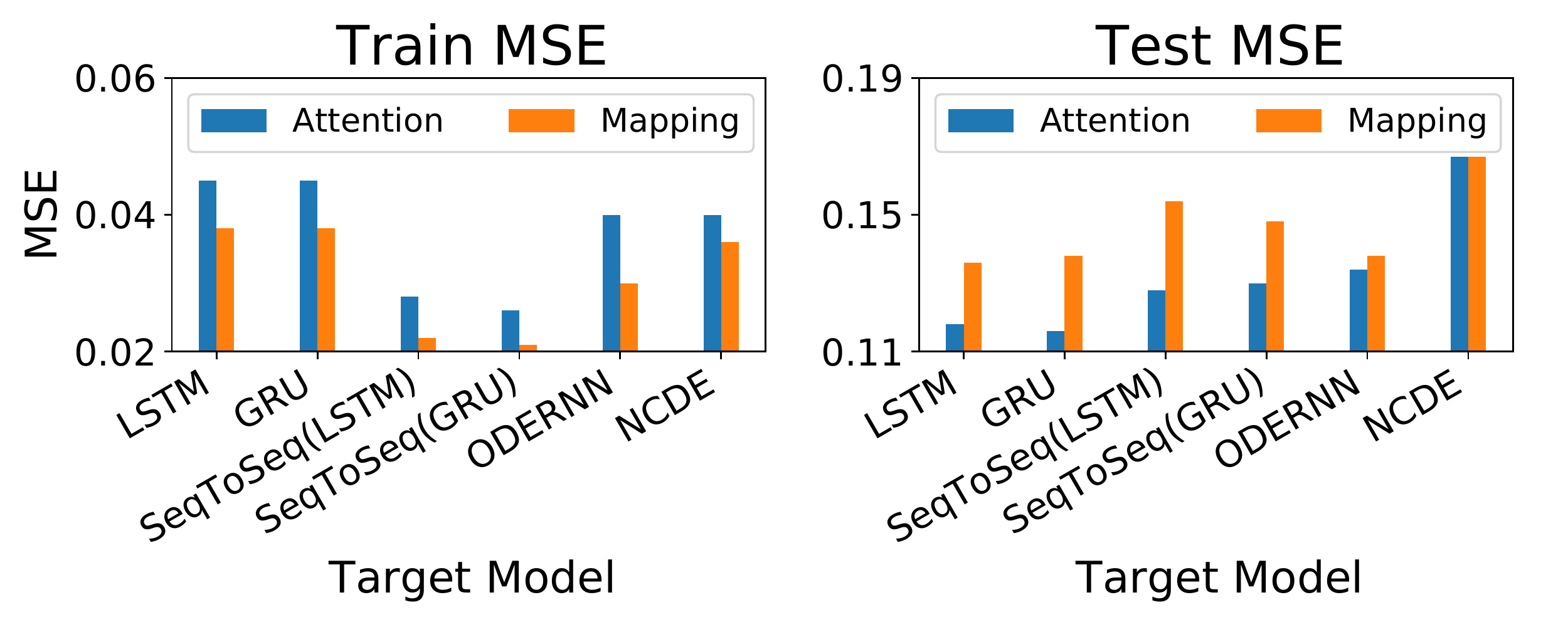}}
    \caption{Experiments for an design choice}
    \label{fig:design}
\end{figure}

\subsection{RQ.3: Ablation Studies}\label{sec:abl}

We justify our specific design choices for \textit{HyperGPA}. Refer to Appendix~\ref{appendix:adddesign} for experiments on other design choices.

\noindent \textbf{With or Without Graph Function.} In order to define two ablation models, we remove AGC from $L1$ (Eq.~\eqref{eq:agc}) or GAT from $L2$ (Eq.~\eqref{eq:gat}) --- note that we use GAT as the graph convolution function of $L2$.
Fig.~\ref{fig:design} (a) shows MSE for each case where the target model is \texttt{LSTM}. \textit{HyperGPA} denotes our full model; \textit{neither} means our model without AGC or GAT. The normalized MSE is used for fair comparison. In almost all cases, \textit{HyperGPA} shows the lowest errors. Higher errors of \textit{no AGC} show that considering different correlated time series simultaneously helps \textit{HyperGPA} to better discover underlying dynamics. Also, \textit{no GAT} has worse results, which means that the information of a computation graph is needed for generating reliable parameters. As such, we conclude that both the computation graph of a target model and the correlation between multiple time series are helpful for accurate model parameter generations.

\noindent \textbf{Attention-Based Method or Simple Mapping Method.}
Fig.~\ref{fig:design} (b) and (c) show the advantages of our attention-based parameter generation method in Sec.~\ref{sub:propose33}, compared to the simple mapping method where $\mathbf{h}_{i,N}$ are directly mapped into the parameters of a target model through linear layers. According to Fig.~\ref{fig:design} (b), as the hidden size of target models increases, the required number of parameters of \textit{HyperGPA} increases more rapidly in the simple mapping method than in the attention-based method. In addition, our attention-based method is more robust to overfitting as in Fig.~\ref{fig:design} (c). 








\section{CONCLUSION}
\label{sec:conclusion}

We presented a novel approach to defend against the temporal drift by designing a hypernetwork generating the future parameters of a target model. Our hypernetwork, called \textit{HyperGPA}, is \emph{model-agnostic} as our experiments showed that it works well for all the tested target models and marks the best accuracy in most cases. In addition, we showed that with \textit{HyperGPA}, small target models are sufficient to be used in real-world applications (if their parameters are carefully configured by our method), which drastically reduces the maintenance overhead of target models. 

\noindent \textbf{Limitations.} Because \textit{HyperGPA} is based on NCDEs which are unknown for their accuracy for long time series, we will further extend our research towards it.

\noindent \textbf{Societal Impacts.} Our method can enhance time series forecasting which has many real-world applications, e.g., forecasting stock prices, weather conditions, etc. In addition, we showed that our method results in using small target models, which reduce carbon emissions during inference. 
Because our main contribution is to improve forecasting performances under temporal drifts, we think \textit{HyperGPA} has no significant negative impact.



\newpage
\onecolumn
\begin{multicols}{2}
    \bibliography{our_work}
    \bibliographystyle{ACM-Reference-Format}
\end{multicols}

\newpage
\appendix

\section{HYPERGRU}\label{appendix:hypergru}
We define \textit{HyperGRU} since only \textit{HyperLSTM} is defined in~\citep{ha2016hypernetworks}. Let $\mathbf{x}_t$ be a time series observation at time $t$. GRU~\citep{cho2014learning} is defined as follows:
\begin{align}
\mathbf{r}_t &= \sigma( W^{\mathbf{r}}_{\mathbf{x}}\mathbf{x}_t + W^{\mathbf{r}}_{\mathbf{h}}\mathbf{h}_{t-1} +  b^{\mathbf{r}}), \\
\mathbf{z}_t &= \sigma( W^{\mathbf{z}}_{\mathbf{x}}\mathbf{x}_t + W^{\mathbf{z}}_{\mathbf{h}}\mathbf{h}_{t-1} +  b^{\mathbf{z}}), \\
\mathbf{g}_t &= \delta( W^{\mathbf{g}}_{\mathbf{x}}\mathbf{x}_t  + \mathbf{r}_t \odot W^{\mathbf{g}}_{\mathbf{h}}\mathbf{h}_{t-1} +  b^{\mathbf{g}} ), \\
\mathbf{h}_t &= (1-\mathbf{z}_t) \odot \mathbf{g}_t + \mathbf{z}_t \odot \mathbf{h}_{t-1},
\end{align} where $\sigma$ is a sigmoid function, $\delta$ is a hyperbolic tangent function, and $\odot$ is an element-wise multiplication. Likewise, \textit{HyperGRU} is defined as follows:
\begin{align}
\hat{\mathbf{x}}_t &= \mathbf{x}_{t} \oplus \mathbf{h}_{t-1}, \\
\hat{\mathbf{r}}_t &= \sigma( LN(W^{\hat{\mathbf{r}}}_{\hat{\mathbf{x}}}\hat{\mathbf{x}}_t + W^{\hat{\mathbf{r}}}_{\hat{\mathbf{h}}}\hat{\mathbf{h}}_{t-1} +  b^{\hat{\mathbf{r}}} )), \\
\hat{\mathbf{z}}_t &= \sigma( LN(W^{\hat{\mathbf{z}}}_{\hat{\mathbf{x}}}\hat{\mathbf{x}}_t + W^{\hat{\mathbf{z}}}_{\hat{\mathbf{h}}}\hat{\mathbf{h}}_{t-1} +  b^{\hat{\mathbf{z}}} )), \\
\hat{\mathbf{g}}_t &= \delta( LN(W^{\hat{\mathbf{g}}}_{\hat{\mathbf{x}}}\hat{\mathbf{x}}_t  + \hat{\mathbf{r}}_t \odot W^{\hat{\mathbf{g}}}_{\hat{\mathbf{h}}}\hat{\mathbf{h}}_{t-1} +  b^{\hat{\mathbf{g}}} )), \\
\hat{\mathbf{h}}_t &= (1-\hat{\mathbf{z}}_t) \odot \hat{\mathbf{g}}_t + \hat{\mathbf{z}}_t \odot \hat{\mathbf{h}}_{t-1},
\end{align} where $\oplus$ means a concatenation. From $\hat{\mathbf{h}}_t$, the embeddings of each gate $\mathbf{a}^{\mathbf{h},y}_t, \mathbf{a}^{\mathbf{x},y}_t, \mathbf{a}^{\mathbf{b},y}_t$ are generated, where $y \in \{\mathbf{r},\mathbf{z},\mathbf{g}\}$. In addition, those embeddings are transformed into $\mathbf{d}^{\mathbf{h},y}_t, \mathbf{d}^{\mathbf{x},y}_t, \mathbf{d}^{\mathbf{b},y}_t$:
\begin{align}
\mathbf{a}^{\mathbf{h},y}_t &= W_{\hat{\mathbf{h}}}^{\mathbf{h},y}\hat{\mathbf{h}}_{t-1} + b_{\hat{\mathbf{h}}}^{\mathbf{h},y}, \\ 
\mathbf{a}^{\mathbf{x},y}_t &= W_{\hat{\mathbf{h}}}^{\mathbf{x},y}\hat{\mathbf{h}}_{t-1} + b_{\hat{\mathbf{h}}}^{\mathbf{x},y}, \\
\mathbf{a}^{\mathbf{b},y}_t &= W_{\hat{\mathbf{h}}}^{\mathbf{b},y}\hat{\mathbf{h}}_{t-1} + b_{\hat{\mathbf{h}}}^{\mathbf{b},y}, \\
\mathbf{d}^{\mathbf{h},y}_t &= W_{\mathbf{a}}^{\mathbf{h},y}\mathbf{a}^{\mathbf{h},y}_{t} + b_{\mathbf{a}}^{\mathbf{h},y}, \\
\mathbf{d}^{\mathbf{x},y}_t &= W_{\mathbf{a}}^{\mathbf{x},y}\mathbf{a}^{\mathbf{x},y}_{t} + b_{\mathbf{a}}^{\mathbf{x},y}, \\
\mathbf{d}^{\mathbf{b},y}_t &= W_{\mathbf{a}}^{\mathbf{b},y}\mathbf{a}^{\mathbf{b},y}_{t} + b_{\mathbf{a}}^{\mathbf{b},y}.
\end{align}
Finally, $\mathbf{d}^{\mathbf{h},y}_t, \mathbf{d}^{\mathbf{x},y}_t, \mathbf{d}^{\mathbf{b},y}_t$ adjust the parameters of the main GRU and $h_t$ is generated:
\begin{align}
\mathbf{r}_t &= \sigma(  LN(\mathbf{d}^{\mathbf{x},\mathbf{r}}_t \odot W^{\mathbf{r}}_{\mathbf{x}}\mathbf{x}_t + \mathbf{d}^{\mathbf{h},\mathbf{r}}_t \odot W^{\mathbf{r}}_{\mathbf{h}}\mathbf{h}_{t-1} +  \mathbf{d}^{\mathbf{b},\mathbf{r}}_t )), \\
\mathbf{z}_t &= \sigma(  LN(\mathbf{d}^{\mathbf{x},\mathbf{z}}_t \odot W^{\mathbf{z}}_{\mathbf{x}}\mathbf{x}_t + \mathbf{d}^{\mathbf{h},\mathbf{z}}_t \odot W^{\mathbf{z}}_{\mathbf{h}}\mathbf{h}_{t-1} +  \mathbf{d}^{\mathbf{b},\mathbf{z}}_t )), \\
\mathbf{g}_t &= \delta(  LN(\mathbf{d}^{\mathbf{x},\mathbf{g}}_t \odot W^{\mathbf{g}}_{\mathbf{x}}\mathbf{x}_t + \mathbf{r}_t \odot \mathbf{d}^{\mathbf{h},\mathbf{g}}_t \odot W^{\mathbf{g}}_{\mathbf{h}}\mathbf{h}_{t-1} +  \mathbf{d}^{\mathbf{b},\mathbf{g}}_t )), \\
\mathbf{h}_t &= (1-\mathbf{z}_t) \odot \mathbf{g}_t + \mathbf{z}_t \odot \mathbf{h}_{t-1}.
\end{align}


\section{ADDITIONAL DESIGN CHOICE}\label{appendix:adddesign}

\begin{figure}
\centering

\begin{minipage}[t]{0.49\textwidth}
    \centering
    \includegraphics[width=0.98\columnwidth]{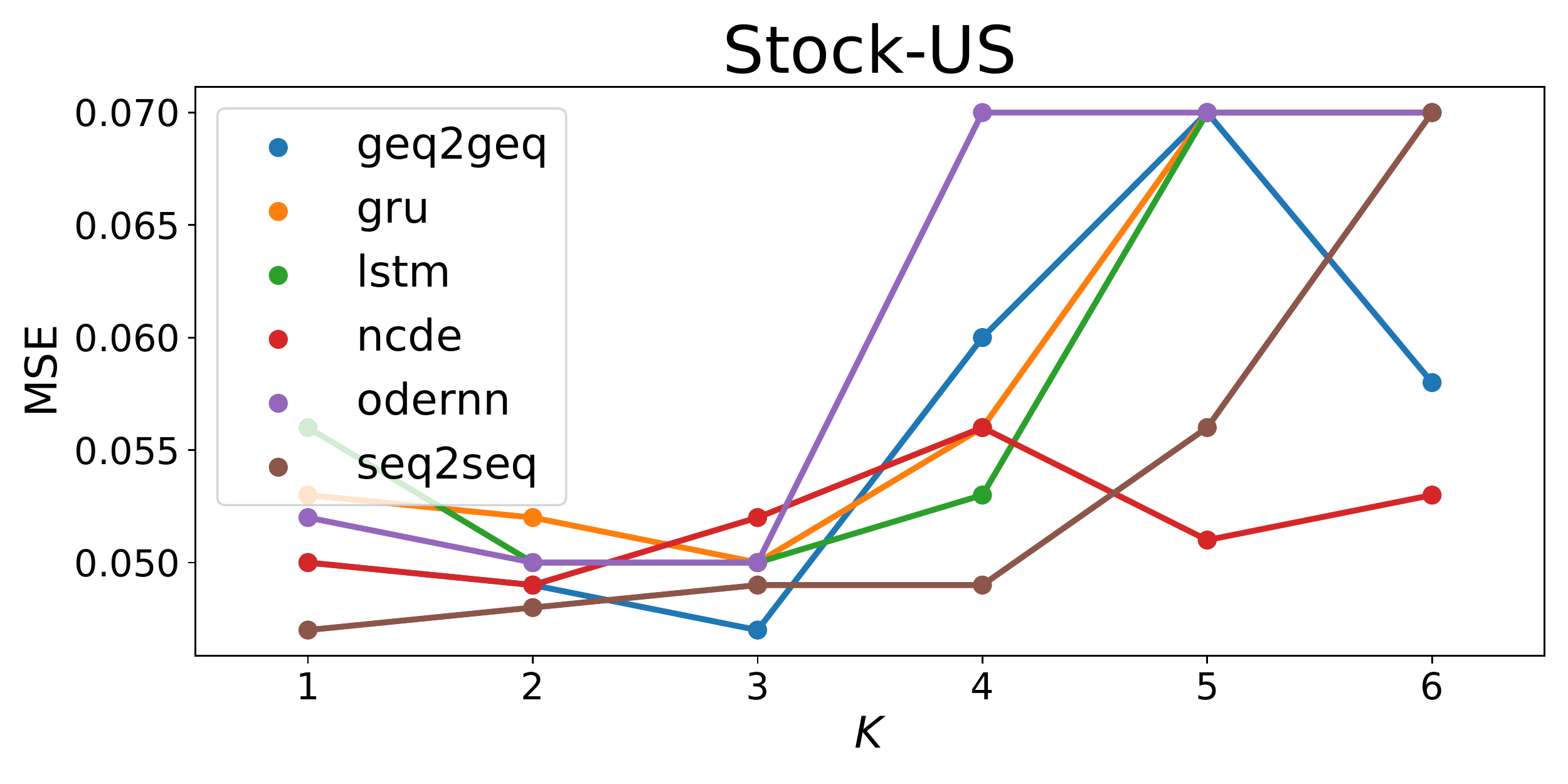}
    \caption{Sensitivity to the input period size $K$}
    \label{fig:sens2}
\end{minipage}
\hfill
\begin{minipage}[t]{0.49\textwidth}
    \centering
    \includegraphics[width=0.49\columnwidth]{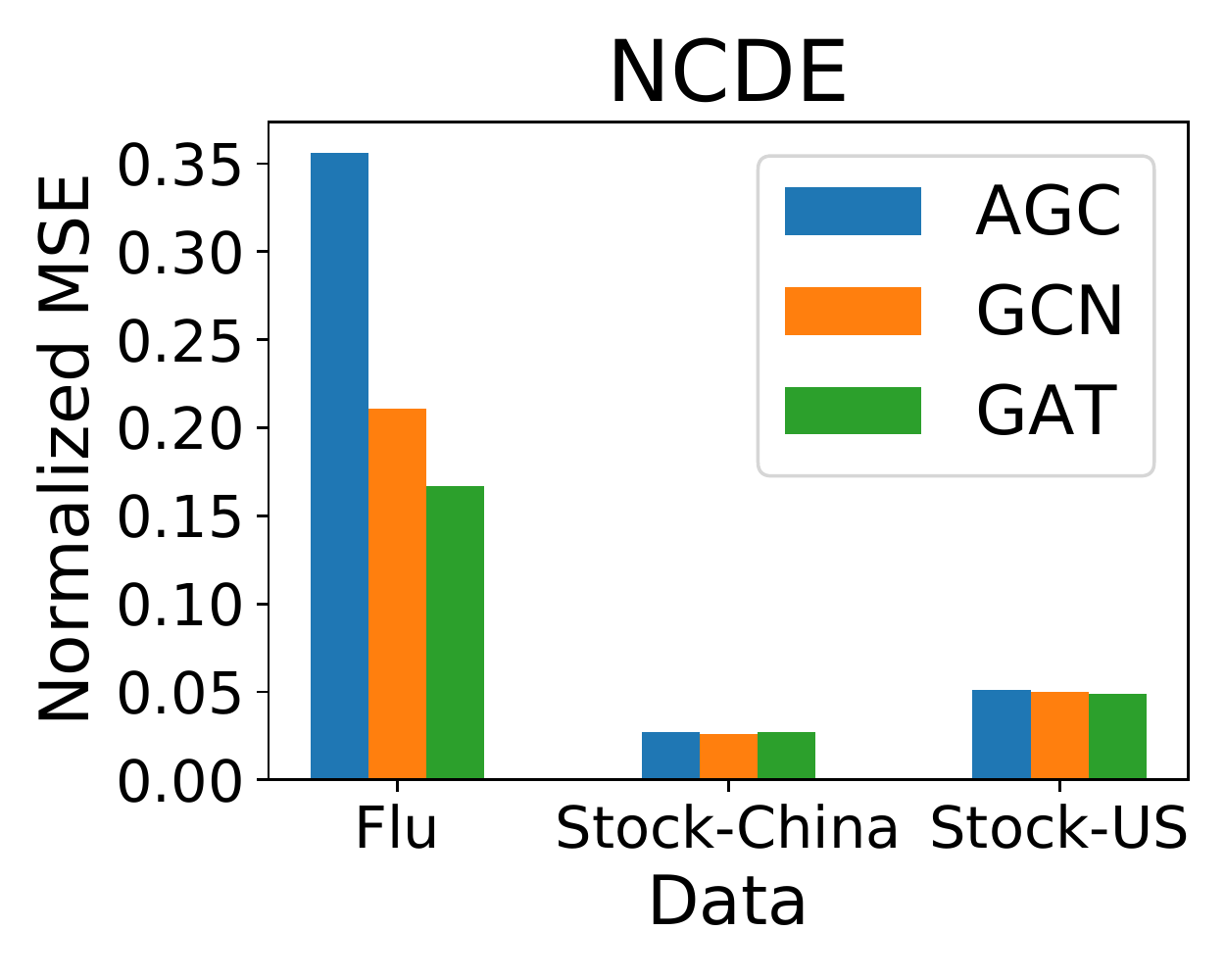}
    \includegraphics[width=0.49\columnwidth]{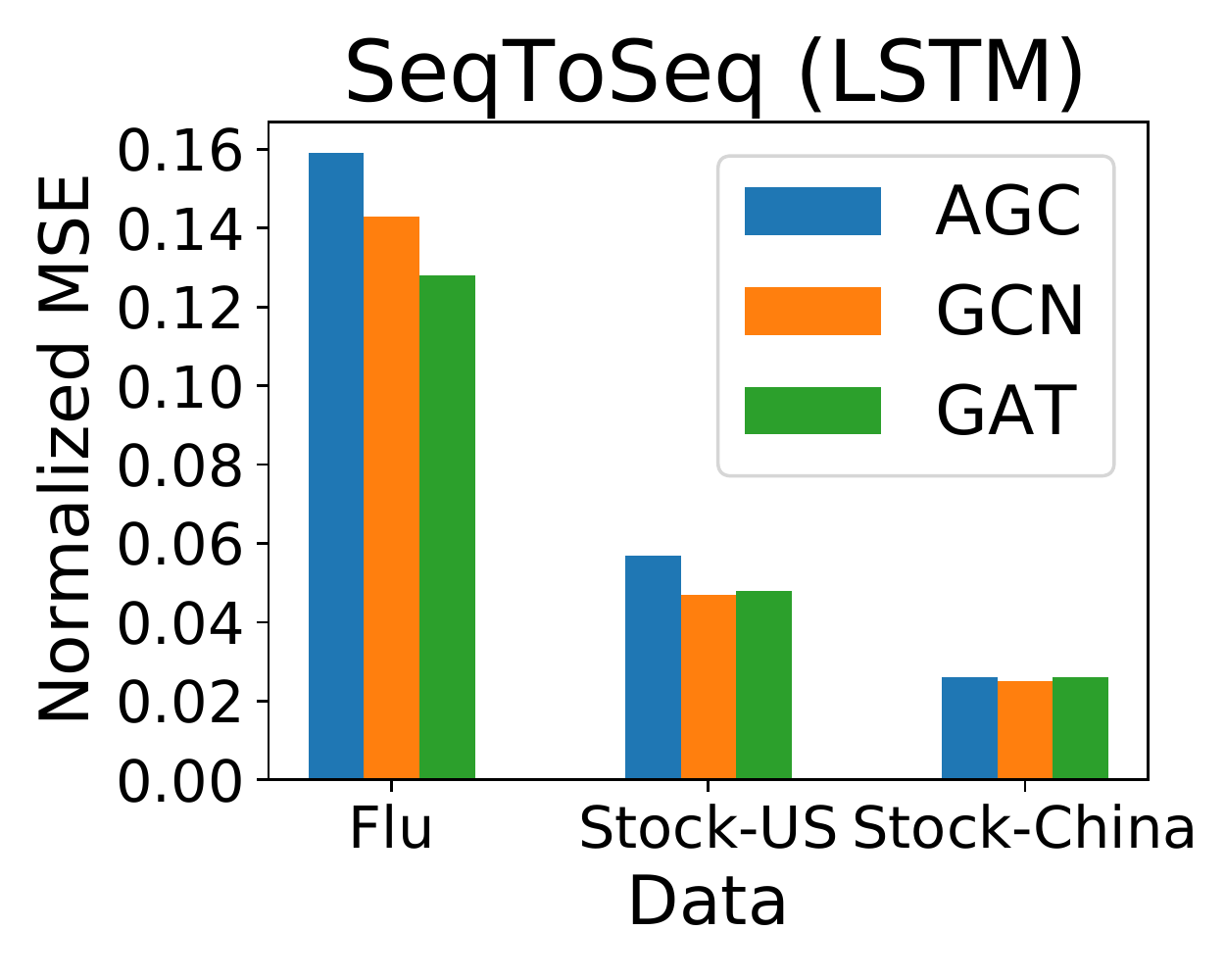}
    \caption{Changing the type of graph functions on the parameter generating layer ($L2$)}
    \label{fig:sens_graph}
\end{minipage}
\end{figure}


\noindent \textbf{Period size $K$.}  Fig.~\ref{fig:sens2} shows sensitivity analyses w.r.t. the input period size $K$. As shown, $K \leq 3$ produces the best outcomes. Feeding too much information leads to sub-optimal outcomes. For this reason, we use 2 or 3 as $K$.

\noindent \textbf{The type of graph function in $L2$.} Fig.~\ref{fig:sens_graph} shows the results by varying the graph function of the parameter generating layer ($L2$) in Eq.~\eqref{eq:gat}. In addition to our choice of GAT, we test with a graph convolutional network (GCN)~\citep{kipf2017semisupervised} and an adaptive graph convolutional network (AGC)~\citep{bai2020adaptive}. In most cases, GAT and GCN show reasonable performance. On the other hand, AGC, which does not use the computation graph but learns its own virtual graph, sometimes shows the worst performance --- note that, however, AGC is effective for the shared multi-task layer.

\section{REPRODUCIBILITY}\label{appendix:reproducibility}

We implement \textit{HyperGPA} with Python v3.7 and PyTorch v1.8 and run our experiments on a machine equipped with Nvidia RTX Titan or 3090. 
\subsection{Datasets}\label{appendix:data}
\texttt{Flu} 
contains the number of flu patients, total patients, and providers who give information about patients, in each of the 51 U.S. states, collected weekly by the Centers for Disease Control and Prevention~\citep{CDC}, i.e., $\dim(\mathbf{x})=3$.
The data collection period is 2011--2020.
%
In \textit{HyperGPA}, we set the length of one window $\mathcal{D}_{i,j}$ to a year; the length is 52 (or 53) weeks.
%
For \texttt{Flu}, each target time series model $\mathcal{B}_i$ reads recent $s_{in}=10$ observations and forecasts next $s_{out}=2$ observations.
%
%

We also use the daily historical stock prices of the U.S. and China (\texttt{Stock-US} and \texttt{Stock-China}) collected by~\citep{STK} over 20 months. 
Each dataset contains the opening, closing, highest, and lowest stock prices per day, i.e., $\dim(\mathbf{x})=4$.
We choose the top-30 companies with the highest market capitalization. 
For those two datasets, we set the window length to two months.
%
Note that the stock prices for those companies are interconnected, i.e., they are loosely-coupled data.
$s_{in}$ and $s_{out}$ for stock datasets are 10 and 4, respectively.
%

\texttt{USHCN} is a climate dataset that contains monthly average, minimum, and maximum temperature and precipitation in the U.S. states from the U.S. historical climatology network~\citep{USHCN}, i.e., $\dim(\mathbf{x})=4$. The collection period is 1981--2012. The length of one window is a year, 12 months. We set $s_{in}$ and $s_{out}$ to 10 and 2.


After making the input and output pairs of target models like Step 2 in Sec.~\ref{sub:propose34}, \texttt{Flu} has 20K training pairs, 3K validation pairs, and 3K test pairs. Both \texttt{Stock-US} and \texttt{Stock-China} have 10K training pairs, 1K validation pairs, and 1K test pairs. \texttt{USHCN} has 17K training pairs, 0.5K validation pairs, and 0.5K test pairs.

\subsection{Hyperparameters}\label{appendix:hyperparam}
We note that the hyperparameters of the target models (the number of layer, $n_\mathcal{B}$ and hidden size of target model, $\dim(\mathbf{h}_\mathcal{B})$) are already described in the main paper. In this section, we show the best hyperparameters for our baselines for reproducibility in Tables~\ref{tbl:hypergpahyperparam} to~\ref{tbl:arimahyperparam}.

\subsubsection{Baselines}
For our baselines, we set the learning rate to $10^{-3}$ and the coefficient of $L_2$ regularization term to $10^{-5}$. The size of mini-batch is 256. 
For \textit{Vanilla}, there is no additional hyperparameter. 
For \textit{HyperLSTM/GRU}, there are 2 additional hyperparameters, $\dim(\hat{\mathbf{h}})$ and $\dim(\mathbf{a})$, that are the dimensionality of $\hat{\mathbf{h}}$ and $\mathbf{a}$, defined in~\citep{ha2016hypernetworks} or Appendix~\ref{appendix:hypergru}.  As $\dim(\mathbf{a}) < \dim(\hat{\mathbf{h}}) < \dim(\mathbf{h}_\mathcal{B})$ recommended in~\citep{ha2016hypernetworks}, $\dim(\hat{\mathbf{h}})$ is in $\{32,16,8\}$, and $\dim(\mathbf{a})$ is in $\{16,8,4\}$. For \textit{ANODEV2}, we configure the baseline as in~\citep{zhang2019anodev2} and \textit{RevIN} is just adding an adaptation layer, so additional hyperparameters are not needed. \textit{AdaGRU/LSTM} has an additional hyperparameter, $\lambda$ which is the coefficient of a drift-related loss. In an \textit{ARIMA} case, there are 3 hyperparameters. $p,d,$ and $q$ denote the autoregressive, differences, and moving average components, respectively. $p$ is in \{1,2,3\}, $d$ in \{0,1,2\}, $q$ in \{1,2,3\}. 

\subsubsection{HyperGPA}
For \textit{HyperGPA}, the learning rate is set to $10^{-2}$, the coefficient of $L_2$ regularization term to $10^{-6}$, and the size of mini-batch to $10^{4}$. The reason why the size of the mini-batch for \textit{HyperGPA} is much larger than that for the baselines is that i) \textit{HyperGPA} generates the parameters of all target models, and ii) all target models are trained simultaneously. Because there are about 30 to 50 target models, the size of mini-batch for one target model is about 256. Each of $\dim(\mathbf{h}_\mathcal{B})$ and $n_\mathcal{B}$ in \textit{HyperGPA} is set to 16 and 1, respectively.

The hidden size and the number of layers for $\Gamma$ in Eq.~\ref{eq:gamma} are set to 32 and 2, respectively. The hidden size of NCDEs, $\dim(\mathbf{h'})$ in Eq.~\eqref{eq:agc}, is in $\{32,64,128\}$. In AGC function, the node embedding size and the output size are set to 32. $\Phi$ is a one-layer fully connected layer without bias. The size of initial query vectors in Eq.~\ref{eq:initquery}, $\dim(\mathbf{z})$, is in $\{512, 1024, 2048\}$. In Eq.~\eqref{eq:gat}, $\text{GAT}$ has three hyperparameters: i) number of heads, ii) depth, and iii) hidden size. The number of heads is set to 4, the depth to 3, and the hidden size to 128. In an attention-based generation method, the number of candidates model $C$ is in $\{3, 5, 10, 20, 48\}$, and the regularization coefficient of $MSE_2$, $\lambda$, is in $\{0, 0.1, 0.01\}$.
For input window size $K$, we use $K=2$ in \texttt{Flu}, \texttt{Stock-China}, and \texttt{Stock-China}, and $K=3$ in \texttt{USHCN}.

\begin{table}
\begin{minipage}[b]{0.58\linewidth}
\scriptsize
\setlength{\tabcolsep}{2pt}
\centering
\caption{Hyperparameters for \textit{HyperGPA}}\label{tbl:hypergpahyperparam}
\renewcommand{\arraystretch}{1.0}
\begin{tabular}{|cc|cccc|}
\hline
Data & \makecell{Target Model}  & $\dim(\mathbf{h'})$ & $\dim(\mathbf{z})$ & $C$ &  $\lambda$ \\ \specialrule{1pt}{1pt}{1pt} 

\multirow{6}{*}{\rotatebox[origin=c]{0}{\texttt{Flu}}} & \texttt{LSTM} & 128 & 2048 & 3 & 0.1 \\ 
& \texttt{GRU} & 128 & 1024 & 3 & 0.01 \\ 
& \makecell{\texttt{SeqToSeq(LSTM)}} & 128 & 1024 & 10 & 0.1 \\ 
& \makecell{\texttt{SeqToSeq(GRU)}} & 128 & 2048 & 3 & 0.01 \\ 
& \texttt{ODERNN} & 128 & 2048 & 10 & 0.0 \\ 
& \texttt{NCDE} & 128 & 512 & 3 & 0.1 \\ \hline 
\multirow{6}{*}{\rotatebox[origin=c]{0}{\texttt{Stock-US}}} & \texttt{LSTM} & 128 & 512 & 3 & 0.0 \\ 
& \texttt{GRU} & 128 & 512 & 10 & 0.01 \\ 
& \makecell{\texttt{SeqToSeq(LSTM)}} & 128 & 512 & 3 & 0.1 \\ 
& \makecell{\texttt{SeqToSeq(GRU)}} & 128 & 1024 & 5 & 0.0 \\ 
& \texttt{ODERNN} & 128 & 512 & 10 & 0.01 \\ 
& \texttt{NCDE} & 128 & 512 & 10 & 0.0 \\ \hline 
\multirow{6}{*}{\rotatebox[origin=c]{0}{\texttt{Stock-China}}} & \texttt{LSTM} & 128 & 512 & 3 & 0.1 \\ 
& \texttt{GRU} & 128 & 512 & 10 & 0.0 \\ 
& \makecell{\texttt{SeqToSeq(LSTM)}} & 128 & 2048 & 5 & 0.1 \\ 
& \makecell{\texttt{SeqToSeq(GRU)}} & 128 & 512 & 3 & 0.01 \\ 
& \texttt{ODERNN} & 128 & 1024 & 3 & 0.0 \\ 
& \texttt{NCDE} & 128 & 1024 & 5 & 0.0 \\ \hline 
\multirow{6}{*}{\rotatebox[origin=c]{0}{\texttt{USHCN}}} & \texttt{LSTM} & 128 & 512 & 48 & 0.01 \\ 
& \texttt{GRU} & 32 & 512 & 48 & 0.1 \\ 
& \makecell{\texttt{SeqToSeq(LSTM)}} & 32 & 512 & 48 & 0.01 \\ 
& \makecell{\texttt{SeqToSeq(GRU)}} & 128 & 512 & 48 & 0.1 \\ 
& \texttt{ODERNN} & 64 & 512 & 48 & 0.1 \\ 
& \texttt{NCDE} & 32 & 512 & 48 & 0.01 \\ \hline 

\end{tabular}
\end{minipage}
\hfill
\begin{minipage}[b]{0.38\linewidth}
\scriptsize
\setlength{\tabcolsep}{2pt}
\centering
\caption{Hyperparameters for \textit{Vanilla}}\label{tbl:wohyperhyperparam}
\renewcommand{\arraystretch}{1.0}
\begin{tabular}{|cc|cc|}
\hline
Data & Target Model & $\dim(\mathbf{h}_\mathcal{B})$ & $n_\mathcal{B}$ \\ \specialrule{1pt}{1pt}{1pt} 

\multirow{6}{*}{\rotatebox[origin=c]{0}{\texttt{Flu}}} & \texttt{LSTM} & 64 & 1 \\ 
& \texttt{GRU} & 64 & 1 \\ 
& \makecell{\texttt{SeqToSeq} \texttt{(LSTM)}} & 64 & 1 \\ 
& \makecell{\texttt{SeqToSeq} \texttt{(GRU)}} & 64 & 1 \\ 
& \texttt{ODERNN} & 32 & 1 \\ 
& \texttt{NCDE} & 32 & 3 \\ \hline 
\multirow{6}{*}{\rotatebox[origin=c]{0}{\texttt{Stock-US}}} & \texttt{LSTM} & 64 & 1 \\ 
& \texttt{GRU} & 64 & 1 \\ 
& \makecell{\texttt{SeqToSeq} \texttt{(LSTM)}} & 64 & 1 \\ 
& \makecell{\texttt{SeqToSeq} \texttt{(GRU)}} & 64 & 1 \\ 
& \texttt{ODERNN} & 64 & 3 \\ 
& \texttt{NCDE} & 16 & 2 \\ \hline 
\multirow{6}{*}{\rotatebox[origin=c]{0}{\texttt{Stock-China}}} & \texttt{LSTM} & 64 & 1 \\ 
& \texttt{GRU} & 16 & 1 \\ 
& \makecell{\texttt{SeqToSeq} \texttt{(LSTM)}} & 64 & 1 \\ 
& \makecell{\texttt{SeqToSeq} \texttt{(GRU)}} & 64 & 1 \\ 
& \texttt{ODERNN} & 32 & 1 \\ 
& \texttt{NCDE} & 64 & 2 \\ \hline 
\multirow{6}{*}{\rotatebox[origin=c]{0}{\texttt{USHCN}}} & \texttt{LSTM} & 32 & 1 \\ 
& \texttt{GRU} & 16 & 1 \\ 
& \makecell{\texttt{SeqToSeq} \texttt{(LSTM)}} & 16 & 1 \\ 
& \makecell{\texttt{SeqToSeq} \texttt{(GRU)}} & 32 & 1 \\ 
& \texttt{ODERNN} & 64 & 2 \\ 
& \texttt{NCDE} & 32 & 3 \\ \hline 

\end{tabular}
\end{minipage}
\end{table}

\begin{table}
\begin{minipage}[b]{0.58\linewidth}
\scriptsize
\setlength{\tabcolsep}{2pt}
\centering
\caption{Hyperparameters for \textit{HyperLSTM/GRU}}\label{tbl:davidhahyperparam}
\renewcommand{\arraystretch}{1.0}
\begin{tabular}{|cc|cccc|}
\hline
Data & Target Model & $\dim(\mathbf{h}_\mathcal{B})$ & $n_\mathcal{B}$ & $\dim(\mathbf{a})$ & $\dim(\hat{\mathbf{h}})$ \\ \specialrule{1pt}{1pt}{1pt} 

\multirow{4}{*}{\rotatebox[origin=c]{0}{\texttt{Flu}}} & \texttt{LSTM} & 64 & 1 & 16 & 8 \\ 
& \texttt{GRU} & 32 & 1 & 8 & 8 \\ 
& \makecell{\texttt{SeqToSeq} \texttt{(LSTM)}} & 64 & 1 & 4 & 8 \\ 
& \makecell{\texttt{SeqToSeq} \texttt{(GRU)}} & 16 & 1 & 4 & 16 \\ \hline 
\multirow{4}{*}{\rotatebox[origin=c]{0}{\texttt{Stock-US}}} & \texttt{LSTM} & 32 & 1 & 4 & 16 \\ 
& \texttt{GRU} & 16 & 1 & 16 & 8 \\ 
& \makecell{\texttt{SeqToSeq} \texttt{(LSTM)}} & 32 & 1 & 8 & 8 \\ 
& \makecell{\texttt{SeqToSeq} \texttt{(GRU)}} & 64 & 1 & 4 & 8 \\ \hline 
\multirow{4}{*}{\rotatebox[origin=c]{0}{\texttt{Stock-China}}} & \texttt{LSTM} & 16 & 1 & 4 & 8 \\ 
& \texttt{GRU} & 16 & 1 & 4 & 8 \\ 
& \makecell{\texttt{SeqToSeq} \texttt{(LSTM)}} & 16 & 1 & 4 & 8 \\ 
& \makecell{\texttt{SeqToSeq} \texttt{(GRU)}} & 32 & 1 & 4 & 8 \\ \hline 
\multirow{4}{*}{\rotatebox[origin=c]{0}{\texttt{USHCN}}} & \texttt{LSTM} & 32 & 1 & 4 & 16 \\ 
& \texttt{GRU} & 64 & 1 & 4 & 8 \\ 
& \makecell{\texttt{SeqToSeq} \texttt{(LSTM)}} & 64 & 1 & 4 & 32 \\ 
& \makecell{\texttt{SeqToSeq} \texttt{(GRU)}} & 64 & 1 & 4 & 8 \\ \hline 

\end{tabular}

\end{minipage}
\hfill
\begin{minipage}[b]{0.38\linewidth}
\scriptsize
\setlength{\tabcolsep}{2pt}
\centering
\caption{Hyperparameters for \textit{RevIN}}\label{tbl:revinhyperparam}
\renewcommand{\arraystretch}{1.0}
\begin{tabular}{|cc|cc|}
\hline
 Data & Target Model & $\dim(\mathbf{h}_\mathcal{B})$ & $n_\mathcal{B}$ \\ \specialrule{1pt}{1pt}{1pt} 
\multirow{6}{*}{\rotatebox[origin=c]{0}{\texttt{Flu}}} & \texttt{LSTM} & 16 & 3 \\ 
& \texttt{GRU} & 16 & 2 \\ 
& \makecell{\texttt{SeqToSeq} \texttt{(LSTM)}} & 16 & 1 \\ 
& \makecell{\texttt{SeqToSeq} \texttt{(GRU)}} & 32 & 1 \\ 
& \texttt{ODERNN} & 32 & 3 \\ 
& \texttt{NCDE} & 64 & 3 \\ \hline 
\multirow{6}{*}{\rotatebox[origin=c]{0}{\texttt{Stock-US}}} & \texttt{LSTM} & 16 & 1 \\ 
& \texttt{GRU} & 16 & 1 \\ 
& \makecell{\texttt{SeqToSeq} \texttt{(LSTM)}} & 16 & 1 \\ 
& \makecell{\texttt{SeqToSeq} \texttt{(GRU)}} & 16 & 1 \\ 
& \texttt{ODERNN} & 32 & 2 \\ 
& \texttt{NCDE} & 16 & 3 \\ \hline 
\multirow{6}{*}{\rotatebox[origin=c]{0}{\texttt{Stock-China}}} & \texttt{LSTM} & 32 & 1 \\ 
& \texttt{GRU} & 16 & 2 \\ 
& \makecell{\texttt{SeqToSeq} \texttt{(LSTM)}} & 16 & 1 \\ 
& \makecell{\texttt{SeqToSeq} \texttt{(GRU)}} & 16 & 1 \\ 
& \texttt{ODERNN} & 64 & 3 \\ 
& \texttt{NCDE} & 32 & 3 \\ \hline 
\multirow{6}{*}{\rotatebox[origin=c]{0}{\texttt{USHCN}}} & \texttt{LSTM} & 32 & 1 \\ 
& \texttt{GRU} & 16 & 3 \\ 
& \makecell{\texttt{SeqToSeq} \texttt{(LSTM)}} & 64 & 1 \\ 
& \makecell{\texttt{SeqToSeq} \texttt{(GRU)}} & 32 & 3 \\ 
& \texttt{ODERNN} & 64 & 2 \\ 
& \texttt{NCDE} & 64 & 3 \\ \hline 
\end{tabular}
\end{minipage}
\end{table}

\begin{table}[t]
\begin{minipage}[b]{0.48\linewidth}
\scriptsize
\setlength{\tabcolsep}{2pt}
\centering
\caption{Hyperparameters for \textit{ANODEV2}}\label{tbl:anodevhyperparam}
\renewcommand{\arraystretch}{1.0}
\begin{tabular}{|cc|cc|}
\hline
 Data & Target Model & $\dim(\mathbf{h}_\mathcal{B})$ & $n_\mathcal{B}$ \\ \specialrule{1pt}{1pt}{1pt} 
\multirow{2}{*}{\rotatebox[origin=c]{0}{\texttt{Flu}}} & \texttt{ODERNN} & 64 & 3 \\ 
& \texttt{NCDE} & 16 & 3 \\ \hline 
\multirow{2}{*}{\rotatebox[origin=c]{0}{\texttt{Stock-US}}} & \texttt{ODERNN} & 64 & 3 \\ 
& \texttt{NCDE} & 64 & 3 \\ \hline 
\multirow{2}{*}{\rotatebox[origin=c]{0}{\texttt{Stock-China}}} & \texttt{ODERNN} & 64 & 2 \\ 
& \texttt{NCDE} & 16 & 2 \\ \hline 
\multirow{2}{*}{\rotatebox[origin=c]{0}{\texttt{USHCN}}} & \texttt{ODERNN} & 64 & 3 \\ 
& \texttt{NCDE} & 32 & 3 \\ \hline 
\end{tabular}
\end{minipage}
\hfill
\begin{minipage}[b]{0.48\linewidth}
\scriptsize
\setlength{\tabcolsep}{2pt}
\centering
\caption{Hyperparameters for \textit{AdaLSTM/GRU}}\label{tbl:adarnnhyperparam}
\renewcommand{\arraystretch}{1.0}
\begin{tabular}{|cc|ccc|}
\hline
 Data & Target Model & $\dim(\mathbf{h}_\mathcal{B})$ & $n_\mathcal{B}$ & $\lambda$ \\ \specialrule{1pt}{1pt}{1pt} 
\multirow{2}{*}{\rotatebox[origin=c]{0}{\texttt{Flu}}} & \texttt{LSTM} & 64 & 1 & 1.0 \\ 
& \texttt{GRU} & 64 & 1 & 1.0 \\ \hline 
\multirow{2}{*}{\rotatebox[origin=c]{0}{\texttt{Stock-US}}} & \texttt{LSTM} & 64 & 1 & 1.0 \\ 
& \texttt{GRU} & 64 & 1 & 1.0 \\ \hline 
\multirow{2}{*}{\rotatebox[origin=c]{0}{\texttt{Stock-China}}} & \texttt{LSTM} & 64 & 1 & 1.0 \\
& \texttt{GRU} & 64 & 1 & 0.5 \\ \hline 
\multirow{2}{*}{\rotatebox[origin=c]{0}{\texttt{USHCN}}} & \texttt{LSTM} & 64 & 1 & 1.0 \\ 
& \texttt{GRU} & 32 & 1 & 1.0 \\ \hline 
\end{tabular}
\end{minipage}

\vspace{1em}

\begin{minipage}{\linewidth}
\scriptsize
\setlength{\tabcolsep}{2pt}
\centering
\caption{Hyperparameters for \textit{ARIMA}}\label{tbl:arimahyperparam}
\renewcommand{\arraystretch}{1.0}
\begin{tabular}{|c|ccc|}
\hline
 Data & $p$ & $d$ & $q$ \\ \specialrule{1pt}{1pt}{1pt} 
\multirow{1}{*}{\rotatebox[origin=c]{0}{\texttt{Flu}}} &  2 & 1 & 1 \\ \hline
\multirow{1}{*}{\rotatebox[origin=c]{0}{\texttt{Stock-US}}} & 1 & 1 & 2 \\ \hline
\multirow{1}{*}{\rotatebox[origin=c]{0}{\texttt{Stock-China}}} & 1 & 1 & 1 \\ \hline 
\multirow{1}{*}{\rotatebox[origin=c]{0}{\texttt{USHCN}}} & 1 & 1 & 2 \\ \hline
\end{tabular}
\end{minipage}

\end{table}

\section{FULL EXPERIMENTAL RESULTS}\label{appendix:detailedresult}
Full experimental results are shown from Table.~\ref{tbl:flufull} to Table.~\ref{tbl:ushcnfull}. Val.MSE is the MSE value in validation data, which is the criterion for selecting the best models. 
\begin{table*}[t]
\scriptsize
\setlength{\tabcolsep}{3pt}
\centering
\caption{Experimental results for \texttt{Flu}}\label{tbl:flufull}
\renewcommand{\arraystretch}{1.2}
\begin{tabular}{|cc|cccccc|}

\hline
Target Model & Generation Way & Val.MSE & PCC & $R^2$ & Exp. & MSE & MAE  \\ \specialrule{1pt}{1pt}{1pt} 
\multicolumn{2}{|c|}{\textit{ARIMA}} & 0.502±0.000 & 0.695±0.000 & 0.152±0.000 & 0.168±0.000 & 1.091±0.000 & 0.579±0.000 \\ \hline
\multirow{5}{*}{\texttt{LSTM}} & \textit{Vanilla} & 0.174±0.004 & 0.910±0.003 & 0.718±0.023 & 0.730±0.021 & 0.367±0.016 & 0.299±0.008\\
& \textit{HyperLSTM} & 0.236±0.008 & 0.852±0.005 & 0.434±0.038 & 0.471±0.032 & 0.582±0.019 & 0.388±0.010\\
& \textit{RevIN} & 0.230±0.022 & 0.917±0.010 & 0.842±0.017 & 0.844±0.018 & 0.506±0.097 & 0.256±0.007\\
& \textit{AdaLSTM} & 0.516±0.047 & 0.814±0.019 & 0.367±0.099 & 0.478±0.075 & 0.740±0.075 & 0.552±0.036\\
& \textit{HyperGPA} & 0.068±0.002 & 0.971±0.001 & 0.933±0.003 & 0.933±0.004 & 0.118±0.004 & 0.141±0.002\\ \hline
\multirow{5}{*}{\texttt{GRU}} & \textit{Vanilla} & 0.130±0.003 & 0.933±0.002 & 0.807±0.006 & 0.813±0.006 & 0.275±0.006 & 0.250±0.003\\
& \textit{HyperGRU} & 0.207±0.016 & 0.863±0.010 & 0.585±0.035 & 0.601±0.035 & 0.520±0.033 & 0.361±0.011\\
& \textit{RevIN} & 0.199±0.015 & 0.938±0.004 & 0.880±0.007 & 0.881±0.008 & 0.379±0.039 & 0.226±0.007\\
& \textit{AdaGRU} & 0.405±0.031 & 0.849±0.009 & 0.542±0.048 & 0.613±0.056 & 0.616±0.035 & 0.510±0.018\\
& \textit{HyperGPA} & 0.066±0.002 & 0.971±0.001 & 0.938±0.003 & 0.939±0.003 & 0.116±0.004 & 0.140±0.003\\ \hline
\multirow{4}{*}{\makecell{\texttt{SeqToSeq(LSTM)}}} & \textit{Vanilla} & 0.151±0.003 & 0.914±0.002 & 0.733±0.004 & 0.747±0.004 & 0.353±0.005 & 0.290±0.004\\
& \textit{HyperLSTM} & 0.211±0.015 & 0.855±0.007 & 0.502±0.035 & 0.527±0.032 & 0.559±0.021 & 0.372±0.007\\
& \textit{RevIN} & 0.193±0.007 & 0.944±0.004 & 0.888±0.006 & 0.890±0.006 & 0.345±0.021 & 0.219±0.005\\
& \textit{HyperGPA} & 0.070±0.004 & 0.969±0.001 & 0.932±0.006 & 0.932±0.006 & 0.128±0.006 & 0.143±0.004\\ \hline
\multirow{4}{*}{\makecell{\texttt{SeqToSeq(GRU)}}} & \textit{Vanilla} & 0.114±0.004 & 0.939±0.001 & 0.828±0.006 & 0.834±0.005 & 0.250±0.005 & 0.237±0.004\\
& \textit{HyperGRU} & 0.191±0.008 & 0.872±0.007 & 0.574±0.028 & 0.592±0.028 & 0.502±0.023 & 0.347±0.009\\
& \textit{RevIN} & 0.189±0.010 & 0.954±0.003 & 0.905±0.006 & 0.908±0.006 & 0.291±0.032 & 0.204±0.005\\
& \textit{HyperGPA} & 0.066±0.002 & 0.968±0.003 & 0.926±0.012 & 0.926±0.012 & 0.130±0.014 & 0.142±0.004\\ \hline
\multirow{4}{*}{\texttt{ODERNN}} & \textit{Vanilla} & 0.163±0.011 & 0.907±0.010 & 0.761±0.030 & 0.767±0.027 & 0.361±0.039 & 0.301±0.017\\
& \textit{ANODEV2} & 0.139±0.003 & 0.926±0.004 & 0.790±0.007 & 0.796±0.007 & 0.298±0.014 & 0.256±0.009\\
& \textit{RevIN} & 0.272±0.004 & 0.905±0.008 & 0.817±0.011 & 0.820±0.011 & 0.549±0.226 & 0.266±0.014\\
& \textit{HyperGPA} & 0.075±0.003 & 0.967±0.004 & 0.931±0.007 & 0.932±0.007 & 0.134±0.016 & 0.145±0.004\\ \hline
\multirow{4}{*}{\texttt{NCDE}} & \textit{Vanilla} & 0.137±0.019 & 0.900±0.011 & 0.775±0.015 & 0.781±0.016 & 0.387±0.042 & 0.296±0.011\\
& \textit{ANODEV2} & 0.457±0.002 & 0.778±0.004 & 0.041±0.017 & 0.077±0.018 & 0.821±0.011 & 0.480±0.005\\
& \textit{RevIN} & 0.259±0.012 & 0.919±0.003 & 0.843±0.005 & 0.848±0.005 & 0.439±0.022 & 0.251±0.004\\
& \textit{HyperGPA} & 0.071±0.003 & 0.959±0.004 & 0.913±0.009 & 0.914±0.009 & 0.167±0.020 & 0.164±0.008\\ \hline

\end{tabular}
\end{table*}

\begin{table*}[t]
\scriptsize
\setlength{\tabcolsep}{3pt}
\centering
\caption{Experimental results for \texttt{Stock-US}}\label{tbl:stockusfull}
\renewcommand{\arraystretch}{1.2}
\begin{tabular}{|cc|cccccc|}

\hline
Target Model & Generation Way & Val.MSE & PCC & $R^2$ & Exp. & MSE & MAE \\ \specialrule{1pt}{1pt}{1pt} 
\multicolumn{2}{|c|}{\textit{ARIMA}} & 0.205±0.000 & 0.548±0.000 & -0.781±0.000 & -0.748±0.000 & 0.620±0.000 & 0.609±0.000 \\ \hline
\multirow{5}{*}{\texttt{LSTM}} & \textit{Vanilla} & 0.051±0.001 & 0.902±0.005 & 0.569±0.029 & 0.641±0.022 & 0.213±0.010 & 0.307±0.006\\
& \textit{HyperLSTM} & 0.089±0.005 & 0.605±0.071 & -1.045±0.340 & -0.491±0.248 & 0.751±0.103 & 0.586±0.038\\
& \textit{RevIN} & 0.045±0.000 & 0.966±0.000 & 0.930±0.000 & 0.933±0.000 & 0.063±0.001 & 0.183±0.001\\
& \textit{AdaLSTM} & 0.257±0.017 & 0.787±0.068 & 0.400±0.231 & 0.475±0.209 & 0.379±0.102 & 0.483±0.062\\
& \textit{HyperGPA} & 0.034±0.000 & 0.973±0.000 & 0.930±0.005 & 0.933±0.004 & 0.050±0.002 & 0.161±0.004\\ \hline
\multirow{5}{*}{\texttt{GRU}} & \textit{Vanilla} & 0.043±0.001 & 0.952±0.002 & 0.830±0.007 & 0.848±0.006 & 0.102±0.003 & 0.226±0.004\\
& \textit{HyperGRU} & 0.073±0.007 & 0.772±0.030 & -0.167±0.113 & 0.103±0.086 & 0.462±0.046 & 0.453±0.015\\
& \textit{RevIN} & 0.044±0.000 & 0.967±0.000 & 0.933±0.001 & 0.935±0.000 & 0.060±0.001 & 0.178±0.000\\
& \textit{AdaGRU} & 0.176±0.021 & 0.875±0.028 & 0.645±0.073 & 0.693±0.073 & 0.233±0.043 & 0.377±0.045\\
& \textit{HyperGPA} & 0.034±0.000 & 0.972±0.001 & 0.927±0.005 & 0.930±0.004 & 0.052±0.002 & 0.164±0.004\\ \hline
\multirow{4}{*}{\makecell{\texttt{SeqToSeq(LSTM)}}} & \textit{Vanilla} & 0.042±0.001 & 0.926±0.003 & 0.688±0.014 & 0.744±0.011 & 0.167±0.005 & 0.274±0.004\\
& \textit{HyperLSTM} & 0.077±0.004 & 0.674±0.052 & -0.604±0.181 & -0.173±0.120 & 0.643±0.084 & 0.528±0.030\\
& \textit{RevIN} & 0.046±0.001 & 0.967±0.000 & 0.932±0.002 & 0.935±0.001 & 0.061±0.001 & 0.180±0.002\\
& \textit{HyperGPA} & 0.034±0.000 & 0.973±0.001 & 0.936±0.002 & 0.938±0.002 & 0.048±0.001 & 0.157±0.002\\ \hline
\multirow{4}{*}{\makecell{\texttt{SeqToSeq(GRU)}}} & \textit{Vanilla} & 0.039±0.000 & 0.948±0.003 & 0.811±0.012 & 0.835±0.010 & 0.112±0.006 & 0.229±0.003\\
& \textit{HyperGRU} & 0.067±0.002 & 0.769±0.057 & -0.121±0.235 & 0.141±0.175 & 0.464±0.092 & 0.444±0.036\\
& \textit{RevIN} & 0.044±0.001 & 0.967±0.001 & 0.933±0.002 & 0.935±0.002 & 0.060±0.002 & 0.177±0.002\\
& \textit{HyperGPA} & 0.034±0.000 & 0.973±0.001 & 0.935±0.007 & 0.938±0.005 & 0.049±0.003 & 0.159±0.006\\ \hline
\multirow{4}{*}{\texttt{ODERNN}} & \textit{Vanilla} & 0.053±0.002 & 0.894±0.014 & 0.659±0.028 & 0.697±0.023 & 0.200±0.015 & 0.300±0.008\\
& \textit{ANODEV2} & 0.043±0.001 & 0.940±0.009 & 0.801±0.024 & 0.822±0.023 & 0.120±0.013 & 0.236±0.008\\
& \textit{RevIN} & 0.048±0.001 & 0.964±0.000 & 0.926±0.001 & 0.929±0.001 & 0.068±0.001 & 0.188±0.001\\
& \textit{HyperGPA} & 0.034±0.000 & 0.972±0.000 & 0.931±0.001 & 0.933±0.001 & 0.050±0.001 & 0.162±0.002\\ \hline
\multirow{4}{*}{\texttt{NCDE}} & \textit{Vanilla} & 0.045±0.001 & 0.929±0.007 & 0.805±0.029 & 0.822±0.024 & 0.130±0.015 & 0.247±0.010\\
& \textit{ANODEV2} & 0.152±0.000 & 0.850±0.001 & 0.551±0.003 & 0.555±0.003 & 0.244±0.002 & 0.376±0.002\\
& \textit{RevIN} & 0.044±0.001 & 0.967±0.000 & 0.934±0.001 & 0.935±0.000 & 0.060±0.001 & 0.177±0.001\\
& \textit{HyperGPA} & 0.034±0.000 & 0.973±0.001 & 0.934±0.004 & 0.936±0.003 & 0.049±0.002 & 0.159±0.003\\ \hline

\end{tabular}
\end{table*}

\begin{table*}[t]
\scriptsize
\setlength{\tabcolsep}{3pt}
\centering
\caption{Experimental results for \texttt{Stock-China}}\label{tbl:stockchinafull}
\renewcommand{\arraystretch}{1.2}
\begin{tabular}{|cc|cccccc|}

\hline
Target Model & Generation Way & Val.MSE & PCC & $R^2$ & Exp. & MSE & MAE \\ \specialrule{1pt}{1pt}{1pt} 
\multicolumn{2}{|c|}{\textit{ARIMA}} & 0.478±0.000 & 0.840±0.000 & 0.570±0.000 & 0.682±0.000 & 0.439±0.000 & 0.503±0.000 \\ \hline
\multirow{5}{*}{\texttt{LSTM}} & \textit{Vanilla} & 0.084±0.002 & 0.975±0.000 & 0.945±0.001 & 0.946±0.001 & 0.050±0.001 & 0.160±0.002\\
& \textit{HyperLSTM} & 0.188±0.019 & 0.949±0.003 & 0.879±0.009 & 0.881±0.008 & 0.103±0.007 & 0.226±0.004\\
& \textit{RevIN} & 0.081±0.001 & 0.977±0.001 & 0.954±0.002 & 0.955±0.002 & 0.049±0.002 & 0.152±0.002\\
& \textit{AdaLSTM} & 0.359±0.068 & 0.834±0.037 & 0.565±0.183 & 0.571±0.180 & 0.321±0.064 & 0.460±0.039\\
& \textit{HyperGPA} & 0.059±0.001 & 0.987±0.000 & 0.972±0.001 & 0.972±0.001 & 0.026±0.001 & 0.113±0.002\\ \hline
\multirow{5}{*}{\texttt{GRU}} & \textit{Vanilla} & 0.089±0.003 & 0.982±0.001 & 0.959±0.002 & 0.960±0.002 & 0.037±0.002 & 0.138±0.003\\
& \textit{HyperGRU} & 0.156±0.021 & 0.963±0.002 & 0.915±0.003 & 0.917±0.003 & 0.075±0.002 & 0.194±0.004\\
& \textit{RevIN} & 0.081±0.000 & 0.981±0.000 & 0.962±0.000 & 0.962±0.000 & 0.040±0.000 & 0.143±0.001\\
& \textit{AdaGRU} & 0.199±0.015 & 0.913±0.009 & 0.783±0.032 & 0.786±0.032 & 0.170±0.017 & 0.327±0.023\\
& \textit{HyperGPA} & 0.058±0.002 & 0.987±0.000 & 0.972±0.001 & 0.972±0.001 & 0.026±0.001 & 0.114±0.003\\ \hline
\multirow{4}{*}{\makecell{\texttt{SeqToSeq(LSTM)}}} & \textit{Vanilla} & 0.084±0.002 & 0.978±0.001 & 0.951±0.002 & 0.952±0.002 & 0.045±0.001 & 0.151±0.002\\
& \textit{HyperLSTM} & 0.163±0.017 & 0.952±0.023 & 0.889±0.054 & 0.892±0.051 & 0.097±0.045 & 0.205±0.022\\
& \textit{RevIN} & 0.079±0.001 & 0.979±0.001 & 0.958±0.002 & 0.959±0.002 & 0.044±0.002 & 0.144±0.002\\
& \textit{HyperGPA} & 0.058±0.001 & 0.987±0.000 & 0.973±0.001 & 0.973±0.001 & 0.026±0.001 & 0.112±0.001\\ \hline
\multirow{4}{*}{\makecell{\texttt{SeqToSeq(GRU)}}} & \textit{Vanilla} & 0.072±0.002 & 0.983±0.000 & 0.963±0.001 & 0.964±0.001 & 0.035±0.001 & 0.131±0.002\\
& \textit{HyperGRU} & 0.150±0.018 & 0.964±0.004 & 0.920±0.009 & 0.922±0.008 & 0.073±0.008 & 0.187±0.007\\
& \textit{RevIN} & 0.078±0.001 & 0.981±0.001 & 0.962±0.002 & 0.963±0.002 & 0.039±0.003 & 0.138±0.003\\
& \textit{HyperGPA} & 0.058±0.001 & 0.988±0.000 & 0.973±0.001 & 0.973±0.001 & 0.025±0.000 & 0.112±0.001\\ \hline
\multirow{4}{*}{\texttt{ODERNN}} & \textit{Vanilla} & 0.092±0.013 & 0.972±0.004 & 0.938±0.009 & 0.939±0.009 & 0.056±0.007 & 0.171±0.013\\
& \textit{ANODEV2} & 0.079±0.002 & 0.982±0.001 & 0.961±0.001 & 0.962±0.001 & 0.037±0.001 & 0.139±0.002\\
& \textit{RevIN} & 0.086±0.001 & 0.978±0.001 & 0.955±0.002 & 0.956±0.002 & 0.048±0.002 & 0.154±0.002\\
& \textit{HyperGPA} & 0.058±0.001 & 0.987±0.000 & 0.972±0.001 & 0.972±0.001 & 0.026±0.001 & 0.115±0.003\\ \hline
\multirow{4}{*}{\texttt{NCDE}} & \textit{Vanilla} & 0.081±0.004 & 0.980±0.001 & 0.959±0.002 & 0.960±0.002 & 0.040±0.001 & 0.142±0.001\\
& \textit{ANODEV2} & 0.335±0.001 & 0.929±0.001 & 0.832±0.003 & 0.838±0.003 & 0.141±0.001 & 0.292±0.002\\
& \textit{RevIN} & 0.088±0.001 & 0.981±0.000 & 0.962±0.000 & 0.962±0.000 & 0.040±0.001 & 0.142±0.001\\
& \textit{HyperGPA} & 0.060±0.003 & 0.987±0.001 & 0.971±0.001 & 0.971±0.001 & 0.027±0.001 & 0.116±0.003\\ \hline

\end{tabular}
\end{table*}

\begin{table*}[t]
\scriptsize
\setlength{\tabcolsep}{3pt}
\centering
\caption{Experimental results for \texttt{USHCN}}\label{tbl:ushcnfull}
\renewcommand{\arraystretch}{1.2}
\begin{tabular}{|cc|cccccc|}

\hline
Target Model & Generation Way & Val.MSE & PCC & $R^2$ & Exp. & MSE & MAE \\ \specialrule{1pt}{1pt}{1pt} 
\multicolumn{2}{|c|}{\textit{ARIMA}} & 0.277±0.000 & 0.838±0.000 & 0.284±0.001 & 0.394±0.001 & 0.232±0.000 & 0.313±0.000 \\ \hline
\multirow{5}{*}{\texttt{LSTM}} & \textit{Vanilla} & 0.298±0.001 & 0.841±0.001 & 0.152±0.026 & 0.270±0.021 & 0.239±0.003 & 0.345±0.004\\
& \textit{HyperLSTM} & 0.320±0.003 & 0.824±0.003 & 0.075±0.009 & 0.191±0.009 & 0.249±0.004 & 0.350±0.003\\
& \textit{RevIN} & 0.779±0.029 & 0.693±0.009 & -0.019±0.054 & 0.225±0.035 & 0.589±0.015 & 0.610±0.010\\
& \textit{AdaLSTM} & 0.679±0.013 & 0.594±0.015 & -0.533±0.259 & -0.327±0.222 & 0.595±0.026 & 0.613±0.012\\
& \textit{HyperGPA} & 0.299±0.003 & 0.844±0.002 & 0.127±0.040 & 0.245±0.040 & 0.221±0.003 & 0.313±0.009\\ \hline
\multirow{5}{*}{\texttt{GRU}} & \textit{Vanilla} & 0.298±0.001 & 0.840±0.001 & 0.156±0.014 & 0.264±0.012 & 0.232±0.001 & 0.334±0.001\\
& \textit{HyperGRU} & 0.313±0.002 & 0.832±0.002 & 0.124±0.028 & 0.239±0.022 & 0.244±0.004 & 0.346±0.005\\
& \textit{RevIN} & 0.599±0.016 & 0.699±0.008 & 0.057±0.070 & 0.178±0.050 & 0.465±0.007 & 0.522±0.006\\
& \textit{AdaGRU} & 0.545±0.017 & 0.676±0.018 & -0.171±0.310 & -0.035±0.269 & 0.472±0.026 & 0.532±0.014\\
& \textit{HyperGPA} & 0.299±0.003 & 0.841±0.003 & 0.142±0.025 & 0.266±0.009 & 0.229±0.004 & 0.323±0.003\\ \hline
\multirow{4}{*}{\makecell{\texttt{SeqToSeq(LSTM)}}} & \textit{Vanilla} & 0.299±0.001 & 0.835±0.002 & 0.157±0.015 & 0.261±0.011 & 0.236±0.003 & 0.335±0.004\\
& \textit{HyperLSTM} & 0.316±0.002 & 0.826±0.003 & 0.141±0.013 & 0.244±0.014 & 0.243±0.004 & 0.343±0.004\\
& \textit{RevIN} & 0.730±0.023 & 0.682±0.016 & -0.059±0.075 & 0.143±0.052 & 0.585±0.014 & 0.600±0.009\\
& \textit{HyperGPA} & 0.296±0.005 & 0.843±0.002 & 0.242±0.043 & 0.344±0.043 & 0.220±0.007 & 0.307±0.006\\ \hline
\multirow{4}{*}{\makecell{\texttt{SeqToSeq(GRU)}}} & \textit{Vanilla} & 0.294±0.001 & 0.839±0.001 & 0.149±0.016 & 0.256±0.013 & 0.232±0.001 & 0.335±0.002\\
& \textit{HyperGRU} & 0.308±0.003 & 0.831±0.001 & 0.169±0.013 & 0.265±0.014 & 0.236±0.003 & 0.336±0.003\\
& \textit{RevIN} & 0.650±0.007 & 0.678±0.003 & -0.087±0.056 & 0.090±0.047 & 0.519±0.006 & 0.563±0.001\\
& \textit{HyperGPA} & 0.302±0.002 & 0.844±0.005 & 0.182±0.083 & 0.287±0.083 & 0.222±0.004 & 0.315±0.006\\ \hline
\multirow{4}{*}{\texttt{ODERNN}} & \textit{Vanilla} & 0.300±0.001 & 0.840±0.001 & 0.192±0.023 & 0.297±0.018 & 0.235±0.004 & 0.339±0.008\\
& \textit{ANODEV2} & 0.296±0.001 & 0.842±0.001 & 0.181±0.029 & 0.291±0.023 & 0.233±0.004 & 0.337±0.006\\
& \textit{RevIN} & 1.162±0.066 & 0.529±0.021 & -0.784±0.248 & -0.327±0.180 & 0.855±0.046 & 0.757±0.019\\
& \textit{HyperGPA} & 0.291±0.001 & 0.840±0.004 & 0.182±0.077 & 0.286±0.078 & 0.226±0.007 & 0.319±0.011\\ \hline
\multirow{4}{*}{\texttt{NCDE}} & \textit{Vanilla} & 0.304±0.002 & 0.829±0.002 & 0.188±0.038 & 0.294±0.031 & 0.234±0.004 & 0.322±0.002\\
& \textit{ANODEV2} & 0.754±0.001 & 0.616±0.001 & -2.257±0.065 & -1.924±0.056 & 0.483±0.002 & 0.554±0.001\\
& \textit{RevIN} & 1.052±0.014 & 0.467±0.015 & -0.920±0.086 & -0.724±0.087 & 0.713±0.008 & 0.687±0.006\\
& \textit{HyperGPA} & 0.305±0.008 & 0.837±0.003 & 0.134±0.062 & 0.246±0.073 & 0.227±0.004 & 0.316±0.006\\ \hline

\end{tabular}
\end{table*}

\newpage

\section{VISUALIZATION}\label{appendix:visual}
\begin{figure}[ht]
    \centering
    \includegraphics[width=0.32\columnwidth]{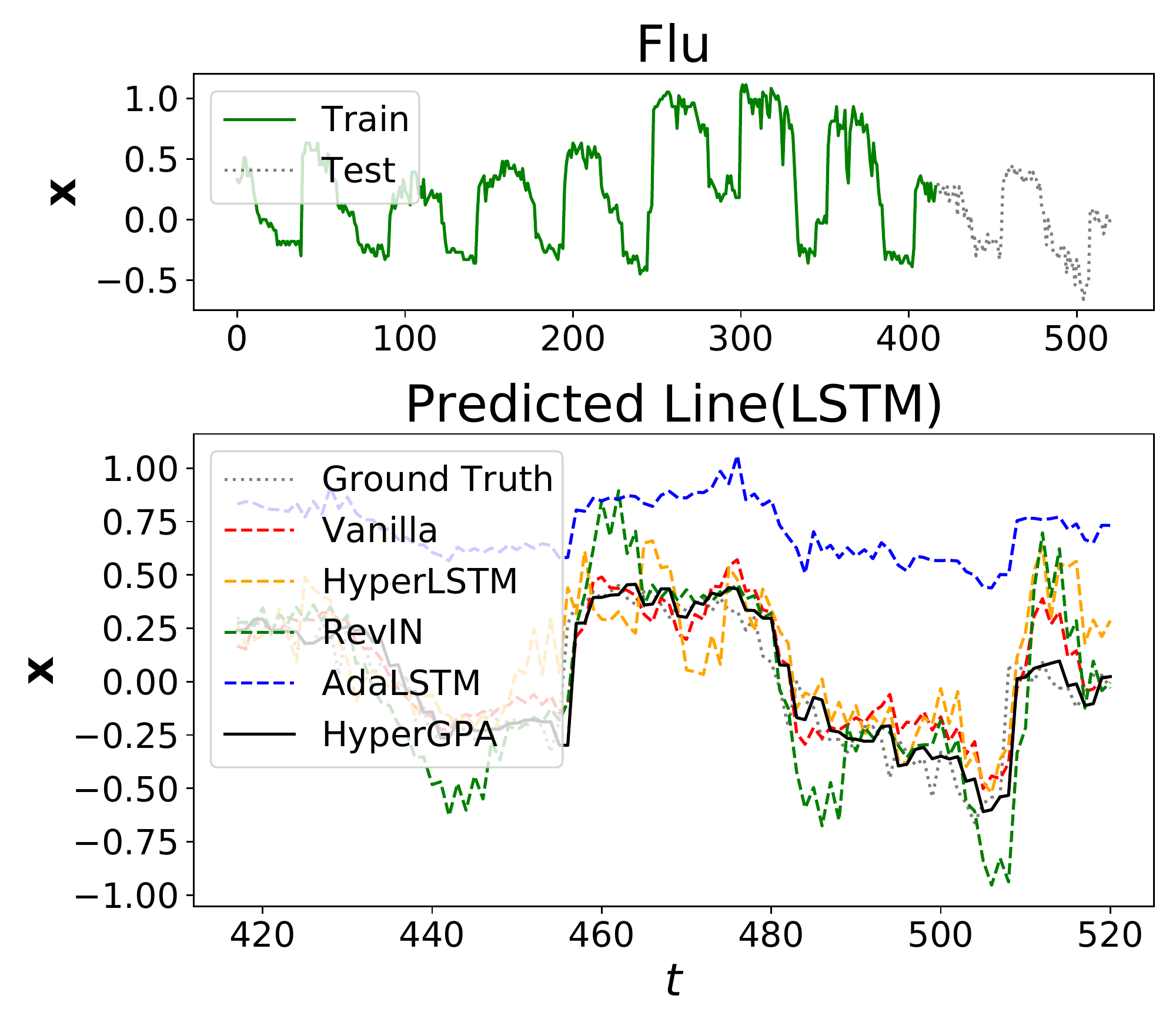}
    \includegraphics[width=0.32\columnwidth]{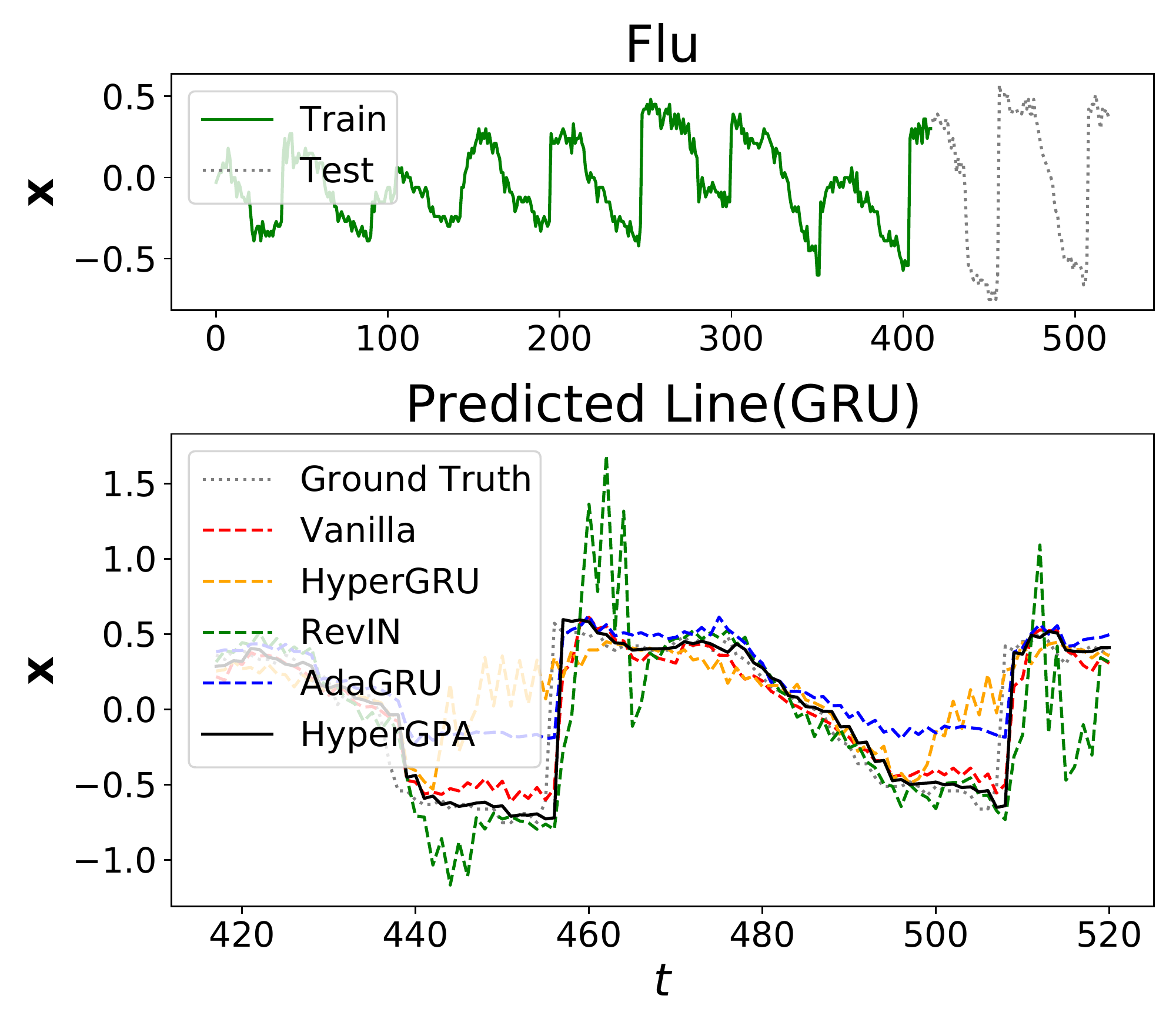}
    \includegraphics[width=0.32\columnwidth]{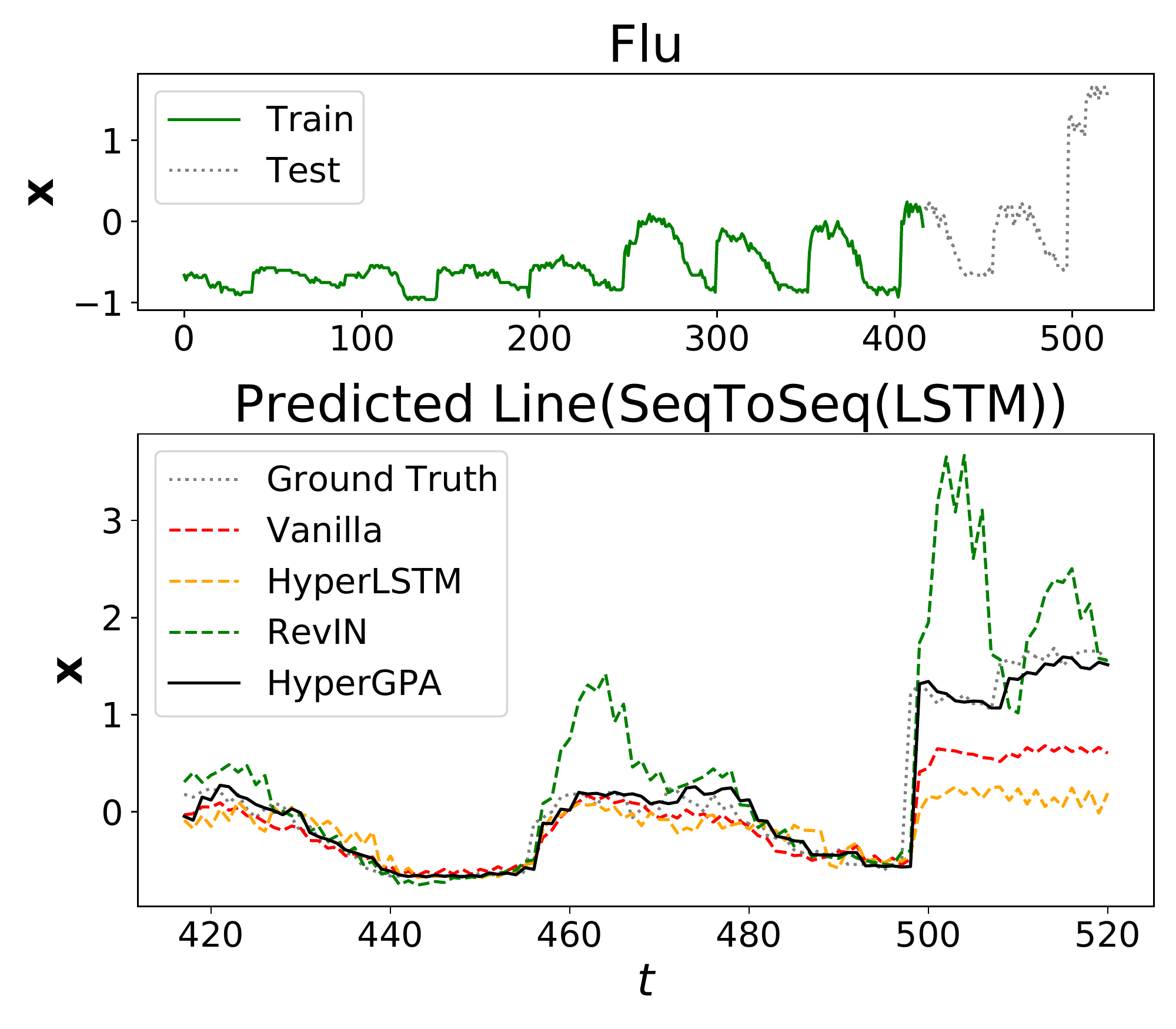}
    \includegraphics[width=0.32\columnwidth]{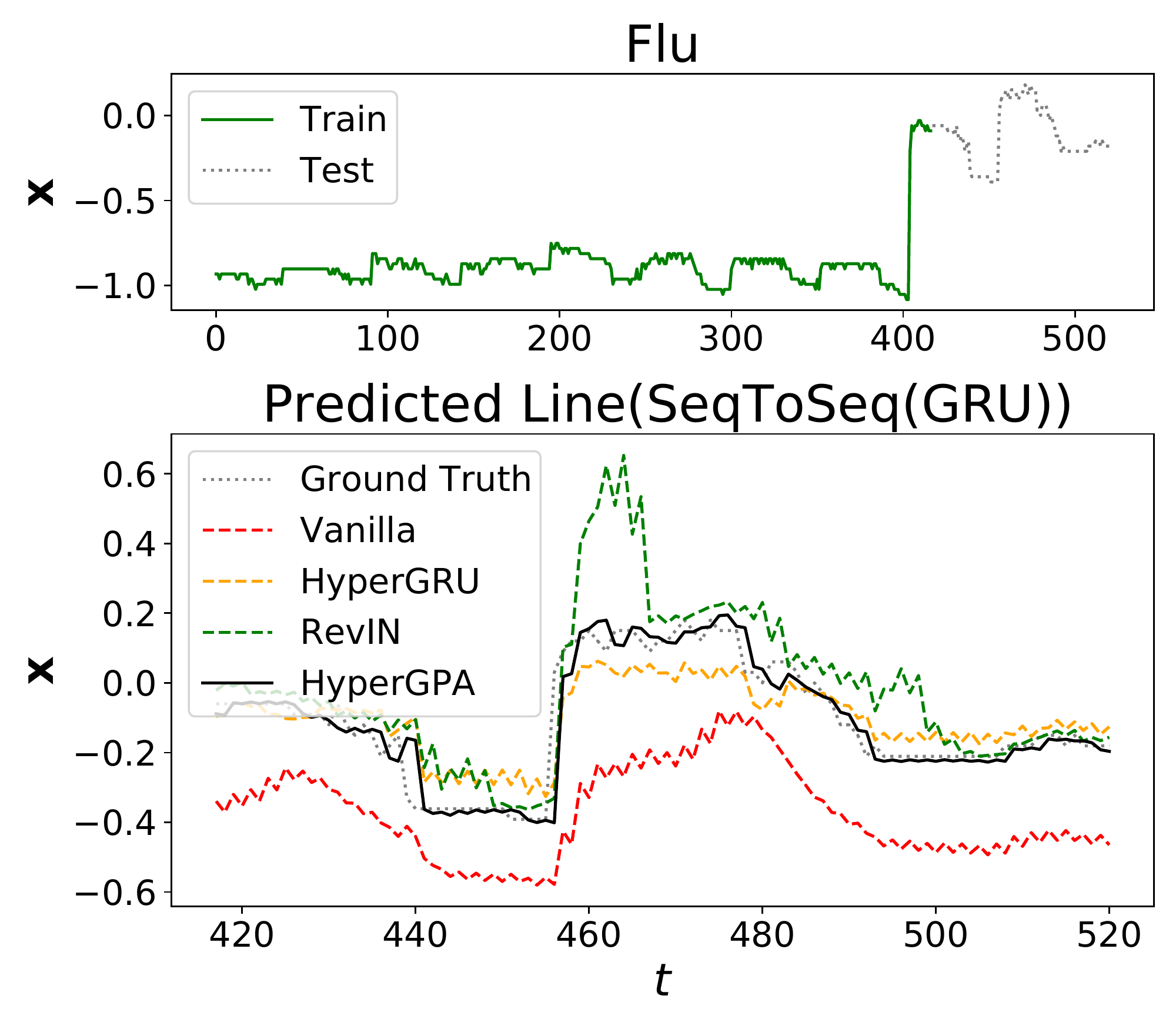}
    \includegraphics[width=0.32\columnwidth]{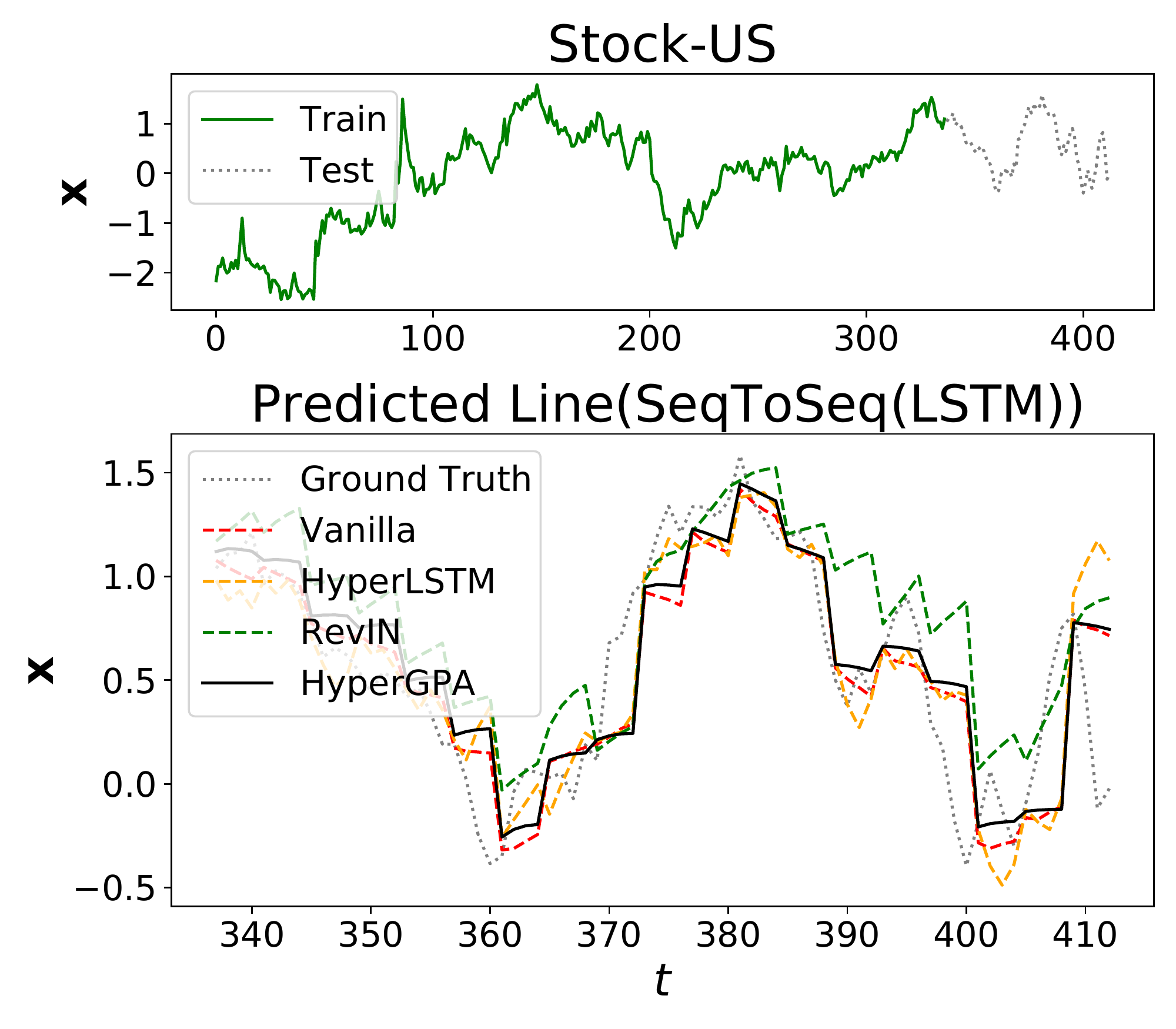}
    \includegraphics[width=0.32\columnwidth]{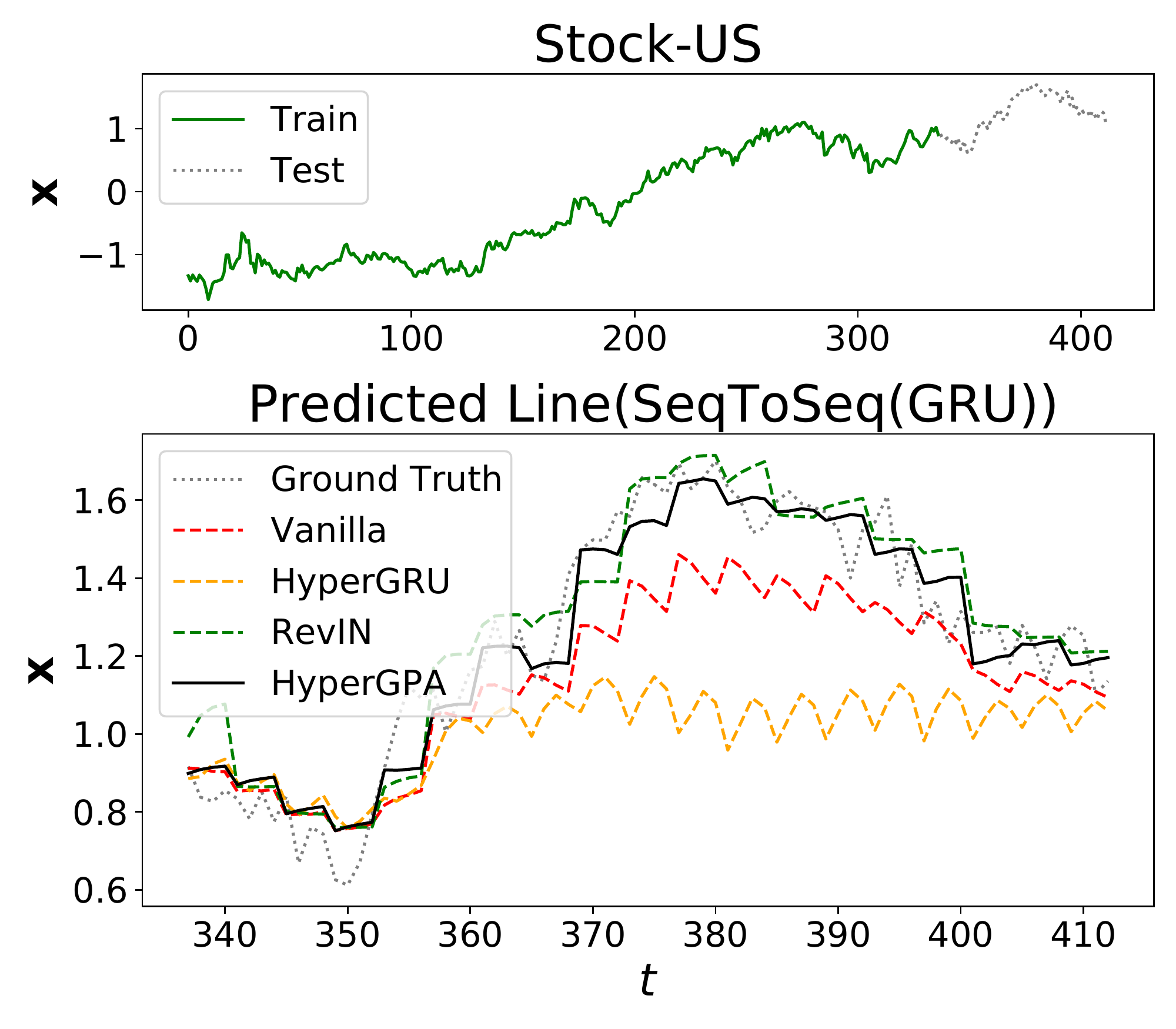}
    \includegraphics[width=0.32\columnwidth]{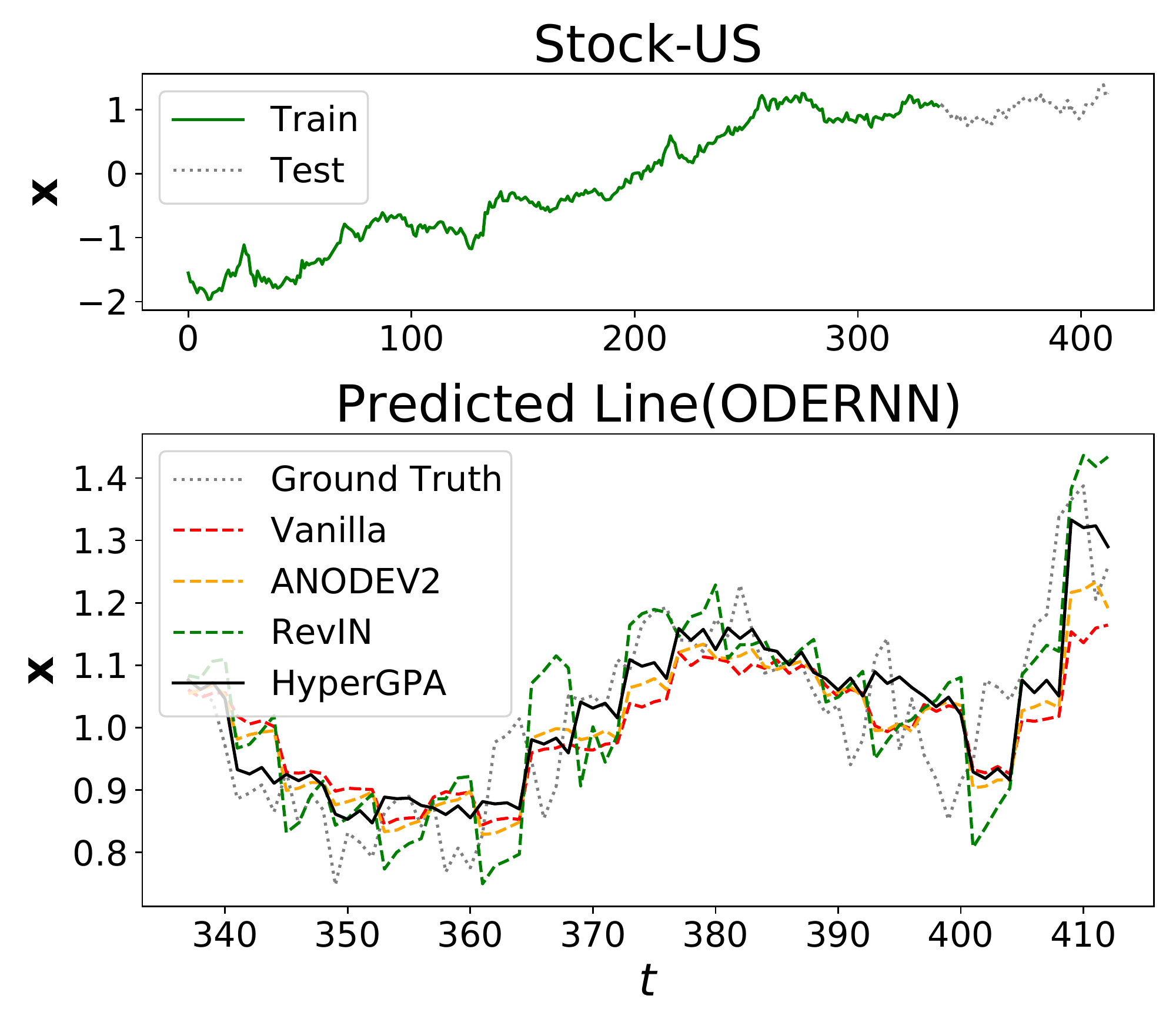}
    \includegraphics[width=0.32\columnwidth]{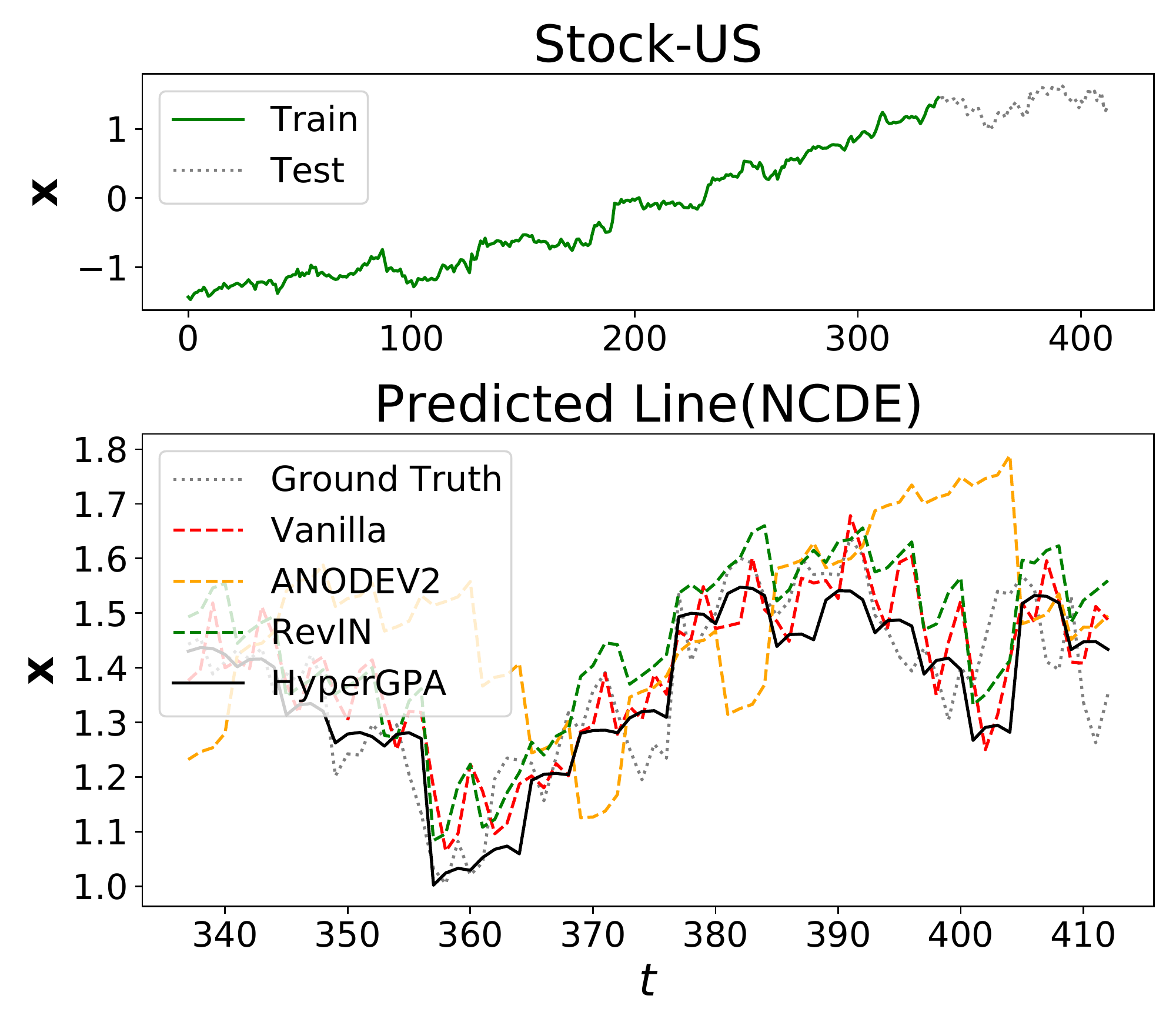}
    \includegraphics[width=0.32\columnwidth]{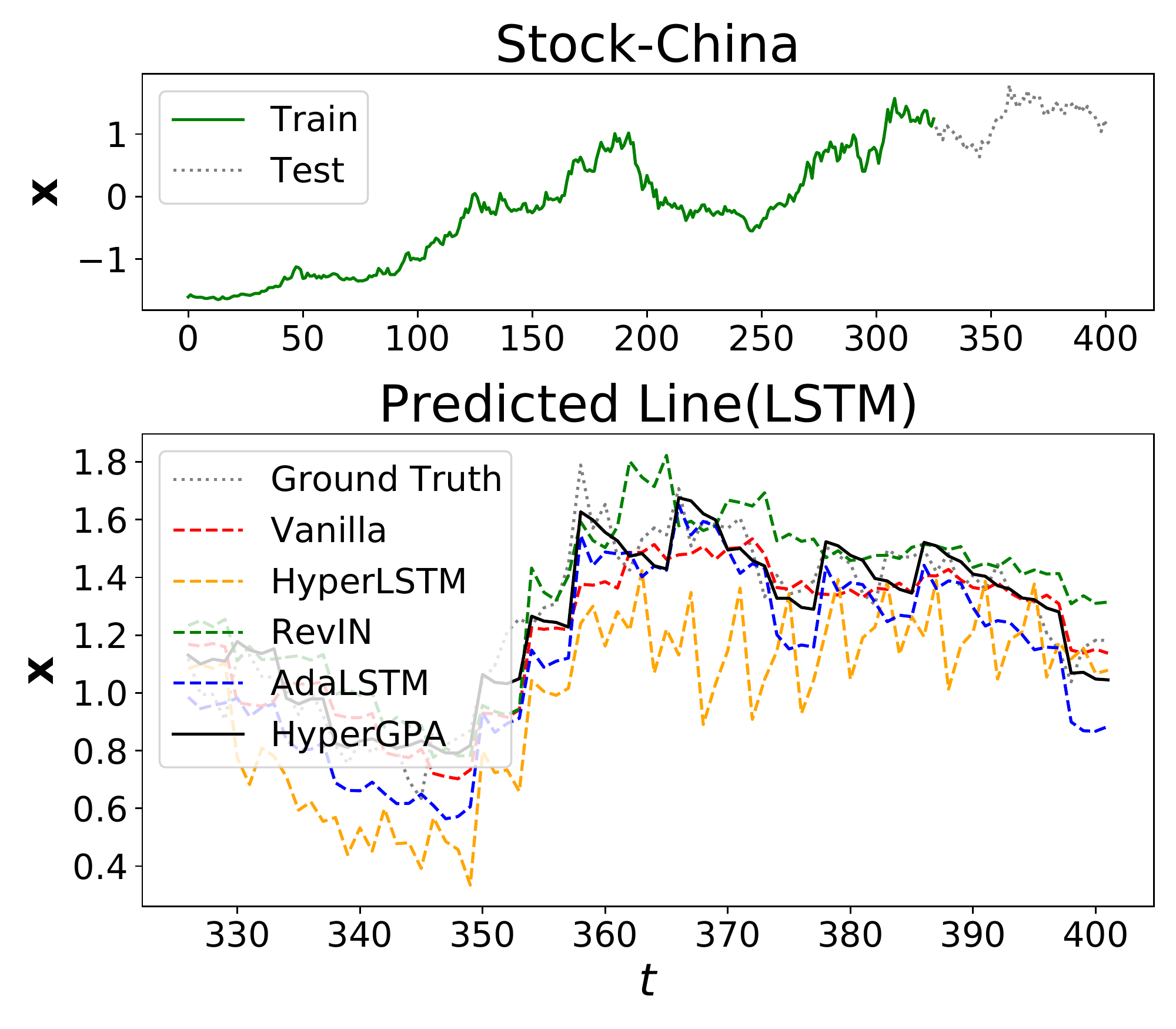}
    \includegraphics[width=0.32\columnwidth]{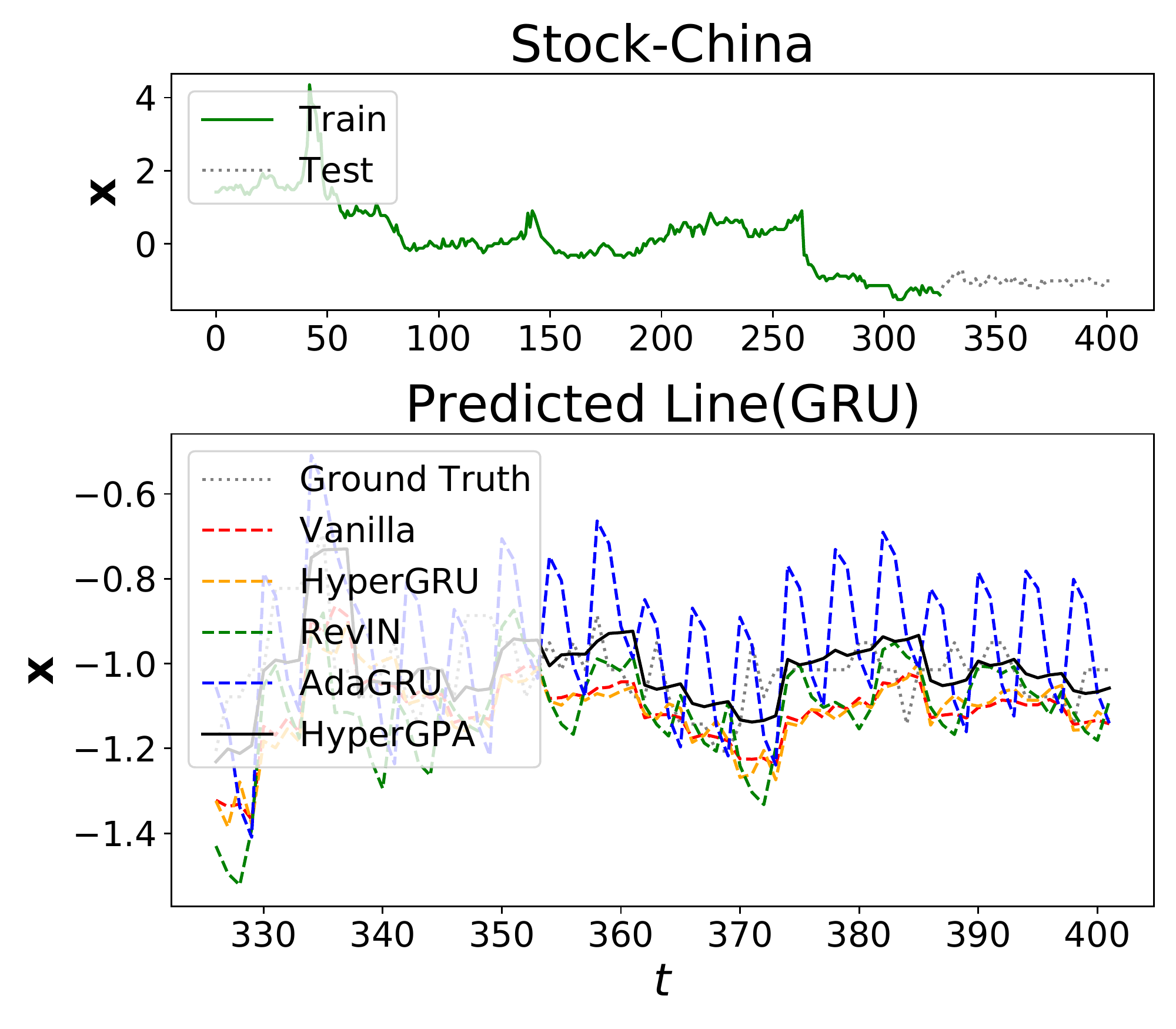}
    \includegraphics[width=0.32\columnwidth]{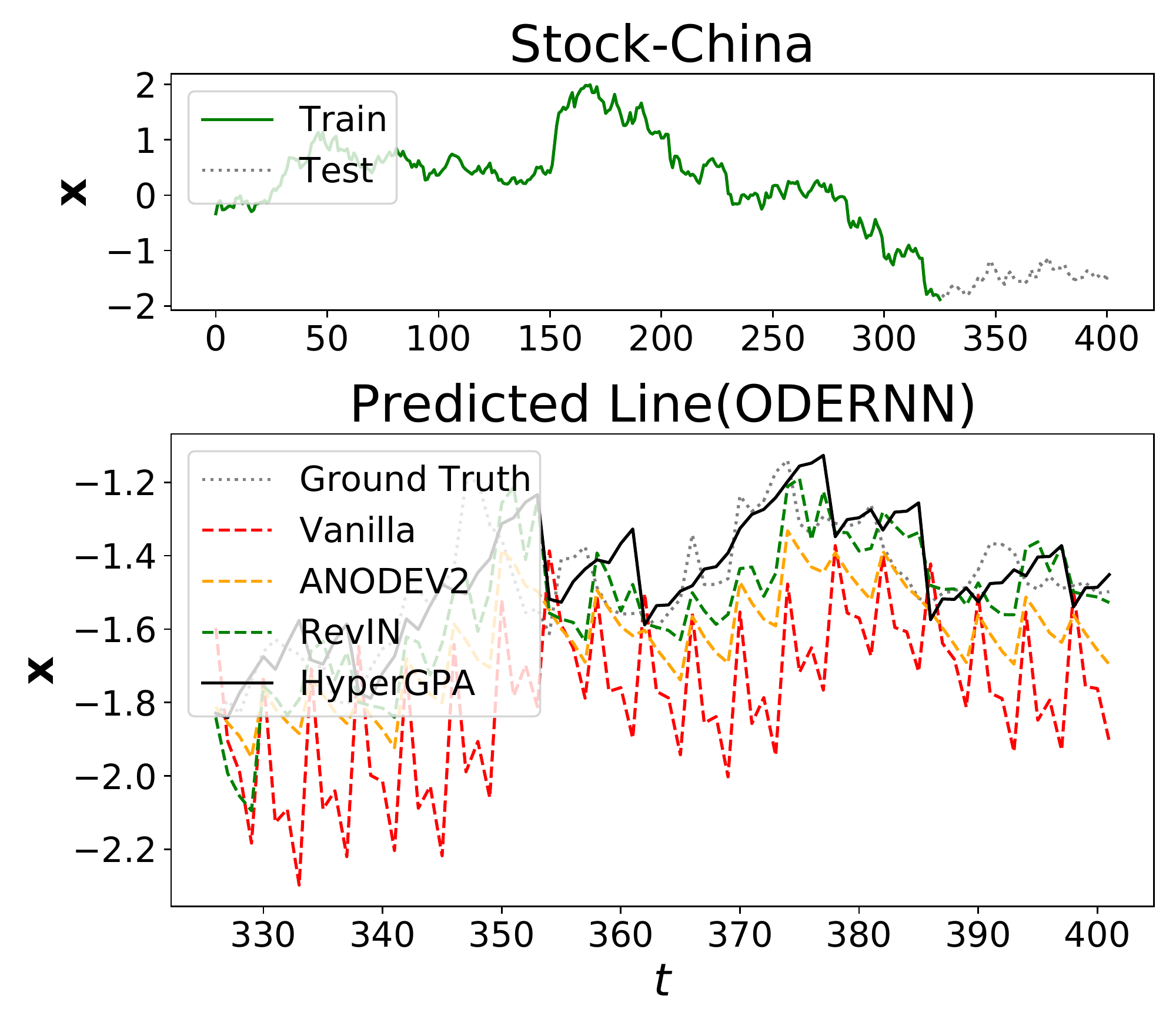}
    \includegraphics[width=0.32\columnwidth]{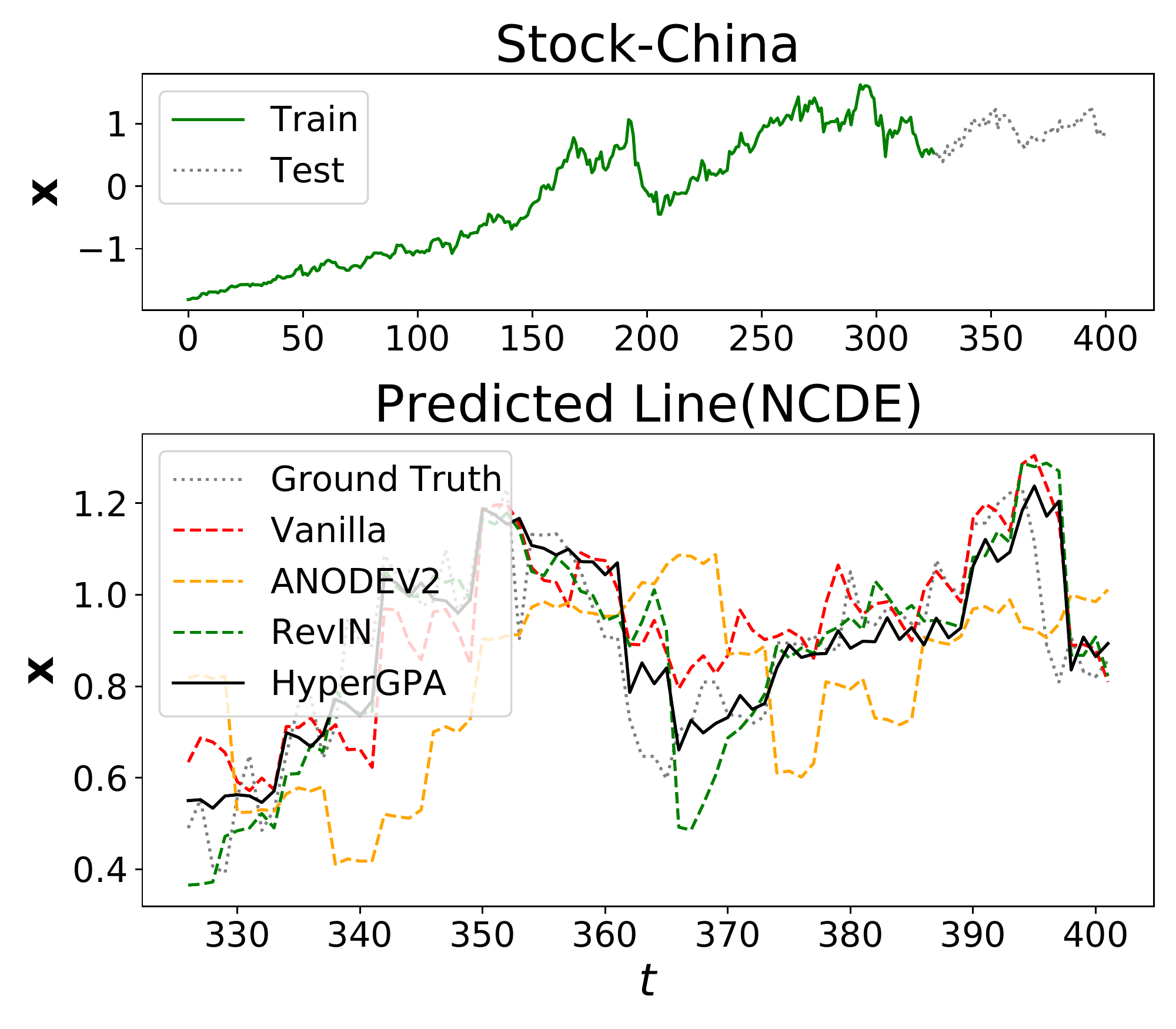}
    \caption{Forecasting visualizations of the baselines and \textit{HyperGPA}}  \label{fig:predline}
\end{figure}

\begin{figure}
    \centering
    \includegraphics[width=0.30\columnwidth]{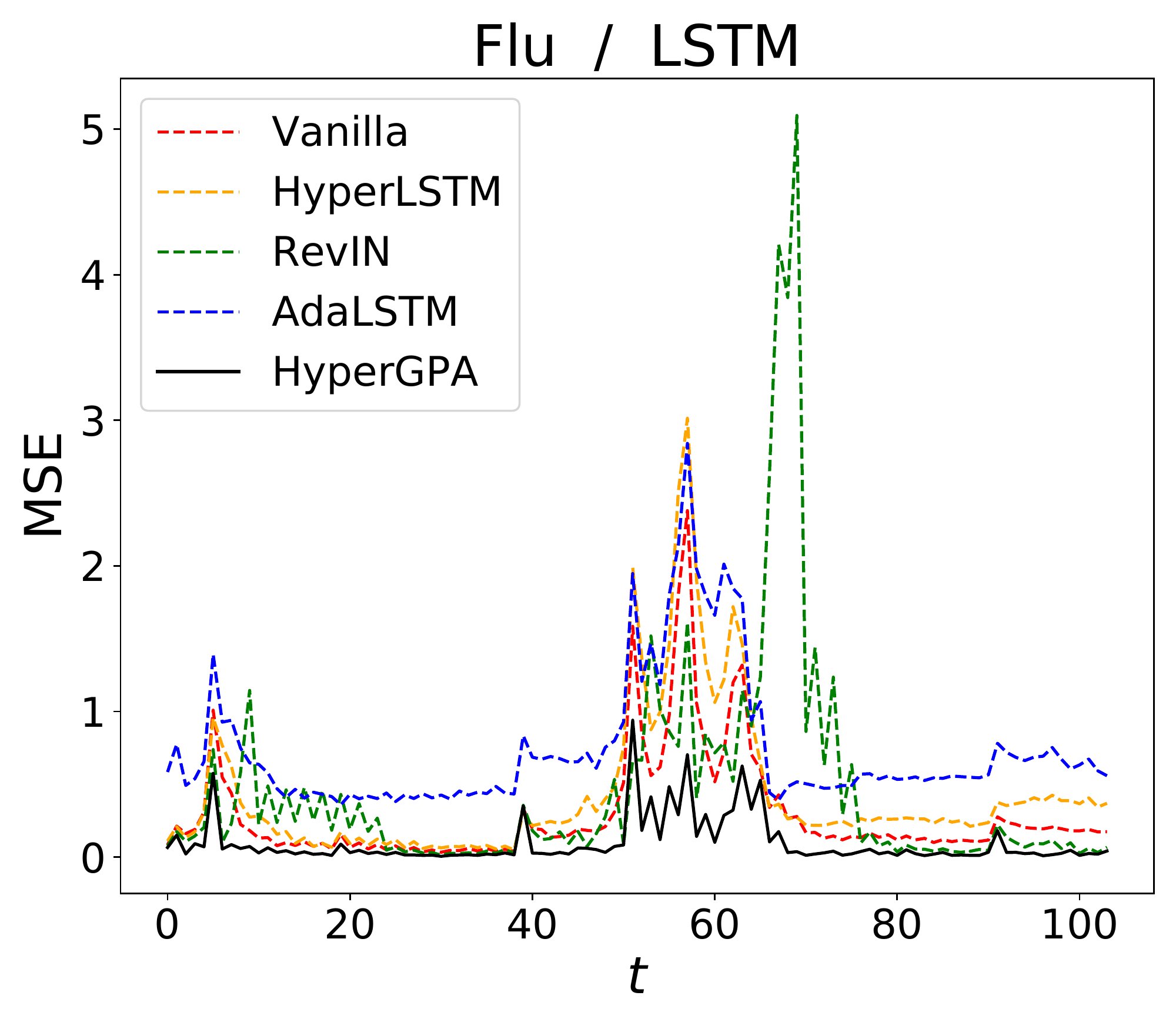}
    \includegraphics[width=0.30\columnwidth]{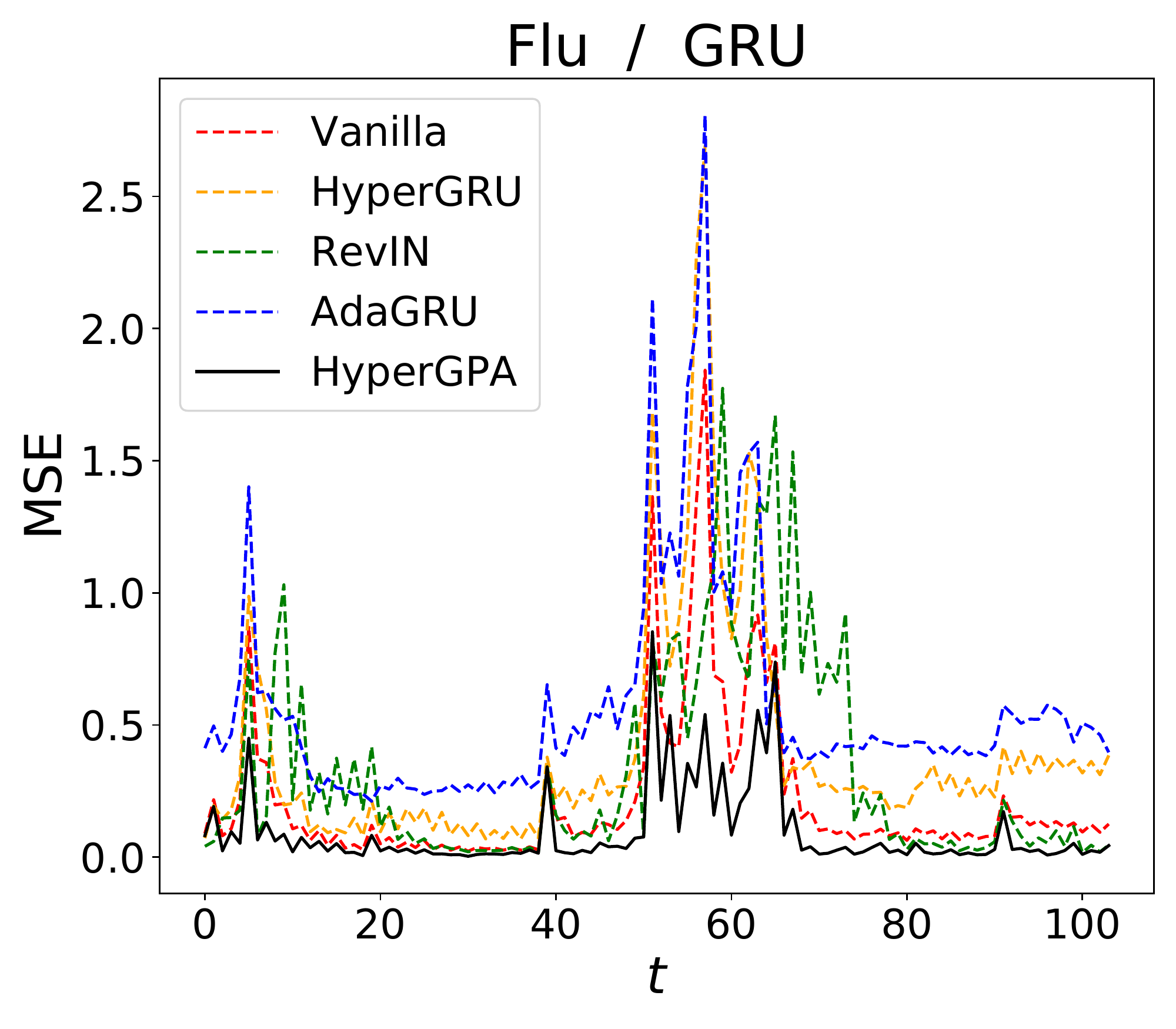}
    \includegraphics[width=0.30\columnwidth]{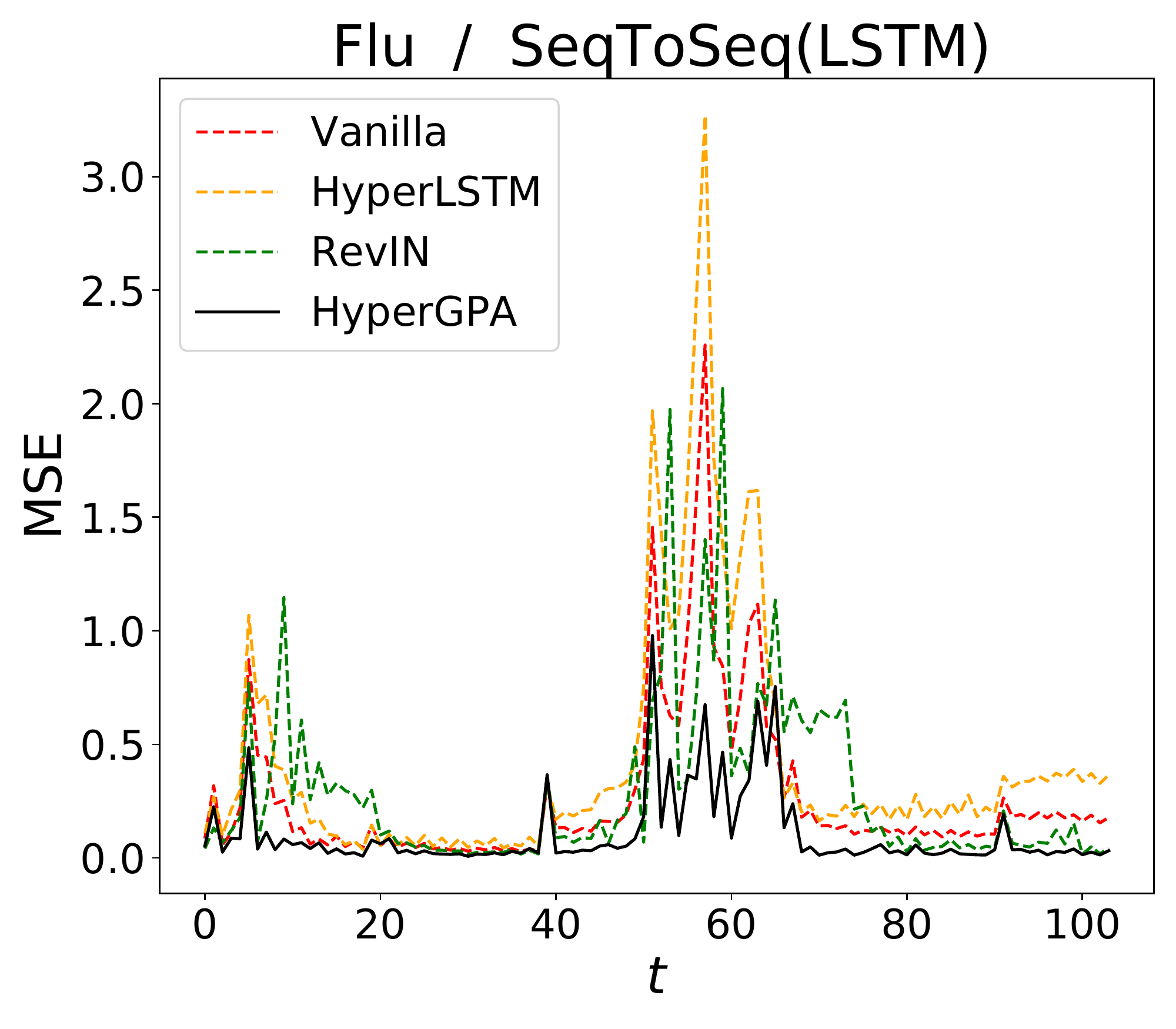}
    \includegraphics[width=0.30\columnwidth]{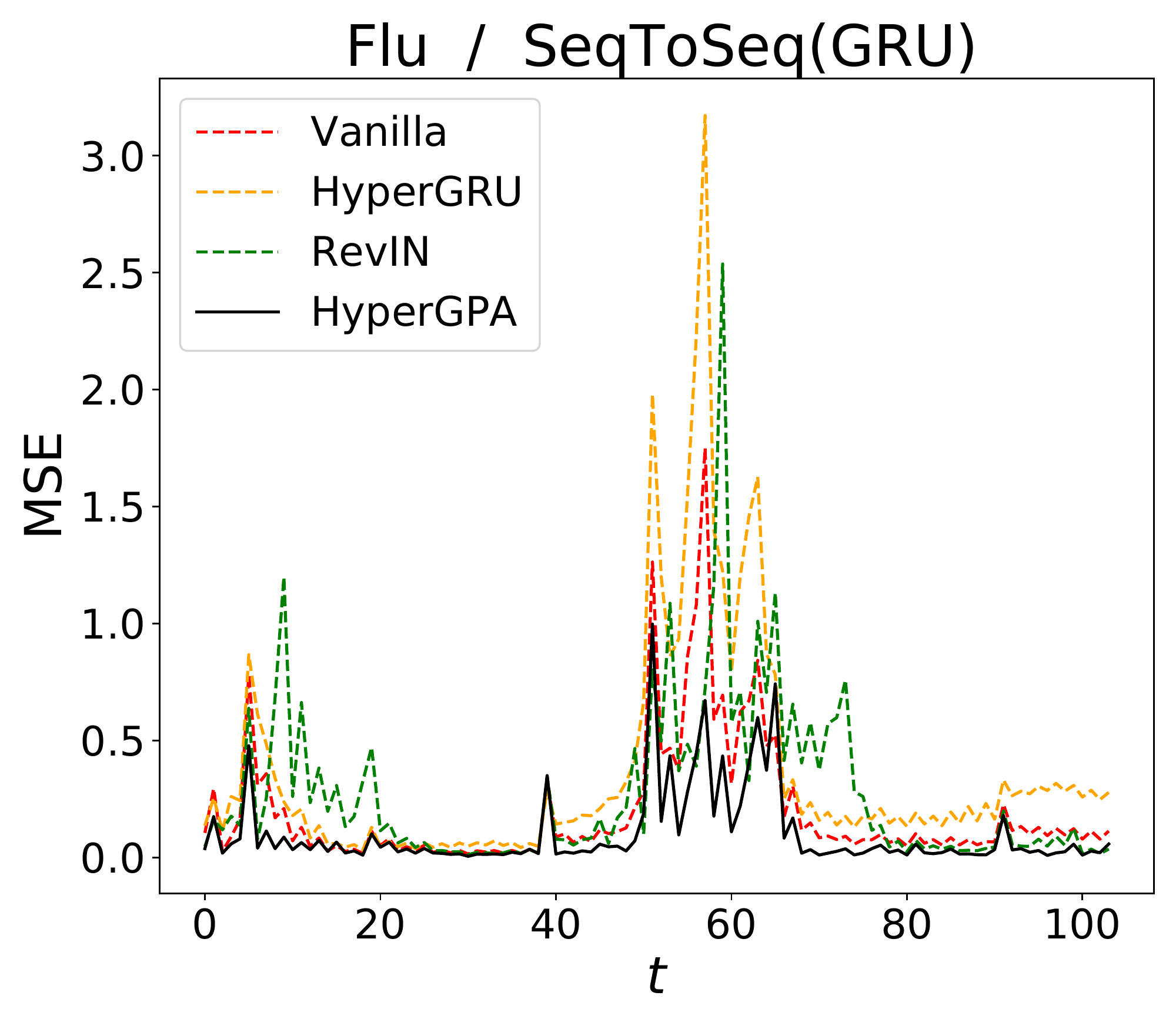}
    \includegraphics[width=0.30\columnwidth]{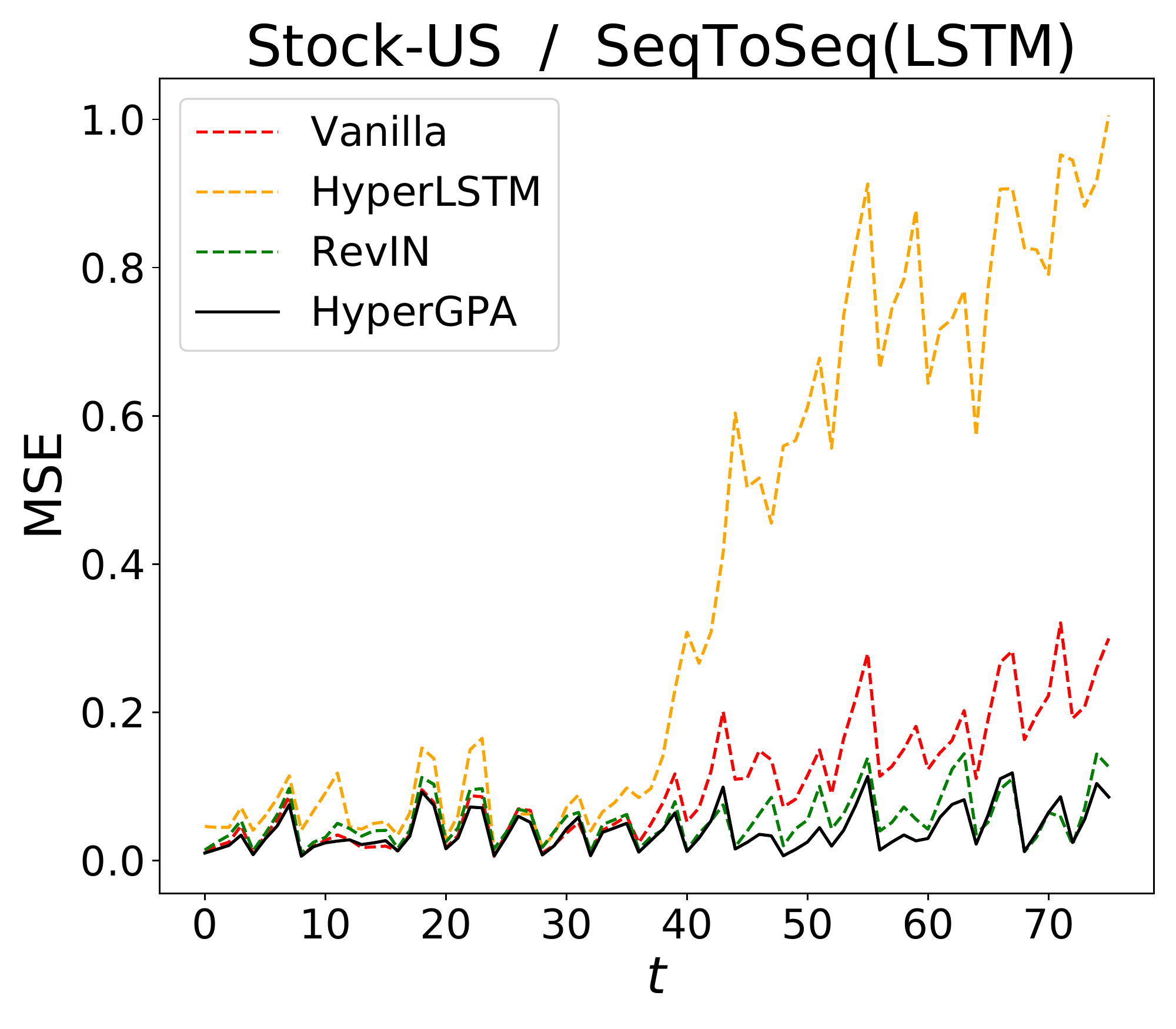}
    \includegraphics[width=0.30\columnwidth]{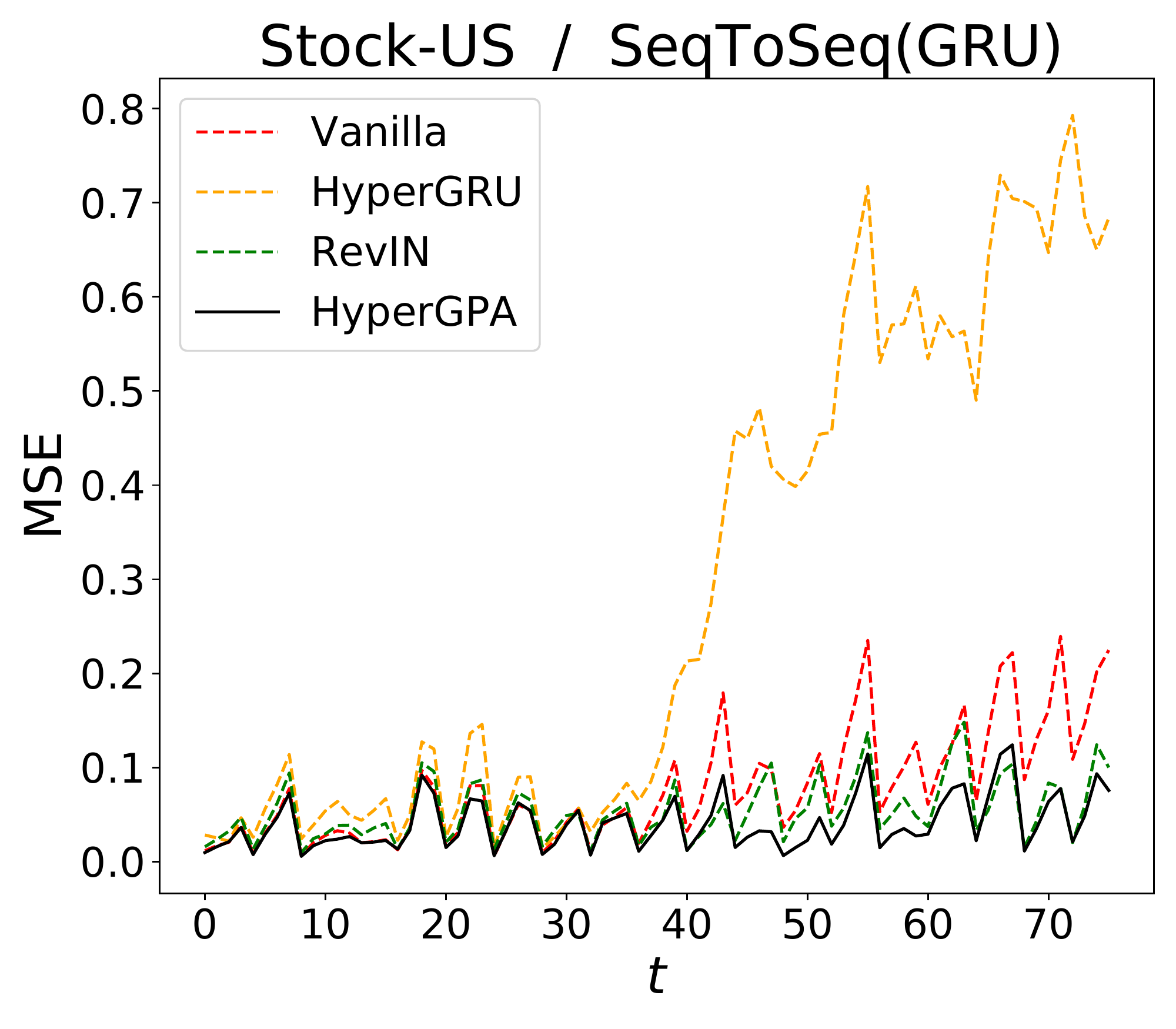}
    \includegraphics[width=0.30\columnwidth]{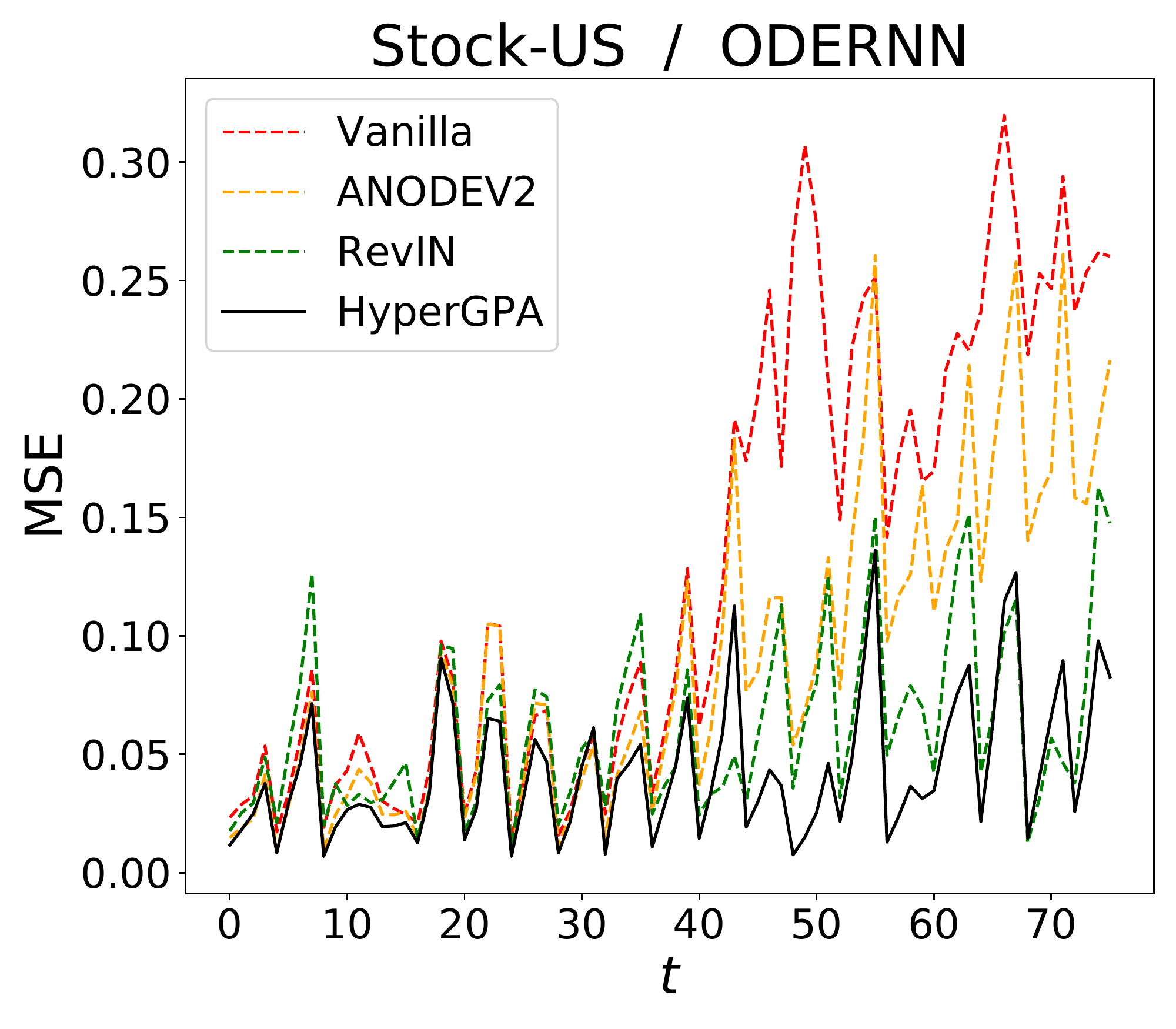}
    \includegraphics[width=0.30\columnwidth]{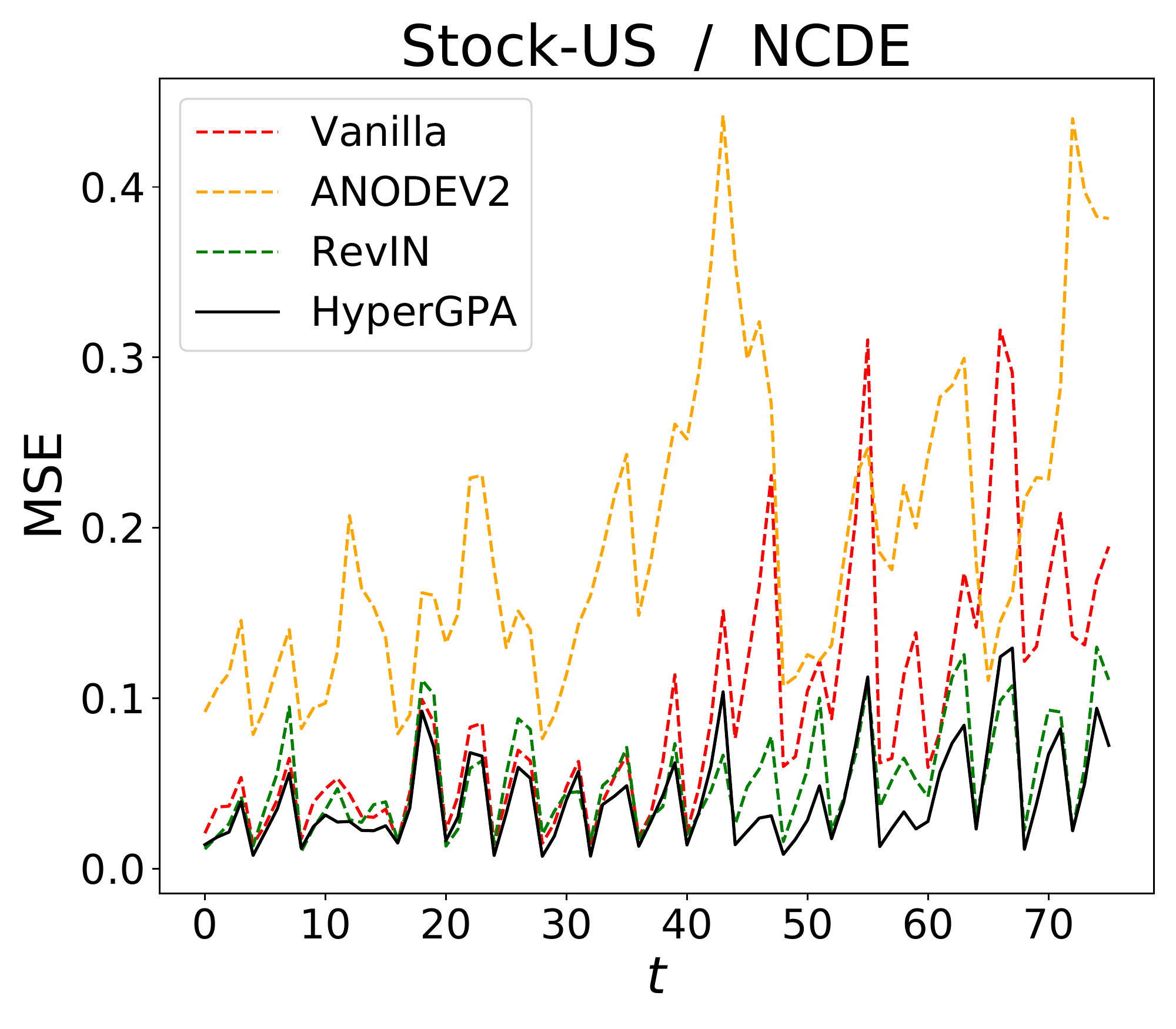}
    \includegraphics[width=0.30\columnwidth]{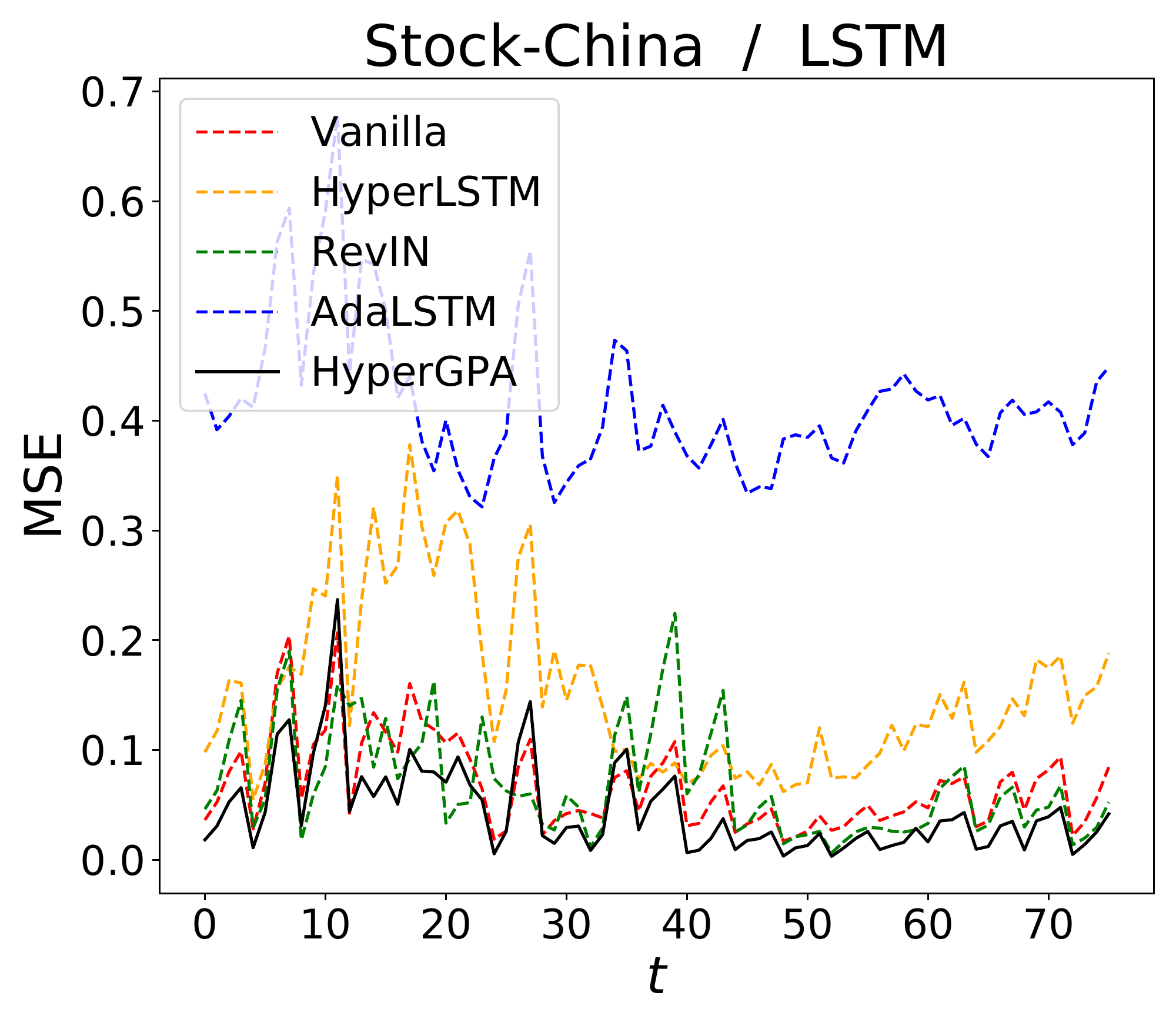}
    \includegraphics[width=0.30\columnwidth]{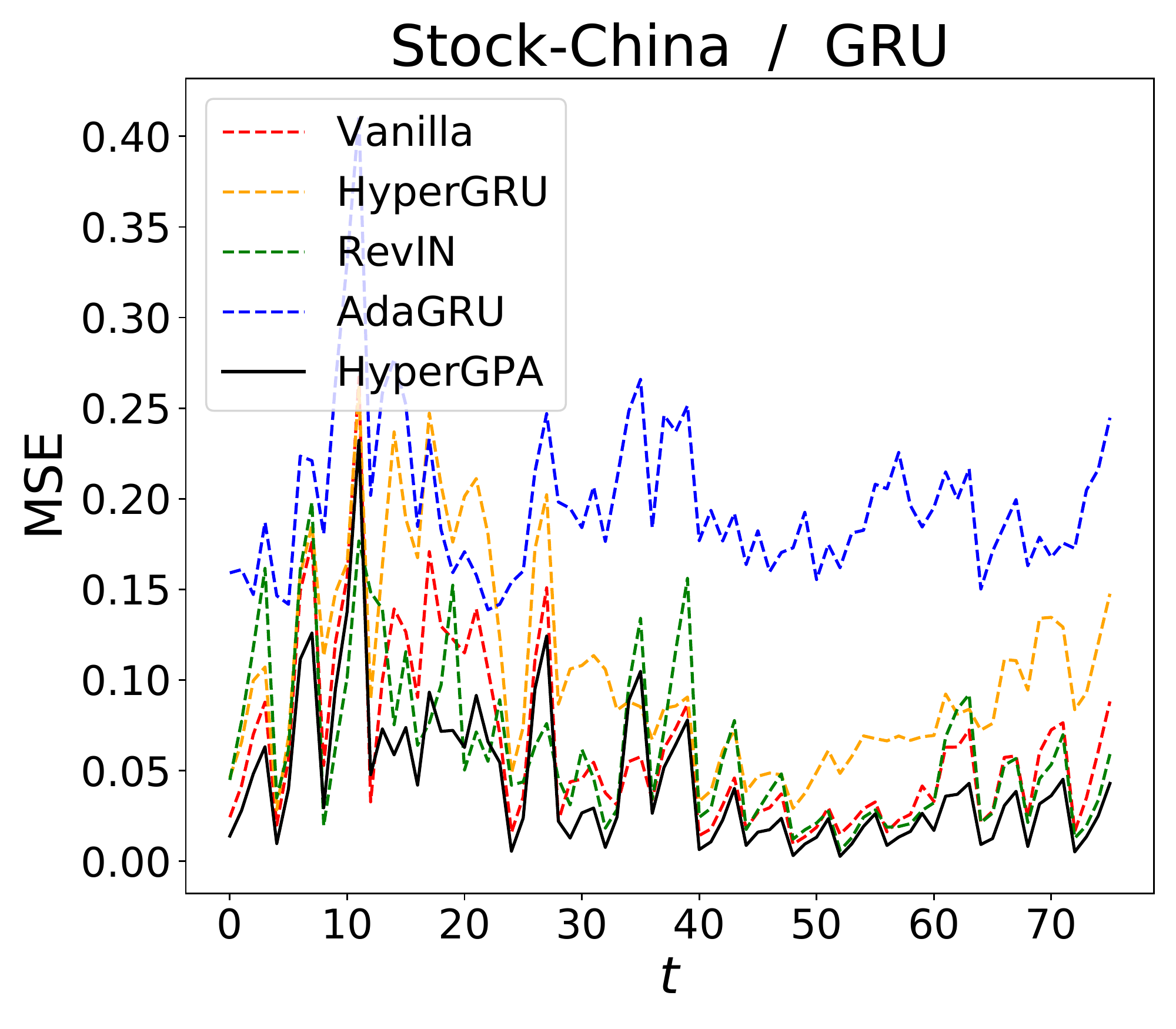}
    \includegraphics[width=0.30\columnwidth]{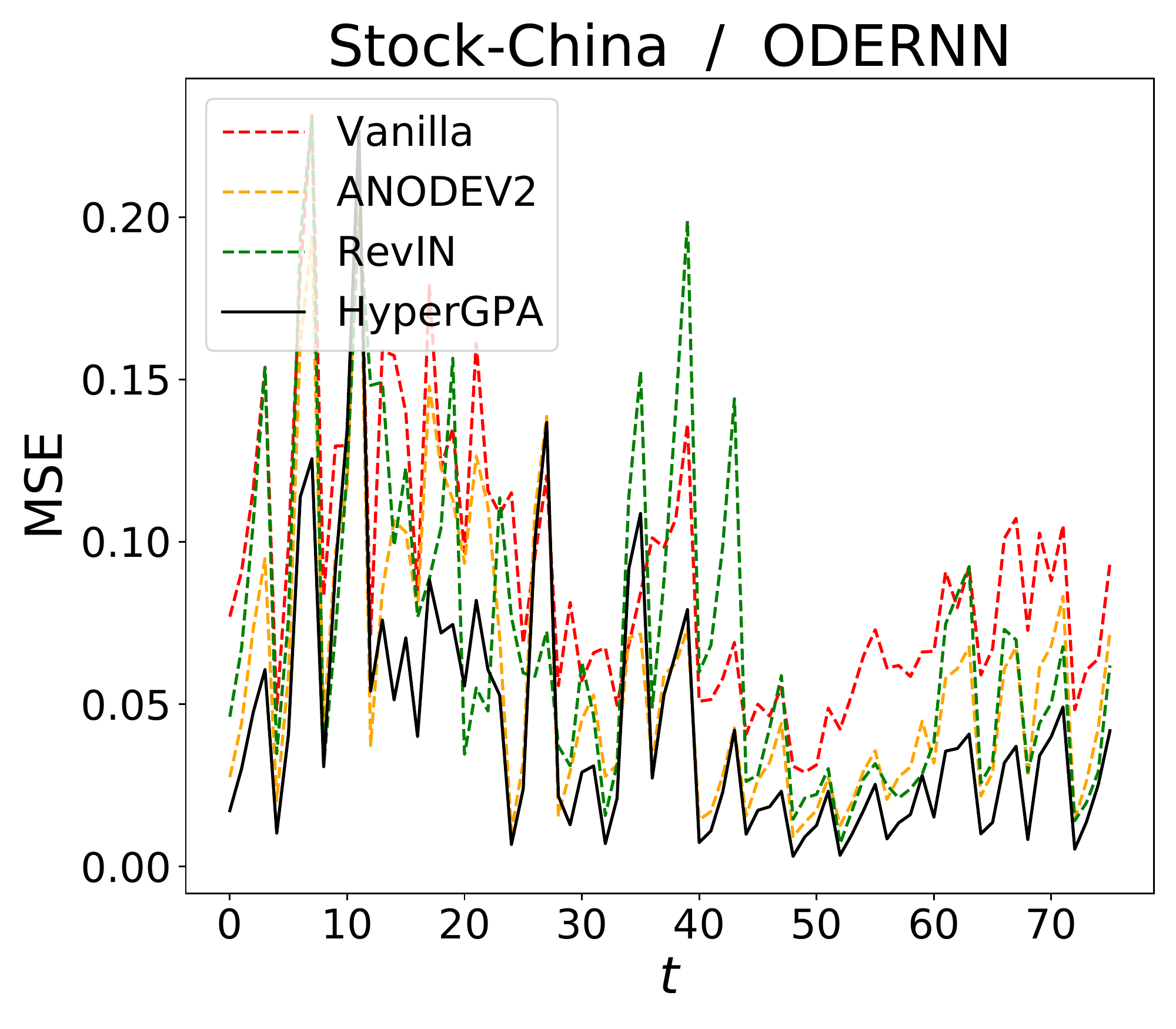}
    \includegraphics[width=0.30\columnwidth]{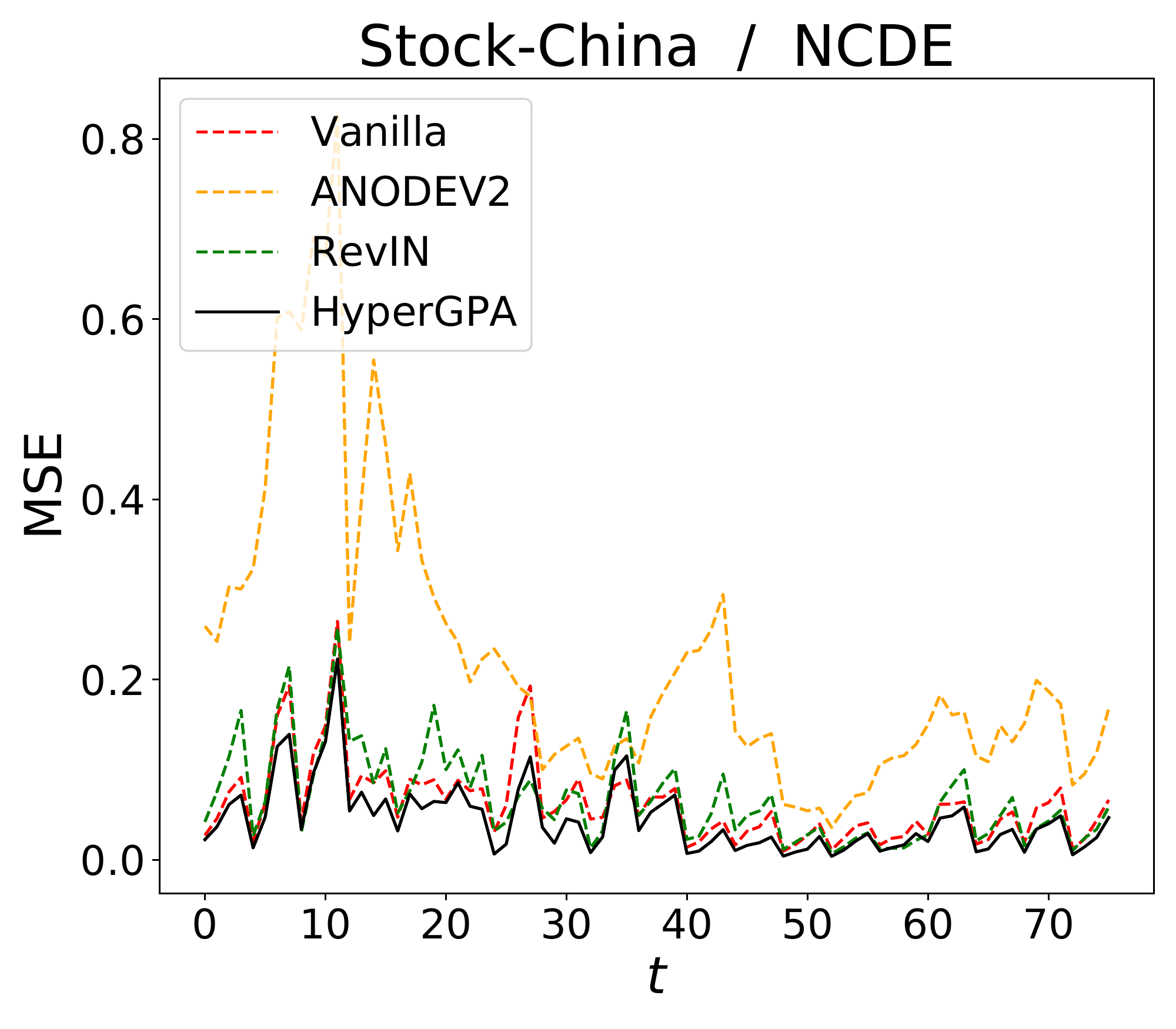}
    \caption{MSE over time}  \label{fig:mseall}
\end{figure}

Fig.~\ref{fig:predline} shows all train, test (ground-truth) and forecast lines by various baselines and \textit{HyperGPA}, and Fig.~\ref{fig:mseall} shows MSE values over time. In almost all cases, \textit{HyperGPA} follows the ground truth line more accurately than others and achieves the lowest MSE across all times. In predicted and MSE lines, lines related to \textit{HyperGPA} are solid, lines related to baseline are dashed, and lines related to ground truth are dotted. We do not include the case of \textit{ARIMA} because \textit{ARIMA} has the worst performance in \textit{Flu}, \textit{Stock-US}, and \textit{Stock-China}.

\end{document}